\documentclass[10pt,twocolumn,letterpaper]{article}

\usepackage[pagenumbers]{cvpr} 

\usepackage{graphicx}
\usepackage{amsmath}
\usepackage{amssymb}
\usepackage{booktabs}

\usepackage[table,xcdraw]{xcolor}
\usepackage{arydshln}
\usepackage[normalem]{ulem}
\useunder{\uline}{\ul}{}
\usepackage{multirow}

\usepackage[pagebackref,breaklinks,colorlinks]{hyperref}

\definecolor{myGreen}{rgb}{0, .6, .0}
\hypersetup{colorlinks,breaklinks,anchorcolor=darkblue,citecolor=myGreen}

\newcommand{\FN}{SimSC\xspace}

\usepackage[capitalize]{cleveref}
\crefname{section}{Sec.}{Secs.}
\Crefname{section}{Section}{Sections}
\Crefname{table}{Table}{Tables}
\crefname{table}{Tab.}{Tabs.}

\def\cvprPaperID{1929} 
\def\confName{CVPR}
\def\confYear{2023}

\newcommand{\xh}[1]{\textcolor{black}{{#1}}}

\begin{document}

\title{SimSC: A Simple Framework for Semantic Correspondence \\ with Temperature Learning}

\author{Xinghui Li\textsuperscript{\textdagger}  \qquad
Kai Han$\textsuperscript{\textdaggerdbl}\footnotemark[1]$
\qquad 
Xingchen Wan\textsuperscript{\textdagger} \qquad 
Victor Adrian Prisacariu\textsuperscript{\textdagger}\\[0.3em]
\textsuperscript{\textdagger}University of Oxford \qquad 
\textsuperscript{\textdaggerdbl}The University of Hong Kong\\
{\tt\small \{xinghui,xwan,victor\}@robot.ox.ac.uk \qquad kaihanx@hku.hk}
}

\twocolumn[{%
\renewcommand\twocolumn[1][]{#1}%
\maketitle
\vspace{-5mm}
\begin{center}
    \centering
    \includegraphics[width=0.95\linewidth]{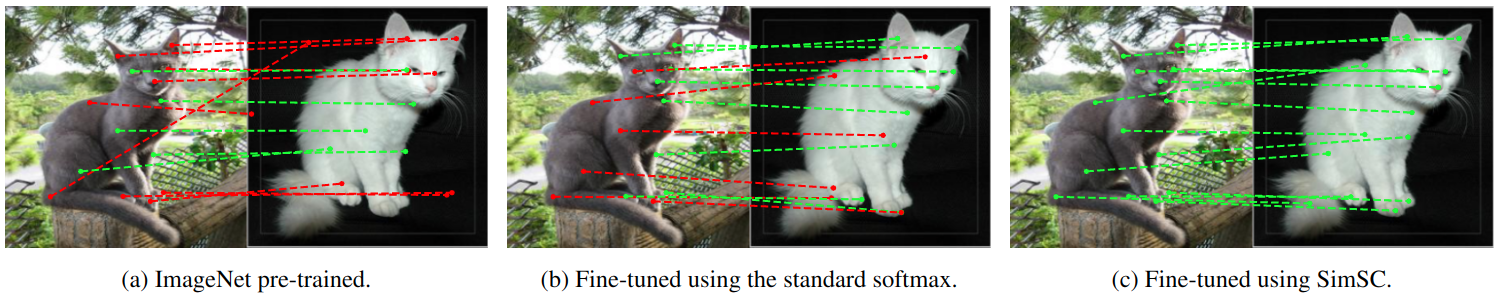}
    \captionof{figure}{Semantic correspondence matching using ResNet101 features. As illustrated, \FN significantly improves the backbone fine-tuning. The matching solely uses the features from the ResNet101 without any learned matching head.}
\end{center}\label{fig:teaser}
}]

\renewcommand{\thefootnote}{\fnsymbol{footnote}}
\footnotetext[1]{Corresponding author.}



\begin{abstract}
 We propose SimSC, a remarkably simple framework, to address the problem of semantic matching \xh{only based on the feature backbone}. \xh{We discover that when fine-tuning ImageNet pre-trained backbone on the semantic matching task, L2 normalization of the feature map, a standard procedure in feature matching, produces an overly smooth matching distribution and significantly hinders the fine-tuning process. By setting an appropriate temperature to the softmax, this over-smoothness can be alleviated and the quality of features can be substantially improved.} We employ a learning module to predict the optimal temperature for fine-tuning feature backbones. This module is trained together with the backbone and the temperature is updated online. \xh{We evaluate our method on three public datasets and demonstrate that we can achieve accuracy on par with state-of-the-art methods under the same backbone without using a learned matching head.
 Our method is versatile and works on various types of backbones. We show that the accuracy of our framework can be easily improved by coupling it with more powerful backbones.}
\end{abstract}


\section{Introduction}


Semantic correspondence matches semantically similar points between two instances of the same category. It is a challenging computer vision problem since two semantically similar points may have significant appearance differences. Semantic correspondence datasets often consist of image pairs with sparse keypoint annotations. The majority of methods in the literature have two parts \cite{ncnet, ancnet, chm, cats, scorrsan, mmnet, patchmatch, sfnet, dccnet}: a feature backbone that is pre-trained on the large-scale dataset such as ImageNet \cite{imagenet}, and a matching head. The general-purpose feature maps are extracted from source and target images by feature backbones. They are subsequently sent to the matching head, which is further trained on a semantic correspondence dataset. 

Recently, attention has been given to developing powerful matching heads. Various architectures have been proposed, ranging from high-dimensional convolutions \cite{chm, ncnet, ancnet, mmnet, patchmatch} to transformers \cite{cats, transformatcher}. The head is either directly applied to feature maps themselves \cite{sfnet, scorrsan, hpf, dhpf}, the 4D matching score space formed by feature maps \cite{chm, ncnet, ancnet, patchmatch, transformatcher} or a combination of both \cite{cats}. These methods have demonstrated promising results in capturing intra-category appearance differences. Meanwhile, the densification of the sparse supervision has also attracted recent research efforts. Through cycle-consistency, data augmentation \cite{semimatch, pwarpc} or teacher-student training strategy \cite{scorrsan}, pseudo-labels are generated to increase the supervision signal, which has also been proven to be effective in improving the performance of the model. 

As the community is developing more complicated matching heads and training losses, we look back and investigate the overlooked fundamental module: the feature backbone. One of the most straightforward solutions to the semantic matching task is to fine-tune ImageNet pre-trained feature backbones on semantic correspondence datasets. \xh{Particularly, a correlation tensor is constructed from L2 normalized feature maps and converted to a probability map by the softmax. The backbone is then supervised by the cross-entropy loss. However, such a vanilla pipeline does not yield satisfactory results. We investigate each stage of the pipeline and discover that the poor performance is mainly attributed to the over-smoothness of the probability map caused by the L2 normalization, which hinders the back-propagation of the supervision signal. By choosing an appropriate temperature in the softmax, the over-smoothness can be alleviated, and the performance of fine-tuning can be improved significantly and achieve accuracy on par with the state-of-the-art methods.}

\xh{However, a manual search for the optimal temperature is tedious and may require different searches for different datasets and feature backbones. We, therefore, propose a temperature learning module that predicts the optimal temperature based on the dense feature maps of input images. The temperature learning module is jointly trained with the backbone, and the temperature is updated and adjusted online. By doing so, we are free from the grid search for the optimal temperature for each dataset and backbone. Our method is versatile and can be applied to various types of backbone. We demonstrate that the performance of our framework can be further improved by coupling with more powerful backbones. Compared with existing methods, our framework is a backbone-only method without using any learned matching head.
}

We summarize our contributions as follows:
\begin{itemize}

    \item \xh{We discover that the L2 normalization to the feature vector causes over-smooth matching distribution and hinders the fine-tuning of backbones on semantic matching tasks. An appropriate temperature to the softmax can alleviate the over-smoothness and significantly improves the performance of the backbone}
    
    
    \item We propose a simple temperature learning module that is trained together with feature backbones and updates the temperature online in order to avoid the manual search for the optimal temperature. Unlike previous methods, our framework has a remarkably simple architecture without any learned matching head.
    
    \item We achieve performance on par with state-of-the-art methods using the same backbone on PF-Pascal and SPair-71K datasets. By coupling with iBOT \cite{ibot}, a powerful ViT-based feature backbone, the performance can be further improved and outperforms that of previous methods on PF-Pascal and SPair-71K datasets by a large margin.
    
\end{itemize}


\section{Related work}

\begin{figure*}
    \centering
    \includegraphics[width=\linewidth]{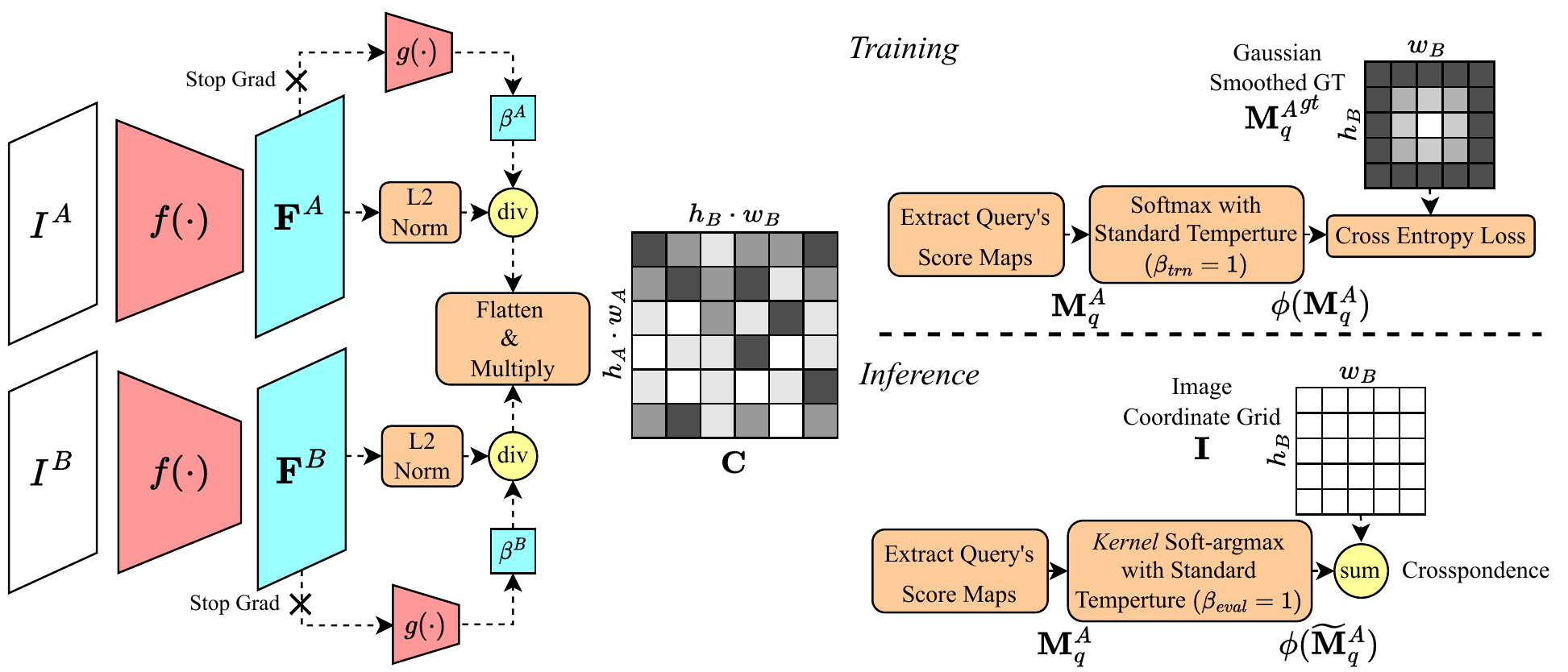}
    \caption{Overview of \FN. (i) The feature backbone $f(\cdot)$ extracts feature maps $\mathbf{F}^{A}$ and $\mathbf{F}^{B}$ from images $I^{A}$ and $I^{B}$ respectively. (ii) The temperature learning module $g(\cdot)$ predicts partial temperatures $\beta^{A}$ and $\beta^{B}$ from $\mathbf{F}^{A}$ and $\mathbf{F}^{B}$. (iii) $\mathbf{F}^{A}$ and $\mathbf{F}^{B}$ are normalized and divided by their corresponding partial temperatures and construct the matching score tensor $\mathbf{C}$. (iv) For training, the matching score map $\mathbf{M}^{A}_{q}$ of the query point is extracted from $\mathbf{C}$ and converted to a probability distribution by the standard softmax before being supervised by ground truth distribution ${\mathbf{M}^{A}_{q}}^{gt}$ using the cross-entropy loss. (v) For inference, the matching score map $\mathbf{M}^{A}_{q}$ is processed by kernel soft-argmax proposed in \cite{sfnet}. The final correspondence is the marginalization of the probability map $\phi(\widetilde{\mathbf{M}}^{A}_{q})$ multiplied with an image coordinate grid $\mathbf{I}$. Overall, our method has a simple architecture with no learned matching head compared with existing methods.}
    \label{fig:pipeline}
\end{figure*}


Semantic matching initially relied on handcrafted features such as SIFT \cite{sift} or HOG \cite{hog}. SIFT Flow \cite{siftflow} is the pioneering work that matches semantically similar objects using SIFT descriptor. Recently, dense feature maps extracted by deep neural networks \cite{vgg, resnet} have shown great success in many computer vision tasks \cite{dualrc, d2net, superpoint, loftr, superglue, xrcnet}. Han \etal's work \cite{scnet} is one of the first works that adopts the deep learning framework to the semantic correspondence problem. They learn a score function using a convolutional neural network (CNN) to rank the similarity between two patches found by region proposal and find correspondences through the probabilistic Hough matching \cite{phm}. Most other methods share a similar pipeline: dense feature extraction followed by a semantic correspondence matching head. One family of matching heads is directly applied to the feature map. Amir \etal \cite{binning} apply log binning to feature maps to increase the robustness of them. Min \etal \cite{hpf} design an algorithm to select and concatenate multiple feature maps from feature backbones to improve the performance. Their extension work, \cite{dhpf}, formulates the selection process into Gumbel-softmax \cite{gumbel} sampling making the entire process learnable. The other family of matching heads are applied to the 4D matching score maps formed by feature maps. Rocco \etal \cite{ncnet} propose 4D convolutions to filter the 4D matching score maps by enforcing a neighbourhood consensus constraint. Li \etal \cite{ancnet} propose an anisotropic 4D convolutional kernel to capture the scale variation of different objects. Min and Cho \cite{chm} use 6D convolutions and multi-layer feature maps to perform Hough matching in a learnable fashion. Kim \etal \cite{transformatcher} apply a transformer to multi-layer 4D matching score maps to perform global matching attention. Liu \etal \cite{scot} formulate the matching problem as an optimal transport problem and solve it by using the Sinkhorn algorithm. Although matching heads are becoming more advanced and powerful, they are simultaneously getting more complicated and memory-consuming.

\xh{Previous works have also identify the importance of temperature to the network on semantic matching task. An noticeable example is SFNet \cite{sfnet, sfnet_pami}. They use a grid search to find the best temperature value for their proposed kernel soft-argmax operation, a correspondence localizer with sub-pixel accuracy. Later works \cite{sfnet, cats, scorrsan, pwarpc} follow such a design. Our work however is different. They treat the temperature as one of their hyperparameters and manually search for an optimal value for it, while we analyse temperature's role in fine-tuning backbones on semantic matching task and validate our finding with experiments. In addition, we disentangle the training and inference stages to exclude the influence of the localizer from our analysis of the backbone fine-tuning, while SFNet does not differentiate these two stages.
} 

\section{Method}
In this section, we introduce \FN, which is illustrated in \cref{fig:pipeline}. Our framework has a simple architecture: a feature backbone and a two-layer multilayer perceptron (MLP). The feature backbone extracts the dense feature maps from the input image pair, and the MLP predicts the temperature for the softmax operation based on the feature maps. Correspondences are found by the marginalization of the matching probability distribution multiplied with an image coordinate grid. Compared with existing methods, our method is remarkably simple without the need for complicated matching heads, training losses, or training procedures.

In \cref{sec:method:classical_pipeline}, we briefly introduce the basic fine-tuning pipeline of feature backbones. \xh{ In \cref{sec:method:templearning}, we review the softmax, analyze the fine-tuning pipeline and provide the intuition behind the design of our temperature learning module.} We then introduce the training loss in \cref{loss} and finally how to localize the final correspondence in \cref{correspondence}.

\subsection{Learning semantic correspondence with cross-entropy loss}
\label{sec:method:classical_pipeline}
Most deep-learning feature backbones \cite{resnet, vit, ibot, dino, vgg} are trained on large datasets, such as ImageNet, on image classification tasks. Convolutional neural network (CNN) used to be the dominant architecture for feature backbones. Recently, ViT-based feature backbones, such as DINO and iBOT, demonstrate superior capability over their CNN-based counterparts on downstream tasks like classification and segmentation. We explore both types of feature backbones in our framework.

Given a pair of images $I^{A}\in \mathbb{R}^{3\times H_{A}\times W_{A}}$ and $I^{B}\in \mathbb{R}^{3\times H_{B}\times W_{B}}$, the feature backbone $f(\cdot)$ extracts dense feature maps $\mathbf{F}^{A}\in \mathbb{R}^{C\times h_{A}\times w_{A}}$ and $\mathbf{F}^{B}\in \mathbb{R}^{C\times h_{B}\times w_{B}}$ from them respectively. We then flatten the spatial dimensions of $\mathbf{F}^{A}$ and $\mathbf{F}^{B}$ and perform L$2$ normalization along the channel dimension, obtaining $\widehat{\mathbf{F}}^{A}\in \mathbb{R}^{C\times h_{A}w_{A}}$ and $\widehat{\mathbf{F}}^{B}\in \mathbb{R}^{C\times h_{B}w_{B}}$:
\begin{equation}
    \widehat{\mathbf{F}}^{A}_{i\cdot w_{A} + j} = \frac{\mathbf{f}^{A}_{ij}}{\|\mathbf{f}^{A}_{ij}\|}  \ \ \ \ \text{and} \ \ \ \   
    \widehat{\mathbf{F}}^{B}_{k\cdot w_{B} + l} = \frac{\mathbf{f}^{B}_{kl}}{\|\mathbf{f}^{B}_{kl}\|},
\label{eq:l2norm}
\end{equation}
where $\mathbf{f}^{A}_{ij}$ and $\mathbf{f}^{B}_{kl}$ are features from the $i^{th}$ row, $j^{th}$ column and $k^{th}$ row, $l^{th}$ column in the feature maps $\mathbf{F}^{A}$ and $\mathbf{F}^{B}$ respectively. Finally, we calculate the matching score tensor $\mathbf{C}\in \mathbb{R}^{h_{A}w_{A}\times h_{B}w_{B}}$ which stores the cosine similarity between every possible feature pair by:
\begin{equation}
    \mathbf{C} = (\widehat{\mathbf{F}}^{A})^{\top}(\widehat{\mathbf{F}}^{B})
\label{eq:corr}
\end{equation}
Assuming that we are matching from $I^{A}$ to $I^{B}$ and we have a set of ground-truth correspondences $\mathbf{X} = \{(\mathbf{x}^{A}_{q}, \mathbf{x}^{B}_{q})\ |\ q=1,...,n\}$, for each query point $\mathbf{x}^{A}_{q}=(x^{A}_{q}, y^{A}_{q})$, its 2D matching score map $\mathbf{M}^{A}_{q}\in \mathbb{R}^{h_{B}\times w_{B}}$ is extracted from $\mathbf{C}$ using bilinear interpolation to preserve sub-pixel accuracy.
Map $\mathbf{M}^{A}_{q}$ contains similarity scores between $\mathbf{x}^{A}_{q}$ and all features in $\mathbf{F}^{B}$. After that, the standard softmax operation $\phi(\cdot)$ without temperature (or equivalently, with a temperature value $\beta_{trn}=1$) converts $\mathbf{M}^{A}_{q}$ into a probability distribution $\phi(\mathbf{M}^{A}_{q})$ which is supervised by the ground-truth distribution ${\mathbf{M}^{A}_{q}}^{gt}$, constructed from the ground-truth correspondence $\mathbf{x}^{B}_{q}$. The model can then be trained using the cross-entropy loss:
\begin{equation}
    \mathcal{L}_{ce} = -\frac{1}{n}\sum_{q}\sum_{uv}{\mathbf{M}^{A}_{q}}^{gt}_{uv}\cdot \log(\phi(\mathbf{M}^{A}_{q})_{uv}),
    \label{eq:loss_ce}
\end{equation}
where $u$ and $v$ are row and column indices. Both $u$ and $v$ could have sub-pixel accuracy since the corresponding values can be extracted using bilinear interpolation.

\begin{figure}[t]
    \centering
    \begin{subfigure}{.33\linewidth}
        \centering
        \includegraphics[width=\linewidth]{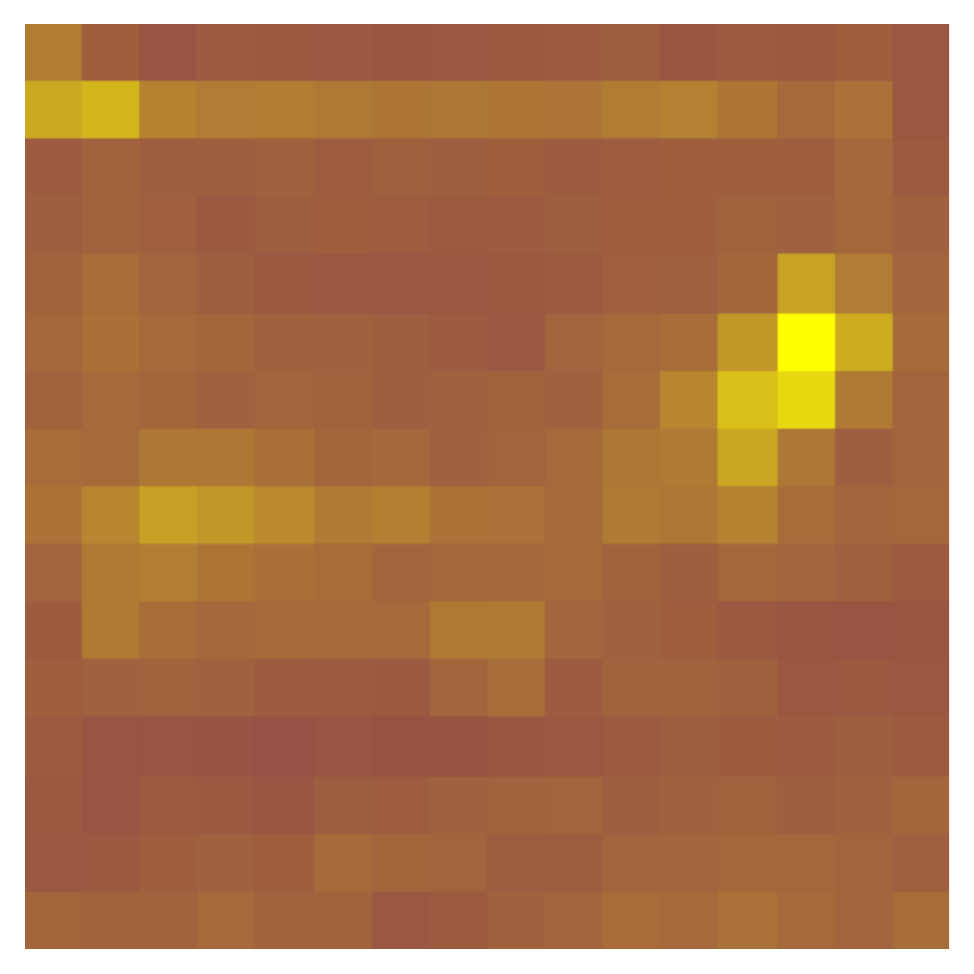}
        \caption{}
    \end{subfigure}%
    \begin{subfigure}{.33\linewidth}
        \centering
        \includegraphics[width=\linewidth]{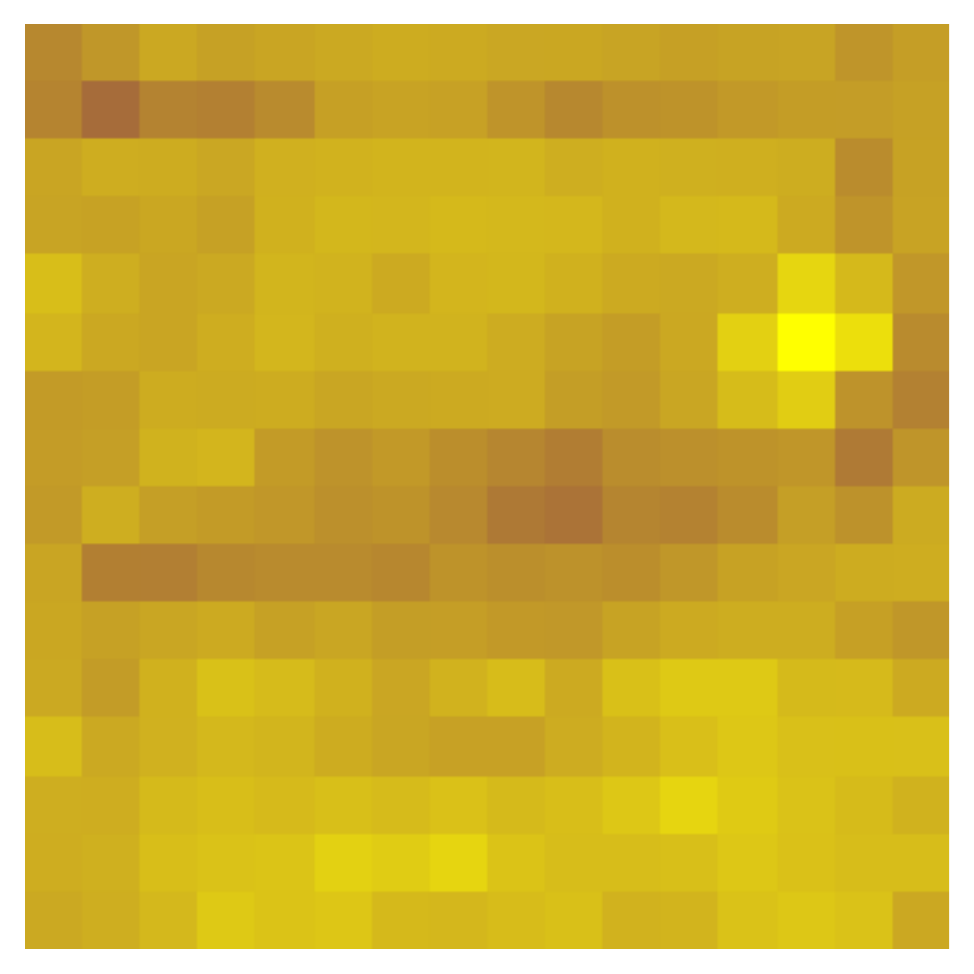}
        \caption{}
    \end{subfigure}%
    \begin{subfigure}{.33\linewidth}
        \centering
        \includegraphics[width=\linewidth]{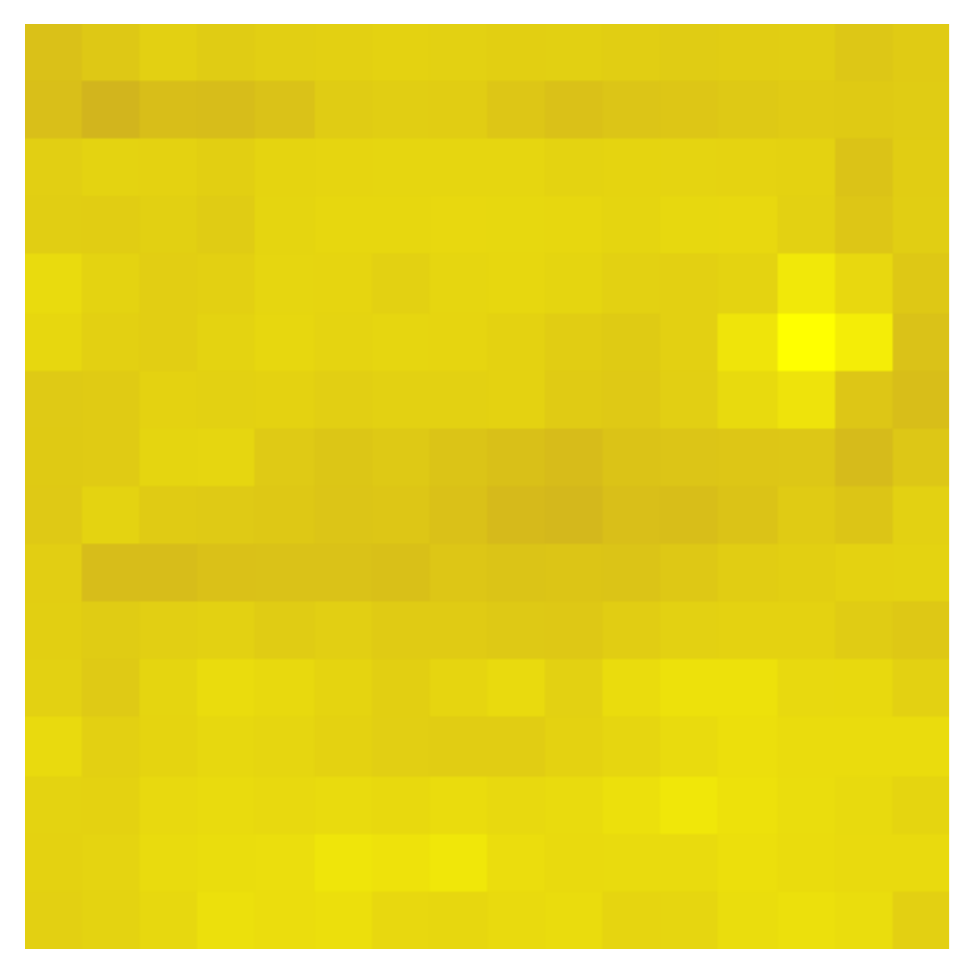}
        \caption{}
    \end{subfigure}
    \caption{(a) Correlation tensor without L2 norm. (b) Correlation tensor with L2 norm. (c) Probability map after L2 norm and the standard softmax. The final probability map is overly smooth which leads to sub-optimal fine-tuning results.}
    \label{fig:smap}
\end{figure}

\subsection{Temperature learning module}
\label{sec:method:templearning}

\xh{Before introducing the temperature learning module, we first explain how the temperature in the softmax affects its output distribution}. The softmax operation $\phi(\cdot)$ converts an input vector $\mathbf{z} = [z_{1}, z_{2}, ..., z_{n}]$ into a discrete probability distribution such that $\sum \phi(\mathbf{z})_{i}=1$. Mathematically, it is defined as:
\begin{equation}
    \phi(\mathbf{z})_{i} = \frac{e^{z_{i}/\beta}}{\sum^{n}_{i}e^{z_{i}/\beta}}
\end{equation}
where $\beta$ is the temperature. In the standard softmax operation, $\beta$ is set to $1$ and is often omitted. This is commonly used in many computer vision tasks where it transfers the output of the neural network into a probability distribution which may subsequently be supervised by the ground-truth distribution. The temperature value $\beta$ in the softmax significantly impacts the output distribution. If $\beta>1$, the absolute differences between elements in $\mathbf{z}$ and the ratio between their exponentials are suppressed, making the output distribution flatter and closer to a uniform distribution. If $\beta<1$, the absolute differences between elements in $\mathbf{z}$ and the ratio between their exponential are enhanced, making the output closer to a Dirac delta distribution. A similar analysis as above has been given in \cite{sfnet_pami, temperaturecheck, networkcalibration}.

Here, we turn to the feature maps to have a more in-depth look. \xh{In \cref{eq:l2norm}, we divide each feature by its L2 norm before calculating its cosine similarities with candidate features. This is a standard procedure in feature matching to eliminate the impact of the feature's magnitude on the similarity score. However, by doing so we simultaneously restrict the range of similarity scores in $\mathbf{M}^{A}_{q}$ between $[-1, 1]$. Since the semantic context within a natural image would not change abruptly, the actual range of similarity scores in $\mathbf{M}^{A}_{q}$ is even smaller. The exponential in the softmax operation further reduces this variance because the magnitudes of the scores are less than $1$. Eventually, the final probability distribution is excessively smooth, making the supervision harder to back-propagate to the feature backbone. An example is shown in \cref{fig:smap}.}

\xh{The over-smoothness can be alleviated in two ways. The first one is to remove the L2 normalization on feature maps. Our experiments have demonstrated that the performance of fine-tuning is improved significantly by removing the L2 norm (see \cref{exp:removel2norm}). However, it simultaneously removes the constraint imposed on the magnitude of the feature. This may lead to undesired effects such as training instability. The second option is setting an appropriate temperature in the softmax operation. As we want to increase the contrast within the probability map $\phi(\mathbf{M}^{A}_{q})$, we need to set the temperature at training stage $\beta_{trn}<1$. }

\begin{figure}
    \centering
    \includegraphics[width=0.7\linewidth]{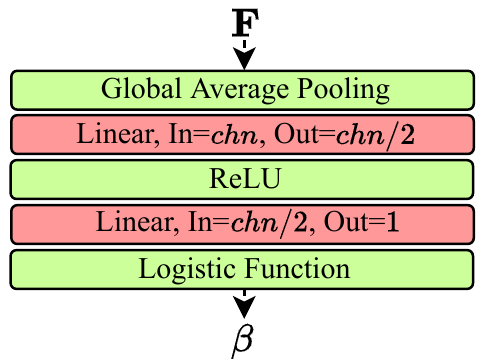}
    \caption{The architecture of the temperature learning module. It starts with the global average pooling followed by a two-layer MLP, which converts the average pooled feature map to a scalar. The logistic function transfers the scalar to the range of $[0, 1]$. In the figure, $chn$ represents the channel dimension of $\mathbf{F}$.}
    \label{fig:temp_learn}
\end{figure}

\xh{The simplest method to find the optimal $\beta_{trn}$ is by brute-force grid search.} However, this method is time-consuming since we might need to repeat the grid search for different feature backbones and datasets. Therefore, we propose a temperature learning module $g(\cdot)$ to predict the optimal temperature for each image pair. As illustrated in Figure \ref{fig:temp_learn}, the module has a very simple architecture. It first performs the global average pooling on input feature map $\mathbf{F}$ and converts it into a vector. A two-layer MLP then projects the vector to a scalar $\beta$ which is mapped to the range $[0, 1]$ by the logistic function $\mathrm{sig}(x)=1/(1+e^{x})$. 

The basic fine-tuning pipeline introduced in~\cref{sec:method:classical_pipeline} is then modified as follows: we process $\mathbf{F}^{A}$ and $\mathbf{F}^{B}$ with $g(\cdot)$ and obtain $\beta^{A} = g(\mathbf{F}^{A})$ and $\beta^{B}=g(\mathbf{F}^{B})$. The normalized feature maps $\widehat{\mathbf{F}}^{A}$ and $\widehat{\mathbf{F}}^{B}$ are then divided by $\beta^{A}$ and $\beta^{B}$ respectively. We finally calculate the matching score tensor $\mathbf{C}$ and replace \cref{eq:corr} with:
\begin{equation}
    \mathbf{C} = \Bigl(\frac{\widehat{\mathbf{F}}^{A}}{\beta^{A}}\Bigr)^{\top}\Bigl(\frac{\widehat{\mathbf{F}}^{B}}{\beta^{B}}\Bigr).
\label{eq:corr_beta}
\end{equation}
By doing so, our method automatically set the temperature value as $\beta_{trn} = \beta^{A}\cdot \beta^{B}$, so we no longer need to set a manual temperature value in the softmax operation.

\subsection{Training objective}
\label{loss}

\begin{figure}[t]
    \centering
    \includegraphics[width=0.9\linewidth]{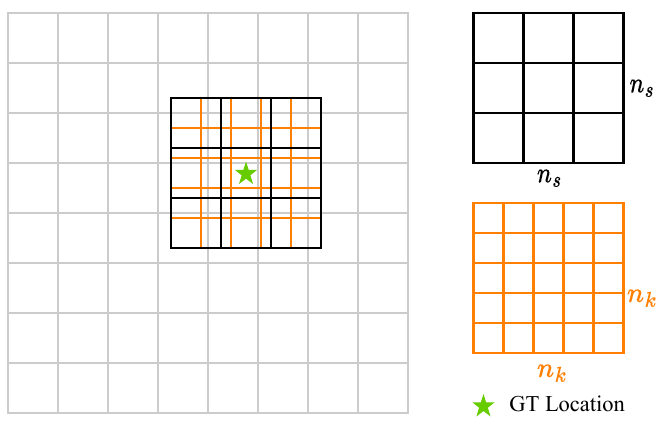}
    \caption{Generation of ground-truth distribution with sub-pixel accuracy. The ground-truth distribution is a $n_{s}\times n_{s}$ region centered at the ground-truth location. Such a region fits with a $n_{k}\times n_{k}$ Gaussian kernel. The rest part is all zero.}
    \label{fig:gt}
\end{figure}

The training objective is the cross-entropy loss between predicted matching probability distribution $\phi(\mathbf{M}^{A}_{q})$ and the ground-truth distribution ${\mathbf{M}^{A}_{q}}^{gt}$, as shown in  \cref{eq:loss_ce}. To reduce overfitting, instead of using a one-hot distribution as the ground truth, we pick the $n_{s}\times n_{s}$ region centered at the ground-truth correspondence and fit a $n_{k}\times n_{k}$  Gaussian kernel with the standard deviation $\lfloor n_{k} / 2 \rfloor$ in it.
We do not round the position of the ground-truth correspondence as we can estimate the sub-pixel values using bilinear interpolation. We set $n_{k}>n_{s}$ to further smooth the ground-truth neighbourhood. Figure \ref{fig:gt} illustrates this process. 

\xh{Occasionally, we observe the backbone has degenerate behavior when rapidly decreasing the temperature, which leads to the collapse of the training in the early stages. Therefore, we propose a temperature regularization loss $\mathcal{L}_{reg}(\beta)$ to prevent the temperature from being too close to $0$}: 
\begin{equation}
    \mathcal{L}_{reg}(\beta) = \max(0, -\log(\beta) + \log(\beta_{thres})),
    \label{eq:l_reg}
\end{equation}
where $\beta$ is the output of temperature learning module $g(\cdot)$ and $\beta_{thres}$ is a constant threshold. If $\beta > \beta_{thres}$, $-\log(\beta) + \log(\beta_{thres}) < 0$ and $\mathcal{L}_{reg} = 0$ so the regularization loss does not have any effect. If $\beta < \beta_{thres}$, $\mathcal{L}_{reg}(\beta) = -\log(\beta) + \log(\beta_{thres})$ takes effect to penalize $\beta$ for being too small and forces it to become larger. Overall, the training loss can be written as:
\begin{equation}
    \mathcal{L} = \mathcal{L}_{ce} + \gamma\mathcal{L}_{reg},
    \label{eq:loss}
\end{equation}
where $\gamma$ is the weight for the regularization. 

\subsection{Localization of correspondences}
\label{correspondence}
During inference, in order to achieve the sub-pixel accuracy, we use the kernel soft-argmax proposed in \cite{sfnet}. Given the query $(x^{A}_{q}, y^{B}_{q})$, its matching score map $\mathbf{M}^{A}_{q}$ is extracted from matching score tensor $\mathbf{C}$ using bilinear interpolation. It then multiplies with a Gaussian kernel $\mathbf{G}\in \mathbb{R}^{h_{B}\times w_{B}}$ with the mean being the maximum of $\mathbf{M}^{A}_{q}$ and the standard deviation of $\sigma$ to suppress the non-maximum modes in $\mathbf{M}^{A}_{q}$:
\begin{equation}
    \widetilde{\mathbf{M}}^{A}_{q} = \mathbf{G} \ast \mathbf{M}^{A}_{q},
    \label{eq:infer_kernel}
\end{equation}
where $\ast$ is the element-wise multiplication between matrices. Note that, since we have already rescaled $\mathbf{C}$ by $\beta^{A}\cdot \beta^{B}$, we do not need to manually set a temperature value $\beta_{eval}$ for evaluation but use $\beta_{eval}=1$. The standard softmax operation directly converts $\widetilde{\mathbf{M}}^{A}_{q}$ to a probability distribution, and the final correspondence can be obtained by:
\begin{equation}
    \bar{\mathbf{x}}^{B}_{q} = \sum_{kl}\phi(\widetilde{\mathbf{M}}^{A}_{q})_{kl}\ast \mathbf{I}_{kl},
    \label{eq:infer_softargmax}
\end{equation}
where $\bar{\mathbf{x}}^{B}_{q}\in \mathbb{R}^{2}$ is the predicted correspondence, $k$ and $l$ are integer spatial indices of $\phi(\widetilde{\mathbf{M}}^{A}_{q})$ and $\mathbf{I}\in \mathbb{Z}^{2\times h_{B}\times w_{B}}$ is an image coordinate grid with $\mathbf{I}_{kl} = [k, l]$. The prediction $\bar{\mathbf{x}}^{B}_{q}$ is interpreted as the expected value of the coordinate grid $\mathbf{I}$ under the probability distribution $\phi(\widetilde{\mathbf{M}}^{A}_{q})$. Note that $\bar{\mathbf{x}}^{B}_{q}$ is in the scale of $\mathbf{F}^{B}$ and it can be transformed to the scale of $I^{B}$ by multiplying with the image-to-feature spatial size ratio $H_{B}/h_{b}$.

\section{Experiments}

\begin{table*}
\centering
\caption{Comparison between previous methods and SimSC on PF-Pascal, PF-Willow and SPair-71K datasets. For PF-Pascal, methods are trained and evaluated on its training and testing splits. For PF-Willow, methods are trained on the training data of PF-Pascal and evaluated on the testing data of PF-Willow. For SPair-71K, we evaluate the models transferred from PF-Pascal and denote results as \textbf{SPair-71K (T)} and models fine-tuned on training split of SPair-71K as \textbf{SPair-71K (F)}. 
All methods in the table use ResNet101 as the feature backbone \xh{and trained by strongly supervised loss. We additionally include ``\textbf{\FN-iBOT*}'' which fine-tunes iBOT, a a more powerful ViT-based feature backbone using our method.} The top four results for each column are picked out and ranked by color in bold: {\color[HTML]{000000} \textbf{first}}, {\color[HTML]{343434} \textbf{second}}, {\color[HTML]{656565} \textbf{third}}, {\color[HTML]{9B9B9B} \textbf{fourth}}.}
\resizebox{\textwidth}{!}{
\begin{tabular}{llcccccccccc}
\hline
 & \multicolumn{1}{l|}{} & \multicolumn{3}{c|}{\textbf{PF-Pascal}} & \multicolumn{3}{c|}{\textbf{PF-Willow}} & \multicolumn{2}{c|}{\textbf{SPair-71K (T)}} & \multicolumn{2}{c}{\textbf{SPair-71K (F)}} \\
 & \multicolumn{1}{l|}{} & \multicolumn{3}{c|}{$\alpha$ with $\theta_{img}$} & \multicolumn{3}{c|}{$\alpha$ with $\theta_{kps}$} & \multicolumn{2}{c|}{$\alpha$ with $\theta_{bbox}$} & \multicolumn{2}{c}{$\alpha$ with $\theta_{bbox}$} \\
Methods & \multicolumn{1}{l|}{Eval. Reso.} & 0.05 & 0.1 & \multicolumn{1}{c|}{0.15} & 0.05 & 0.1 & \multicolumn{1}{c|}{0.15} & 0.05 & \multicolumn{1}{c|}{0.1} & 0.05 & 0.1 \\ \hline
ANC-Net \cite{ancnet} & \multicolumn{1}{l|}{256} & {\color[HTML]{9B9B9B} -} & {\color[HTML]{9B9B9B} 86.1} & \multicolumn{1}{c|}{{\color[HTML]{9B9B9B} -}} & {\color[HTML]{9B9B9B} -} & {\color[HTML]{9B9B9B} -} & \multicolumn{1}{c|}{{\color[HTML]{9B9B9B} -}} & {\color[HTML]{9B9B9B} -} & \multicolumn{1}{c|}{{\color[HTML]{9B9B9B} 28.7}} & {\color[HTML]{9B9B9B} -} & {\color[HTML]{9B9B9B} -} \\
HPF \cite{hpf} & \multicolumn{1}{l|}{max 300} & {\color[HTML]{9B9B9B} 60.1} & {\color[HTML]{9B9B9B} 84.8} & \multicolumn{1}{c|}{{\color[HTML]{9B9B9B} 92.7}} & {\color[HTML]{9B9B9B} 45.9} & {\color[HTML]{9B9B9B} 74.4} & \multicolumn{1}{c|}{{\color[HTML]{9B9B9B} 85.6}} & {\color[HTML]{9B9B9B} -} & \multicolumn{1}{c|}{{\color[HTML]{9B9B9B} -}} & {\color[HTML]{9B9B9B} -} & {\color[HTML]{9B9B9B} -} \\
SCOT \cite{scot} & \multicolumn{1}{l|}{max 300} & {\color[HTML]{9B9B9B} 63.1} & {\color[HTML]{9B9B9B} 85.4} & \multicolumn{1}{c|}{{\color[HTML]{9B9B9B} 92.7}} & {\color[HTML]{9B9B9B} \textbf{47.8}} & {\color[HTML]{656565} \textbf{76.0}} & \multicolumn{1}{c|}{{\color[HTML]{343434} \textbf{87.1}}} & {\color[HTML]{9B9B9B} -} & \multicolumn{1}{c|}{{\color[HTML]{9B9B9B} -}} & {\color[HTML]{9B9B9B} -} & {\color[HTML]{9B9B9B} -} \\
DHPF \cite{dhpf} & \multicolumn{1}{l|}{240} & {\color[HTML]{9B9B9B} 75.7} & {\color[HTML]{9B9B9B} 90.7} & \multicolumn{1}{c|}{{\color[HTML]{9B9B9B} 95.0}} & {\color[HTML]{9B9B9B} 41.4} & {\color[HTML]{9B9B9B} 67.4} & \multicolumn{1}{c|}{{\color[HTML]{9B9B9B} 81.8}} & {\color[HTML]{9B9B9B} 15.4} & \multicolumn{1}{c|}{{\color[HTML]{9B9B9B} 27.4}} & {\color[HTML]{9B9B9B} 20.7} & {\color[HTML]{9B9B9B} 37.3} \\
MMNet \cite{mmnet} & \multicolumn{1}{l|}{$224\times 320$} & {\color[HTML]{9B9B9B} 77.6} & {\color[HTML]{9B9B9B} 89.1} & \multicolumn{1}{c|}{{\color[HTML]{9B9B9B} 94.3}} & {\color[HTML]{9B9B9B} -} & {\color[HTML]{9B9B9B} -} & \multicolumn{1}{c|}{{\color[HTML]{9B9B9B} -}} & {\color[HTML]{9B9B9B} -} & \multicolumn{1}{c|}{{\color[HTML]{9B9B9B} -}} & {\color[HTML]{9B9B9B} -} & {\color[HTML]{9B9B9B} 40.9} \\
CHM \cite{chm} & \multicolumn{1}{l|}{240} & {\color[HTML]{9B9B9B} 80.1} & {\color[HTML]{9B9B9B} 91.6} & \multicolumn{1}{c|}{{\color[HTML]{9B9B9B} -}} & {\color[HTML]{9B9B9B} -} & {\color[HTML]{9B9B9B} 69.6} & \multicolumn{1}{c|}{{\color[HTML]{9B9B9B} -}} & {\color[HTML]{9B9B9B} -} & \multicolumn{1}{c|}{{\color[HTML]{9B9B9B} 30.1}} & {\color[HTML]{9B9B9B} -} & {\color[HTML]{9B9B9B} 46.3} \\
PMNC \cite{patchmatch} & \multicolumn{1}{l|}{400} & {\color[HTML]{9B9B9B} 82.4} & {\color[HTML]{9B9B9B} 90.6} & \multicolumn{1}{c|}{{\color[HTML]{9B9B9B} -}} & {\color[HTML]{9B9B9B} -} & {\color[HTML]{9B9B9B} -} & \multicolumn{1}{c|}{{\color[HTML]{9B9B9B} -}} & {\color[HTML]{9B9B9B} -} & \multicolumn{1}{c|}{{\color[HTML]{9B9B9B} 28.8}} & {\color[HTML]{9B9B9B} -} & {\color[HTML]{9B9B9B} 50.4} \\
CATs \cite{cats} & \multicolumn{1}{l|}{256} & {\color[HTML]{9B9B9B} 75.4} & {\color[HTML]{9B9B9B} 92.6} & \multicolumn{1}{c|}{{\color[HTML]{656565} \textbf{96.4}}} & {\color[HTML]{9B9B9B} 40.9} & {\color[HTML]{9B9B9B} 69.5} & \multicolumn{1}{c|}{{\color[HTML]{9B9B9B} 83.2}} & {\color[HTML]{9B9B9B} 13.6} & \multicolumn{1}{c|}{{\color[HTML]{9B9B9B} 27.0}} & {\color[HTML]{9B9B9B} 27.9} & {\color[HTML]{9B9B9B} 49.9} \\
TransforMatcher \cite{transformatcher} & \multicolumn{1}{l|}{240} & {\color[HTML]{9B9B9B} 80.8} & {\color[HTML]{9B9B9B} 91.8} & \multicolumn{1}{c|}{{\color[HTML]{9B9B9B} -}} & {\color[HTML]{9B9B9B} -} & {\color[HTML]{9B9B9B} 65.3} & \multicolumn{1}{c|}{{\color[HTML]{9B9B9B} -}} & {\color[HTML]{9B9B9B} -} & \multicolumn{1}{c|}{{\color[HTML]{9B9B9B} 30.1}} & {\color[HTML]{9B9B9B} 32.4} & {\color[HTML]{656565} \textbf{53.7}} \\
SemiMatch-CATs \cite{semimatch} & \multicolumn{1}{l|}{256} & {\color[HTML]{9B9B9B} 80.1} & {\color[HTML]{343434} \textbf{93.5}} & \multicolumn{1}{c|}{{\color[HTML]{343434} \textbf{96.6}}} & {\color[HTML]{9B9B9B} -} & {\color[HTML]{9B9B9B} -} & \multicolumn{1}{c|}{{\color[HTML]{9B9B9B} -}} & {\color[HTML]{9B9B9B} -} & \multicolumn{1}{c|}{{\color[HTML]{9B9B9B} -}} & {\color[HTML]{9B9B9B} -} & {\color[HTML]{9B9B9B} 50.1} \\
NCNet-C.E. Loss \cite{ncnet} & \multicolumn{1}{l|}{ori.} & {\color[HTML]{9B9B9B} 78.6} & {\color[HTML]{9B9B9B} 91.7} & \multicolumn{1}{c|}{{\color[HTML]{9B9B9B} 95.3}} & {\color[HTML]{9B9B9B} 43.0} & {\color[HTML]{9B9B9B} 70.9} & \multicolumn{1}{c|}{{\color[HTML]{9B9B9B} 83.9}} & {\color[HTML]{9B9B9B} 17.3} & \multicolumn{1}{c|}{{\color[HTML]{9B9B9B} 32.4}} & {\color[HTML]{9B9B9B} 29.1} & {\color[HTML]{9B9B9B} 50.7} \\
SFNet-C.E. Loss \cite{sfnet} & \multicolumn{1}{l|}{ori.} & {\color[HTML]{9B9B9B} 78.7} & {\color[HTML]{9B9B9B} \textbf{92.9}} & \multicolumn{1}{c|}{{\color[HTML]{9B9B9B} 96.0}} & {\color[HTML]{9B9B9B} 43.2} & {\color[HTML]{9B9B9B} 72.5} & \multicolumn{1}{c|}{{\color[HTML]{9B9B9B} 85.9}} & {\color[HTML]{9B9B9B} 13.3} & \multicolumn{1}{c|}{{\color[HTML]{9B9B9B} 27.9}} & {\color[HTML]{9B9B9B} 26.2} & {\color[HTML]{9B9B9B} 50.0} \\
PWarpC-CATs \cite{pwarpc} & \multicolumn{1}{l|}{ori.} & {\color[HTML]{9B9B9B} 79.8} & {\color[HTML]{9B9B9B} 92.6} & \multicolumn{1}{c|}{{\color[HTML]{656565} \textbf{96.4}}} & {\color[HTML]{343434} \textbf{48.1}} & {\color[HTML]{9B9B9B} \textbf{75.1}} & \multicolumn{1}{c|}{{\color[HTML]{9B9B9B} \textbf{86.6}}} & {\color[HTML]{9B9B9B} 15.4} & \multicolumn{1}{c|}{{\color[HTML]{9B9B9B} 27.9}} & {\color[HTML]{9B9B9B} 28.2} & {\color[HTML]{9B9B9B} 48.4} \\
PWarpC-NCNet \cite{pwarpc} & \multicolumn{1}{l|}{ori.} & {\color[HTML]{9B9B9B} 79.2} & {\color[HTML]{9B9B9B} 92.1} & \multicolumn{1}{c|}{{\color[HTML]{9B9B9B} 95.6}} & {\color[HTML]{656565} \textbf{48.0}} & {\color[HTML]{343434} \textbf{76.2}} & \multicolumn{1}{c|}{{\color[HTML]{656565} \textbf{86.8}}} & {\color[HTML]{656565} \textbf{21.5}} & \multicolumn{1}{c|}{{\color[HTML]{343434} \textbf{37.1}}} & {\color[HTML]{9B9B9B} \textbf{31.6}} & {\color[HTML]{9B9B9B} \textbf{52.0}} \\
PWarpC-DHPF \cite{pwarpc} & \multicolumn{1}{l|}{ori.} & {\color[HTML]{9B9B9B} 79.1} & {\color[HTML]{9B9B9B} 91.3} & \multicolumn{1}{c|}{{\color[HTML]{9B9B9B} 96.1}} & {\color[HTML]{000000} \textbf{48.5}} & {\color[HTML]{9B9B9B} 74.4} & \multicolumn{1}{c|}{{\color[HTML]{9B9B9B} 85.4}} & {\color[HTML]{9B9B9B} 16.4} & \multicolumn{1}{c|}{{\color[HTML]{9B9B9B} 28.6}} & {\color[HTML]{9B9B9B} 23.3} & {\color[HTML]{9B9B9B} 38.7} \\
PWarpC-SFNet \cite{pwarpc} & \multicolumn{1}{l|}{ori.} & {\color[HTML]{9B9B9B} 78.3} & {\color[HTML]{9B9B9B} 92.2} & \multicolumn{1}{c|}{{\color[HTML]{9B9B9B} \textbf{96.2}}} & {\color[HTML]{9B9B9B} 47.5} & {\color[HTML]{000000} \textbf{77.7}} & \multicolumn{1}{c|}{{\color[HTML]{000000} \textbf{88.8}}} & {\color[HTML]{9B9B9B} 17.3} & \multicolumn{1}{c|}{{\color[HTML]{9B9B9B} 32.5}} & {\color[HTML]{9B9B9B} 27.0} & {\color[HTML]{9B9B9B} 48.8} \\
SCorrSAN \cite{scorrsan} & \multicolumn{1}{l|}{256} & {\color[HTML]{9B9B9B} 81.5} & {\color[HTML]{656565} \textbf{93.3}} & \multicolumn{1}{c|}{{\color[HTML]{343434} \textbf{96.6}}} & {\color[HTML]{9B9B9B} -} & {\color[HTML]{9B9B9B} -} & \multicolumn{1}{c|}{{\color[HTML]{9B9B9B} -}} & {\color[HTML]{9B9B9B} -} & \multicolumn{1}{c|}{{\color[HTML]{9B9B9B} -}} & {\color[HTML]{9B9B9B} -} & {\color[HTML]{343434} \textbf{55.3}} \\ \hdashline
\textbf{\FN-ResNet101} & \multicolumn{1}{l|}{256} & {\color[HTML]{9B9B9B} \textbf{84.4}} & {\color[HTML]{9B9B9B} 92.3} & \multicolumn{1}{c|}{{\color[HTML]{9B9B9B} 95.6}} & {\color[HTML]{9B9B9B} 37.1} & {\color[HTML]{9B9B9B} 60.6} & \multicolumn{1}{c|}{{\color[HTML]{9B9B9B} 73.3}} & {\color[HTML]{9B9B9B} 18.1} & \multicolumn{1}{c|}{{\color[HTML]{9B9B9B} 33.5}} & {\color[HTML]{9B9B9B} 30.9} & {\color[HTML]{9B9B9B} 51.0} \\
\textbf{\FN-ResNet101} & \multicolumn{1}{l|}{ori.} & {\color[HTML]{9B9B9B} 82.2} & {\color[HTML]{9B9B9B} 92.7} & \multicolumn{1}{c|}{{\color[HTML]{9B9B9B} 95.2}} & {\color[HTML]{9B9B9B} 36.1} & {\color[HTML]{9B9B9B} 63.5} & \multicolumn{1}{c|}{{\color[HTML]{9B9B9B} 76.6}} & {\color[HTML]{000000} \textbf{24.5}} & \multicolumn{1}{c|}{{\color[HTML]{656565} \textbf{36.8}}} & {\color[HTML]{656565} \textbf{35.4}} & {\color[HTML]{9B9B9B} 51.0} \\
\textbf{\FN-ResNet101*} & \multicolumn{1}{l|}{256} & {\color[HTML]{343434} \textbf{86.2}} & {\color[HTML]{9B9B9B} 92.6} & \multicolumn{1}{c|}{{\color[HTML]{9B9B9B} 95.3}} & {\color[HTML]{9B9B9B} 42.5} & {\color[HTML]{9B9B9B} 65.4} & \multicolumn{1}{c|}{{\color[HTML]{9B9B9B} 78.1}} & {\color[HTML]{9B9B9B} \textbf{18.2}} & \multicolumn{1}{c|}{{\color[HTML]{9B9B9B} 31.4}} & {\color[HTML]{9B9B9B} 31.3} & {\color[HTML]{9B9B9B} 51.6} \\
\textbf{\FN-ResNet101*} & \multicolumn{1}{l|}{ori.} & {\color[HTML]{656565} \textbf{85.3}} & {\color[HTML]{656565} \textbf{93.3}} & \multicolumn{1}{c|}{{\color[HTML]{9B9B9B} 96.0}} & {\color[HTML]{9B9B9B} 40.7} & {\color[HTML]{9B9B9B} 68.0} & \multicolumn{1}{c|}{{\color[HTML]{9B9B9B} 80.6}} & {\color[HTML]{000000} \textbf{24.5}} & \multicolumn{1}{c|}{{\color[HTML]{9B9B9B} \textbf{36.0}}} & {\color[HTML]{343434} \textbf{36.6}} & {\color[HTML]{9B9B9B} 51.5} \\ \hline
\textbf{\FN-iBOT*} & \multicolumn{1}{l|}{256} & {\color[HTML]{000000} \textbf{88.4}} & {\color[HTML]{000000} \textbf{95.6}} & \multicolumn{1}{c|}{{\color[HTML]{000000} \textbf{97.3}}} & {\color[HTML]{9B9B9B} 44.9} & {\color[HTML]{9B9B9B} 71.4} & \multicolumn{1}{c|}{{\color[HTML]{9B9B9B} 84.5}} & {\color[HTML]{343434} \textbf{22.0}} & \multicolumn{1}{c|}{{\color[HTML]{000000} \textbf{37.9}}} & {\color[HTML]{000000} \textbf{43.0}} & {\color[HTML]{000000} \textbf{63.5}} \\ \hline
\end{tabular}
}
\label{tab:result}
\end{table*}

We evaluate our method on three public datasets: PF-Pascal \cite{pf}, PF-Willow \cite{pfwillow}, and SPair-71K \cite{spair}. In \cref{dataset}, we briefly introduce three public datasets and the evaluation metric. In \cref{implementation}, we elaborate on our experimental setup, which includes configurations of models and hyperparameters. We show the comparison between our method and state-of-the-art methods in  \cref{result} and provide model analysis in  \cref{analysis}.

\subsection{Datasets and evaluation metric}
\label{dataset}
\paragraph{Datasets} PF-Pascal consists of $2941$ training image pairs, $308$ validation pairs, and $299$ testing pairs spanning across $20$ categories of objects. Its supplement PF-Willow contains $900$ testing pairs split into $3$ categories with no training and validation data. SPair-71K, on the other hand, is a larger and more challenging dataset with $53,340$ training pairs, $5,384$ validation pairs, and $12,234$ testing pairs across $18$ categories of objects with large scale and appearance variation. For all three datasets, each image pair has non-uniform numbers of ground-truth correspondences. SPair-71K additionally provides the object's bounding box for each image.

\paragraph{Evaluation Metric} We follow the common practice in the literature and use the percentage of corrected keypoints (PCK) as the evaluation metric. Given an image pair $(I^{A}, I^{B})$ and its associated correspondence set $\mathbf{X} = \{(\mathbf{x}^{A}_{q}, \mathbf{x}^{B}_{q})\ |\ q=1,2,...,n\}$, for each $\mathbf{x}^{A}_{q}=(x^{A}_{q}, y^{A}_{q})$, we find its predicted correspondence $\bar{\mathbf{x}}^{B}_{q}$ and calculate PCK for the image pair by:
\begin{equation}
    PCK(I^{A}, I^{B}) = \frac{1}{n}\sum^{n}_{q}\mathbb{I}(\|\bar{\mathbf{x}}^{B}_{q}-\mathbf{x}^{B}_{q}\| \leq \alpha * \theta)
\end{equation}
where $\theta$ is the base threshold, $\alpha$ is a number less than $1$ and $\mathbb{I}(\cdot)$ is the binary indicator function with $\mathbb{I}(\textrm{\texttt{true}})=1$ and $\mathbb{I}(\textrm{\texttt{false}})=0$. For PF-Pascal, $\theta$ is set as $\theta_{img} = \max(H_{B}, W_{B})$. For PF-Willow, base threshold is set as $\theta_{kps}=\max(\max_{q}(x_{q}^{B})-\min
_{q}(x_{q}^{B}), \max_{q}(y_{q}^{B})-\min_{q}(y_{q}^{B}))$ and for SPair-71K, base threshold is set as $\theta_{bbox} = \max(h_{bbox}, w_{bbox})$ where $h_{bbox}$ and $w_{bbox}$ are height and width of the bounding box. The choices of all three base thresholds align with the convention in the literature.

\subsection{Experimental details}
\label{implementation}
\paragraph{Feature backbone}
We implement our framework in Python using the PyTorch \cite{pytorch} library. We evaluate our fine-tuning method on ImageNet pre-trained ResNet101 \cite{resnet}, which has the CNN architecture and is widely used in literature. Specifically, ResNet101 is truncated before \texttt{layer4}, so the feature dimension is $1024$, and the image-to-feature spatial size ratio is $16$. In addition, we also validate our method on ResNet50, DINO \cite{dino}, and iBOT \cite{ibot} to show that our method works for both CNN-based and ViT-based \cite{vit} feature backbones. 
Note that ResNet101 and ResNet50 are pretrained on ImageNet under a fully supervised setting, while DINO and iBOT are pretrained on ImageNet with self-supervised learning. 
We provide more details for this part in \cref{analysis}.

\paragraph{Training configuration} 
We train two configurations: fine-tuning the \textit{last block} and fine-tuning the \textit{entire feature backbone}, which we denote as \textbf{\FN-ResNet101} and \textbf{\FN-ResNet101*} respectively in \cref{tab:result}. \xh{\FN-ResNet101 corresponds to works that only train their matching heads and \FN-ResNet101* corresponds to works that train both matching heads and backbones.} We use the Adam optimizer \cite{adam} for training, and the learning rate is set to $1\times 10^{-4}$ for fine-tuning the last block and to $2\times 10^{-5}$ for fine-tuning the entire feature backbone. The learning rate for the temperature learning module $g(\cdot)$ is set to $1\times 10^{-4}$ for both configurations. We stop the gradient of $g(\cdot)$ from propagating to the feature backbone to prevent interference between the two modules. We fine-tune feature backbones for $100$ epochs for PF-Pascal dataset and $10$ epochs for SPair-71K dataset. Values for $n_{s}$ and $n_{k}$ are set to $3$ and $5$ respectively. The standard deviation $\sigma$ for $\mathbf{G}$ in \cref{eq:infer_kernel} is set to $7$. $\beta_{thres}$ and $\gamma$ in \cref{eq:loss} are $0.1$ and $0.2$ respectively. Images are resized to $256\times 256$ for all datasets and feature backbones during training. We train all models on $3$ Nvidia GTX 1080Ti GPUs. 

\subsection{Results}
\label{result}

We summarize the comparison between existing methods and our method in \cref{tab:result}. \xh{\emph{All} listed methods are trained with strongly supervised loss. However, SFNet and NCNet are originally trained with weakly supervised losses. They are re-trained with the keypoint cross-entropy loss, denoted as SFNet-C.E. Loss and NCNet-C.E. Loss, for a fair comparison.}
\paragraph{Comparison with existing methods}
Following the convention in the literature, we report four results: fine-tuned result on PF-Pascal (\textbf{PF-Pascal}), transferred result from PF-Pascal to PF-Willow (\textbf{PF-Willow}), transferred result from PF-Pascal to SPair-71K (\textbf{SPair-71K (T)}), and fine-tuned result on SPair-71K (\textbf{SPair-71K (F)}). The majority of the existing methods are evaluated on either an image size of $256\times 256$ or the original image size, so we evaluate \textbf{\FN-ResNet101} and \textbf{\FN-ResNet101*} on both sizes. Under the same feature backbone, we achieve accuracy on par with state-of-the-art methods despite the remarkable simplicity of our method. We achieve the best results at threshold $\alpha=0.05$ for PF-Pascal, SPair-71K (T) and SPair-71K (F). We note that our transferred result from PF-Pascal to PF-Willow is not as significant as to SPair-71K. At the same time, we also notice that methods that transfer strongly from PF-Pascal to PF-Willow tend to transfer deficiently from PF-Pascal to SPair-71K. We suspect that this is because the data distributions between PF-Willow and SPair-71K are different and represent two directions of generalization. Moving toward one means moving away from the other. 

We also include the results of fine-tuning a more powerful backbone named iBOT \cite{ibot}, which has the ViT-B/16 architecture, denoted as \textbf{SimSC-iBOT*}. As shown, \xh{it outperforms other methods on PF-Pascal, SPair-71K (T) and SPair-71K (F) at all thresholds by a notable margin.}

\paragraph{Comparison within SimSC}
In \cref{tab:result}, we also compare the effects of fine-tuning the entire feature backbone and fine-tuning only the last block. We can see that overall fine-tuning the entire feature backbone shows more promise than fine-tuning only the last block. We also notice that performance using the original images is better than using squared images. We believe this is attributed to the fact that objects are not distorted if using original images.

\subsection{Analysis}
\label{analysis}
We include several additional analyses of our method in this section. All evaluation is done with an image size of $256\times 256$. A model with the symbol ``$\ast$'' indicates fine-tuning the entire feature backbone, otherwise fine-tuning only the last block. \xh{More experiments can be found in the supplementary materials.} 

\subsubsection{Comparison with the finetuning baselines.}
\label{exp:imgnet_temp1_ours}
\begin{figure}[t]
    \centering
    \includegraphics[width=.95\linewidth]{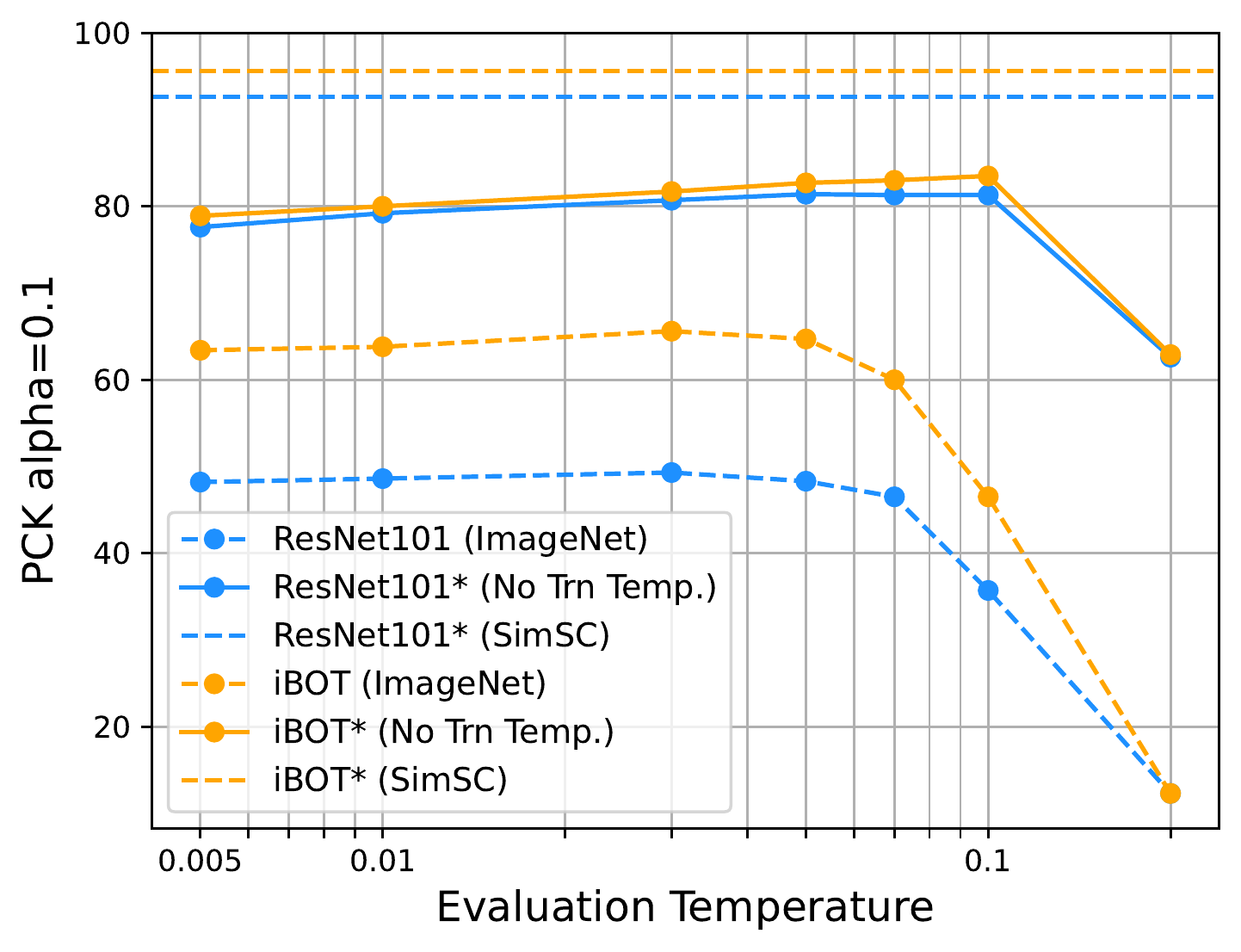}
    \caption{Comparison between ImageNet pre-trained models, fine-tuned models with no training temperature ($\beta_{trn} = 1$), and \FN on PF-Pascal dataset.}
    \label{fig:comparison_imgnet_temp1_ours}
\end{figure}

We compare the performance of ImageNet pre-trained feature backbones, fine-tuned feature backbones with no training temperature ($\beta_{trn}=1$), and fine-tuned feature backbones using SimSC. \xh{The accuracy of kernel soft-argmax is significantly affect by $\beta_{eval}$. Therefore, we evaluate ImageNet pre-trained models and fine-tuned models with no training temperature using different $\beta_{eval}$ on the PF-Pascal dataset and plot the results in \cref{fig:comparison_imgnet_temp1_ours}.}
Our method improves the results of fine-tuning both ResNet101* and iBOT* by $11.2$ percentage points and $12$ percentage points respectively when compared with fine-tuning pipeline with training temperature $\beta_{trn}=1$.
The performance of ImageNet pre-trained models and fine-tuned models with $\beta_{trn}=1$ drops significantly when $\beta_{eval} > 0.1$. This is because $\beta_{eval}$ is too big to suppress incorrect candidates and they keep too much weight in the marginalization in \cref{eq:infer_softargmax}. 

\subsubsection{Effectiveness on different feature backbones}
\label{exp:diff_backbones}

We test our method on four feature backbones: ResNet50 \cite{resnet}, ResNet101 \cite{resnet}, DINO \cite{dino} and iBOT \cite{ibot}. ResNet50 and ResNet101 are CNN-based, and DINO and iBOT have the ViT-B/8 and ViT-B/16 architecture respectively. Following the setup in \cref{implementation}, we fine-tune both the last block and the entire module for each type of feature backbone, except for DINO, where we only fine-tune the last block. This is because the image-to-feature spatial size ratio of DINO is $8$ rather than $16$ (the latter one is the ratio of all other feature backbones). The memory cost would be too large to fit into our GPU if we fine-tune the entire DINO under the same image size. We summarize the performance of each type of feature backbone in \cref{tab:exp:ours_diff_backbone}. All evaluation metrics are the same as described in \cref{dataset}.

As shown in \cref{tab:exp:ours_diff_backbone}, our method works well on all four feature backbones. iBOT appears to be the most effective one due to its raw features' superiority and transformer architecture. DINO takes the second position in PF-Pascal and SPair-71K (F), followed by ResNet101 and ResNet50. 

\begin{table}[t]
\centering
\caption{Performance of \FN using different feature backbones.}
\resizebox{.9\linewidth}{!}{
\begin{tabular}{l|c|c|c|c}
\hline
 & \textbf{PF-Pas.} & \textbf{PF-Wil.} & \textbf{S.P. (T)} & \textbf{S.P. (F)} \\
Feat. Ext. & $\alpha=0.1$ & $\alpha=0.1$ & $\alpha=0.1$ & $\alpha=0.1$ \\ \hline
\textbf{Res101} & {\color[HTML]{9B9B9B} 92.3} & {\color[HTML]{9B9B9B} 60.6} & {\color[HTML]{656565} \textbf{33.5}} & {\color[HTML]{9B9B9B} 51.0} \\
\textbf{Res101*} & {\color[HTML]{9B9B9B} \textbf{92.6}} & {\color[HTML]{656565} \textbf{65.4}} & {\color[HTML]{9B9B9B} \textbf{31.4}} & {\color[HTML]{9B9B9B} \textbf{51.6}} \\
\textbf{Res50} & {\color[HTML]{9B9B9B} 90.4} & {\color[HTML]{9B9B9B} 59.7} & {\color[HTML]{9B9B9B} 31.3} & {\color[HTML]{9B9B9B} 47.0} \\
\textbf{Res50*} & {\color[HTML]{9B9B9B} 91.3} & {\color[HTML]{9B9B9B} \textbf{62.5}} & {\color[HTML]{9B9B9B} 28.5} & {\color[HTML]{9B9B9B} 45.7} \\
\textbf{DINO} & {\color[HTML]{656565} \textbf{92.8}} & {\color[HTML]{9B9B9B} 62.2} & {\color[HTML]{9B9B9B} 28.3} & {\color[HTML]{656565} \textbf{52.6}} \\
\textbf{iBOT} & {\color[HTML]{343434} \textbf{94.9}} & {\color[HTML]{343434} \textbf{65.4}} & {\color[HTML]{343434} \textbf{36.0}} & {\color[HTML]{343434} \textbf{61.8}} \\
\textbf{iBOT*} & {\color[HTML]{000000} \textbf{95.6}} & {\color[HTML]{000000} \textbf{71.4}} & {\color[HTML]{000000} \textbf{37.9}} & {\color[HTML]{000000} \textbf{63.5}} \\ \hline
\end{tabular}}
\label{tab:exp:ours_diff_backbone}
\end{table}

\subsubsection{Removing L2 norm of feature maps}
\label{exp:removel2norm}

\begin{table}[t]
\centering
\caption{Effects of L2 normalization to finetuning ResNet101* on PF-Pascal dataset.}
\resizebox{.9\linewidth}{!}{
\begin{tabular}{c|c|c}
\hline
 & \textbf{PF-Pascal} & \textbf{SPair-71K (T)} \\
Config. & $\alpha$ = 0.05 / 0.1 / 0.15 & $\alpha$ = 0.05 / 0.1 \\ \hline 
WithL2Norm & 57.8 / 80.3 / 87.9 & 6.6 / 16.4 \\ \hline
NoL2Norm & 81.6 / 92.3 / 94.9 & 13.7 / 28.0 \\ \hline
SimSC & \textbf{86.2} / \textbf{92.6} / \textbf{95.3} & \textbf{18.2} / \textbf{31.4}\\ \hline
\end{tabular}}
\label{tab:l2norm}
\vspace{-1mm}
\end{table}

\xh{We remove L2 norm on feature maps and finetune ResNet101* with $\beta_{trn}$=1, denoted as ``NoL2Norm'', on the PF-Pascal dataset. As shown in \cref{tab:l2norm}, removing L2 norm can substantially improve the performance of finetuning from the baseline ``WithL2Norm''. This verifies our analysis of the L2 norm in \cref{sec:method:templearning}. The temperature acts as a re-scaling factor on the overly smooth correlation distribution caused by L2 Norm and allows the gradient to be better propagated to the network. However, compared with SimSC, removing L2 norm degrades the kernel softmax correspondence localizer to picking the nearest neighbour due to an excessively large max value, leading to an inferior result at a small threshold of $\alpha=0.05$. The generalizability to the SPair-71K dataset is also weaker than that of SimSC. We additionally include the analysis of the training gradient the supplementary materials.}


\section{Conclusion}

We propose a remarkably simple framework that tackles the problem of semantic matching. \xh{We discover that when fine-tuning ImageNet pre-trained backbone on the semantic matching task, the L2 normalization to the feature map produces an over-smooth matching distribution and significantly hinders the finetuning process. By setting an appropriate temperature to the softmax, the over-smoothness can be alleviated and quality of features can be substantially improved. In order to avoid manually tuning the temperature, we propose a temperature learning module that is jointly optimized with feature extractors. We evaluate our method on three public datasets and achieve accuracy on par with state-of-the-art methods. By coupling with iBOT, a more powerful self-supervised feature extractor, we outperform existing methods on the PF-Pascal and SPair-71K datasets by a large margin. Unlike existing methods, our method does not use any learned matching head, sophisticated training loss or training procedures. }

\paragraph{Acknowledgements} 
This work is partially supported by Hong Kong Research
Grant Council - Early Career Scheme (Grant No. 27208022) and HKU Seed Fund for Basic Research.


{\small
\bibliographystyle{ieee_fullname}
\bibliography{egbib}
}

\end{document}


\title{SimSC: A Simple Framework for Semantic Correspondence \\ with Temperature Learning\\
-- \textit{Supplementary Material} --}

\author{Xinghui Li\textsuperscript{\textdagger}  \qquad
Kai Han$\textsuperscript{\textdaggerdbl}\footnotemark[1]$
\qquad 
Xingchen Wan\textsuperscript{\textdagger} \qquad 
Victor Adrian Prisacariu\textsuperscript{\textdagger}\\[0.3em]
\textsuperscript{\textdagger}University of Oxford \qquad 
\textsuperscript{\textdaggerdbl}The University of Hong Kong\\
{\tt\small \{xinghui,xwan,victor\}@robot.ox.ac.uk \qquad kaihanx@hku.hk}
}

\maketitle

\renewcommand{\thefootnote}{\fnsymbol{footnote}}
\footnotetext[1]{Corresponding author.}

\section{Learning curve of temperature} 
\begin{figure}[ht]
    \centering
    \begin{subfigure}{.5\linewidth}
        \centering
        \includegraphics[width=\linewidth]{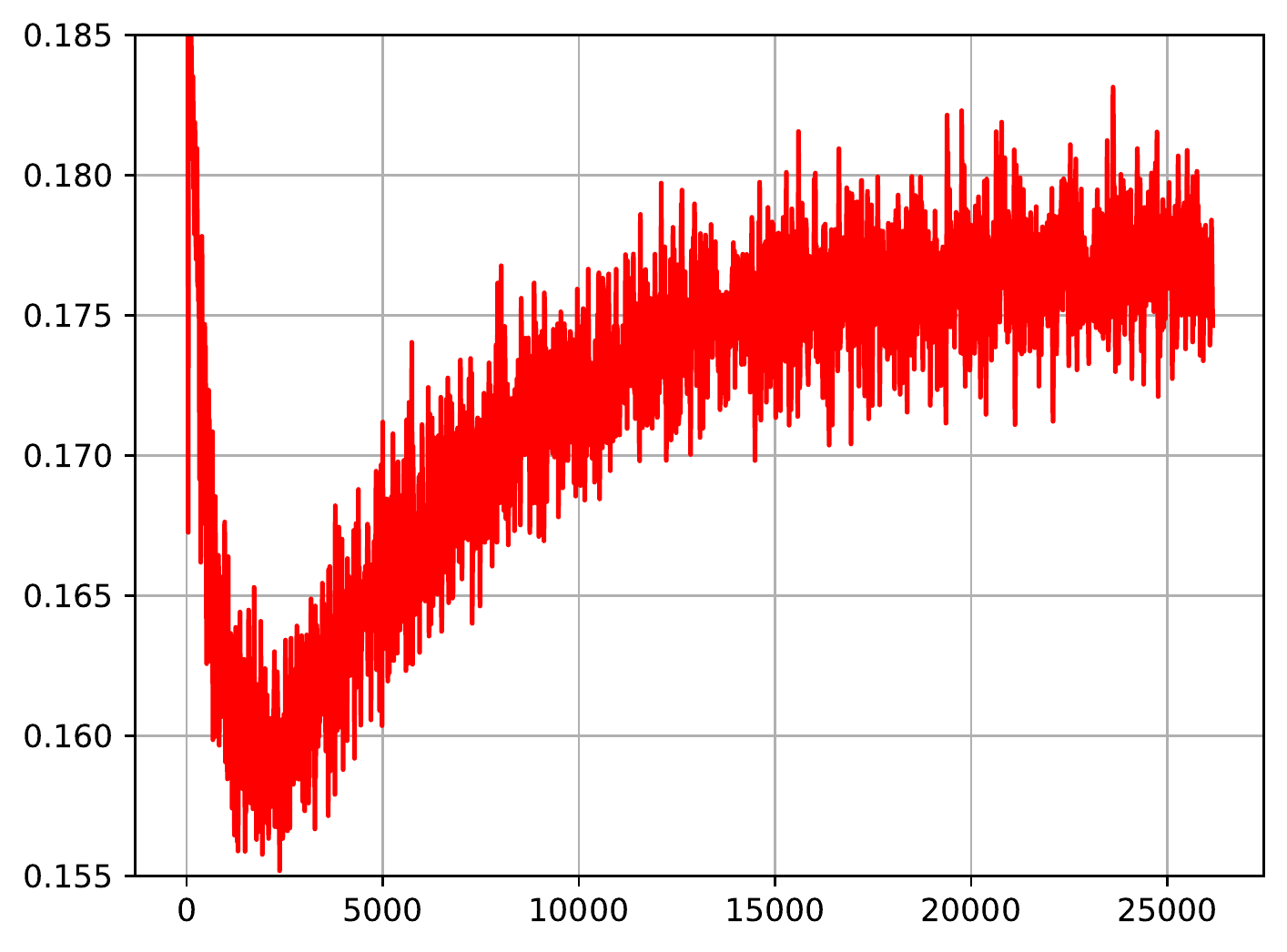}
        \caption{$\beta_{A}$}
    \end{subfigure}%
    \begin{subfigure}{.5\linewidth}
        \centering
        \includegraphics[width=\linewidth]{pdf/supp_show_temp/src_avg_temp.pdf}
        \caption{$\beta_{B}$}
    \end{subfigure}
    \begin{subfigure}{.5\linewidth}
        \centering
        \includegraphics[width=\linewidth]{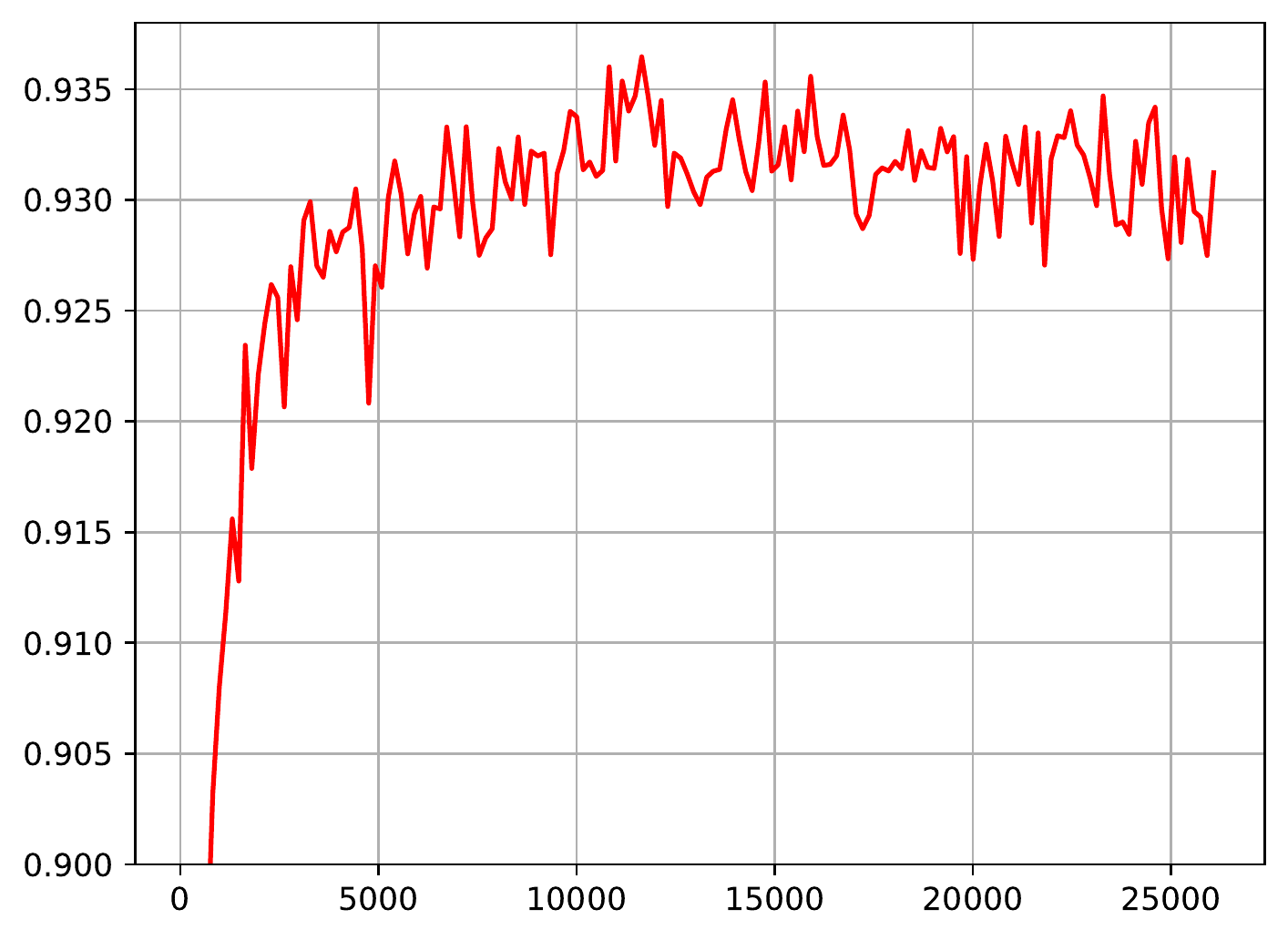}
        \caption{Validation Accuracy}
    \end{subfigure}%
    \begin{subfigure}{.5\linewidth}
        \centering
        \includegraphics[width=\linewidth]{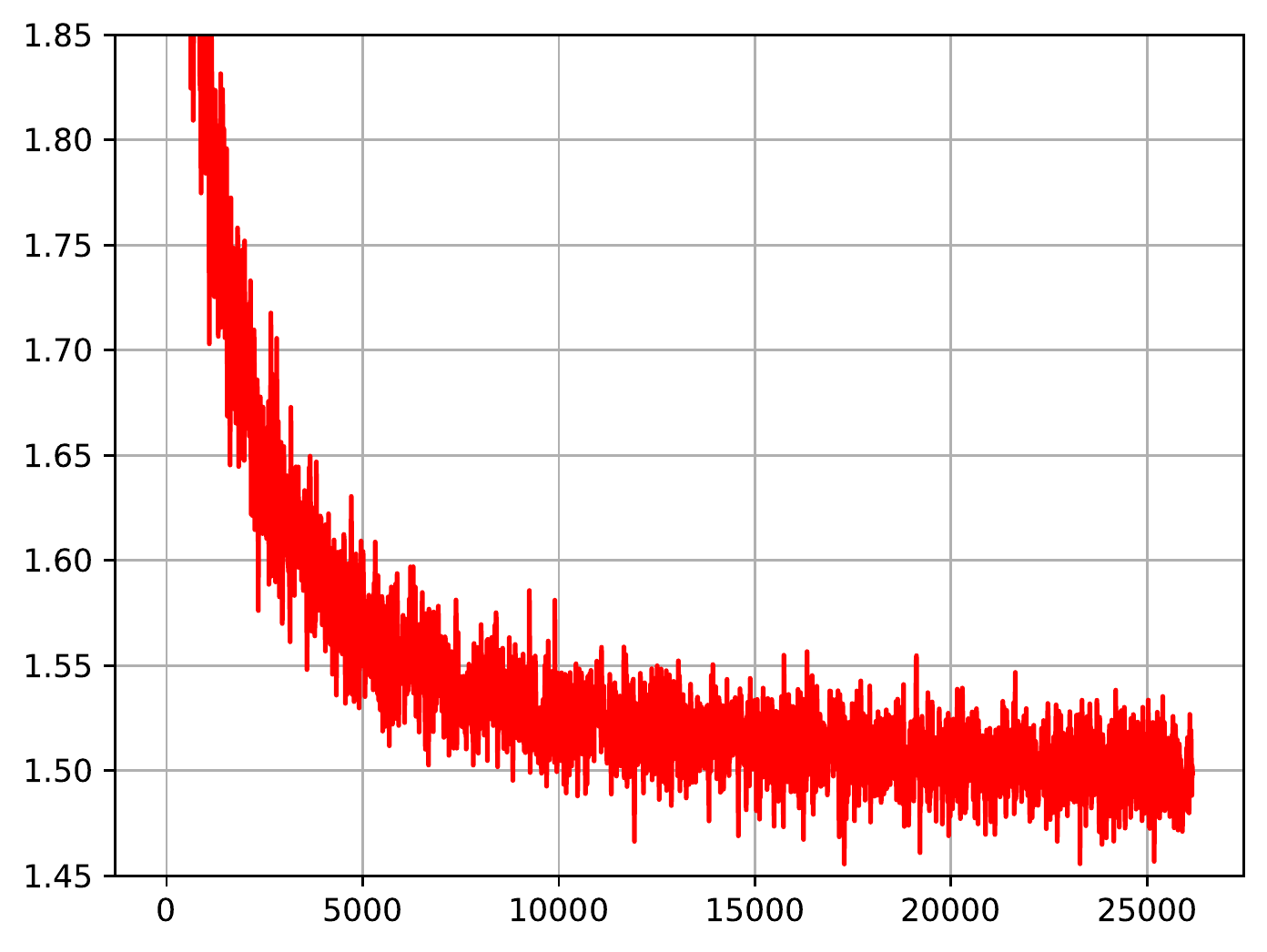}
        \caption{Loss}
    \end{subfigure}
    \caption{Learning curves of $\beta_{A}$, $\beta_{B}$, validation accuracy and loss of SimSC-ResNet101* on the PF-Pascal dataset. \xh{We train the model for 200 epochs in order to provide the entire trend of the curve.}}
    \label{fig:learning_curve_temp}
\end{figure}

We provide the plots of temperatures, validation accuracy and loss of ResNet101*-SimSC on PF-Pascal in \cref{fig:learning_curve_temp} to show how the temperature and backbone are jointly updated. As illustrated, $\beta_{A}$ and $\beta_{B}$ have very similar learning curves. From 0-2k step, the temperature and loss drop rapidly and accuracy is improved, due to the matching distribution drawn closer to the target by both reducing temperature and learning better features. From 2k-12k step, the temperature increases while accuracy keeps improving. This shows that the temperature gradually changes to its optimal position and helps tuning of the backbone. The accuracy starts dropping after 12k step due to overfitting and the temperature is stable after 20k step.

\section{Gradient analysis}

\begin{figure}[t]
    \centering
    \begin{subfigure}{.5\linewidth}
        \centering
        \includegraphics[width=\linewidth]{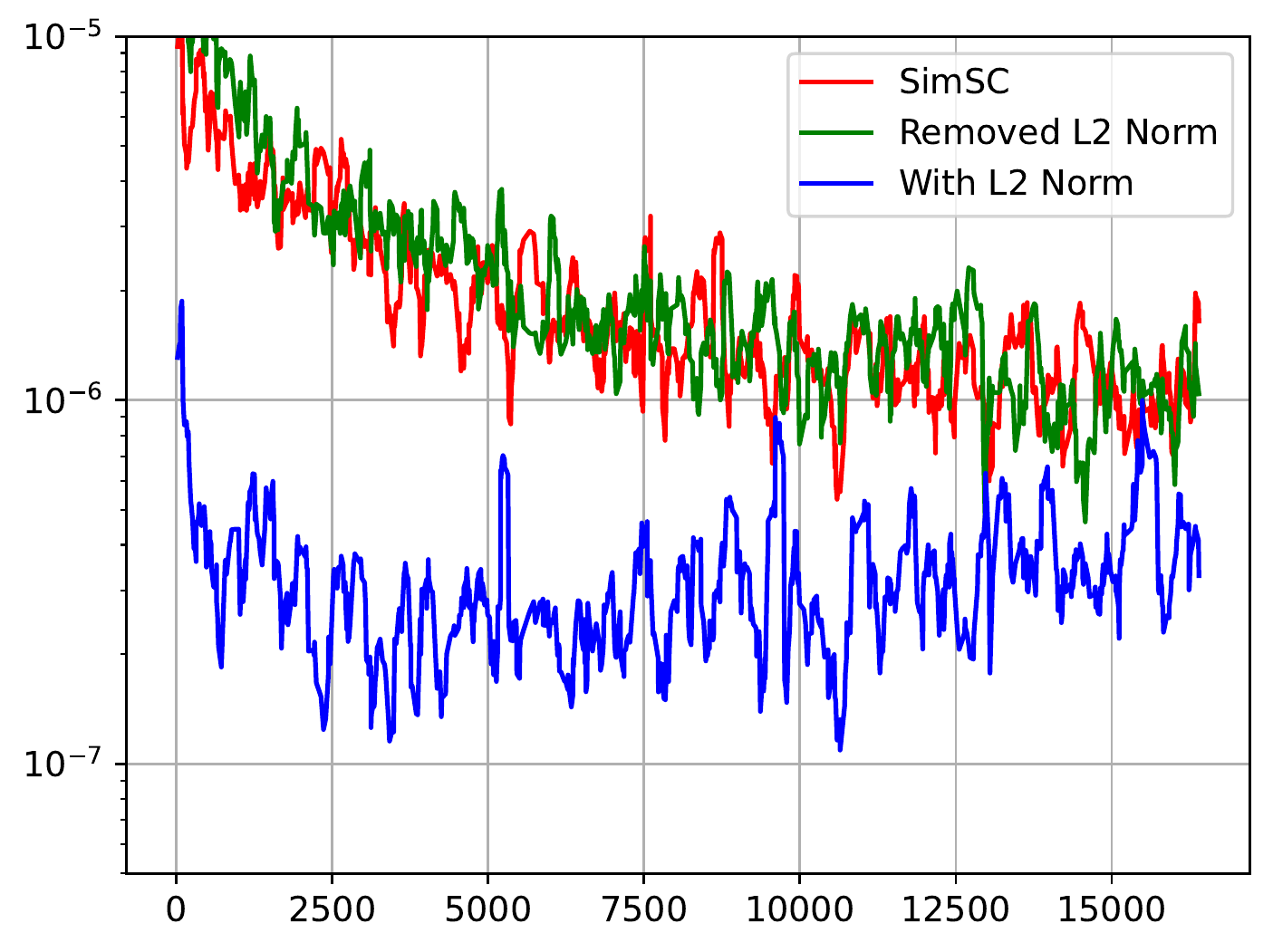}
        \caption{Gradient of \texttt{conv\_1}}
    \end{subfigure}%
    \begin{subfigure}{.5\linewidth}
        \centering
        \includegraphics[width=\linewidth]{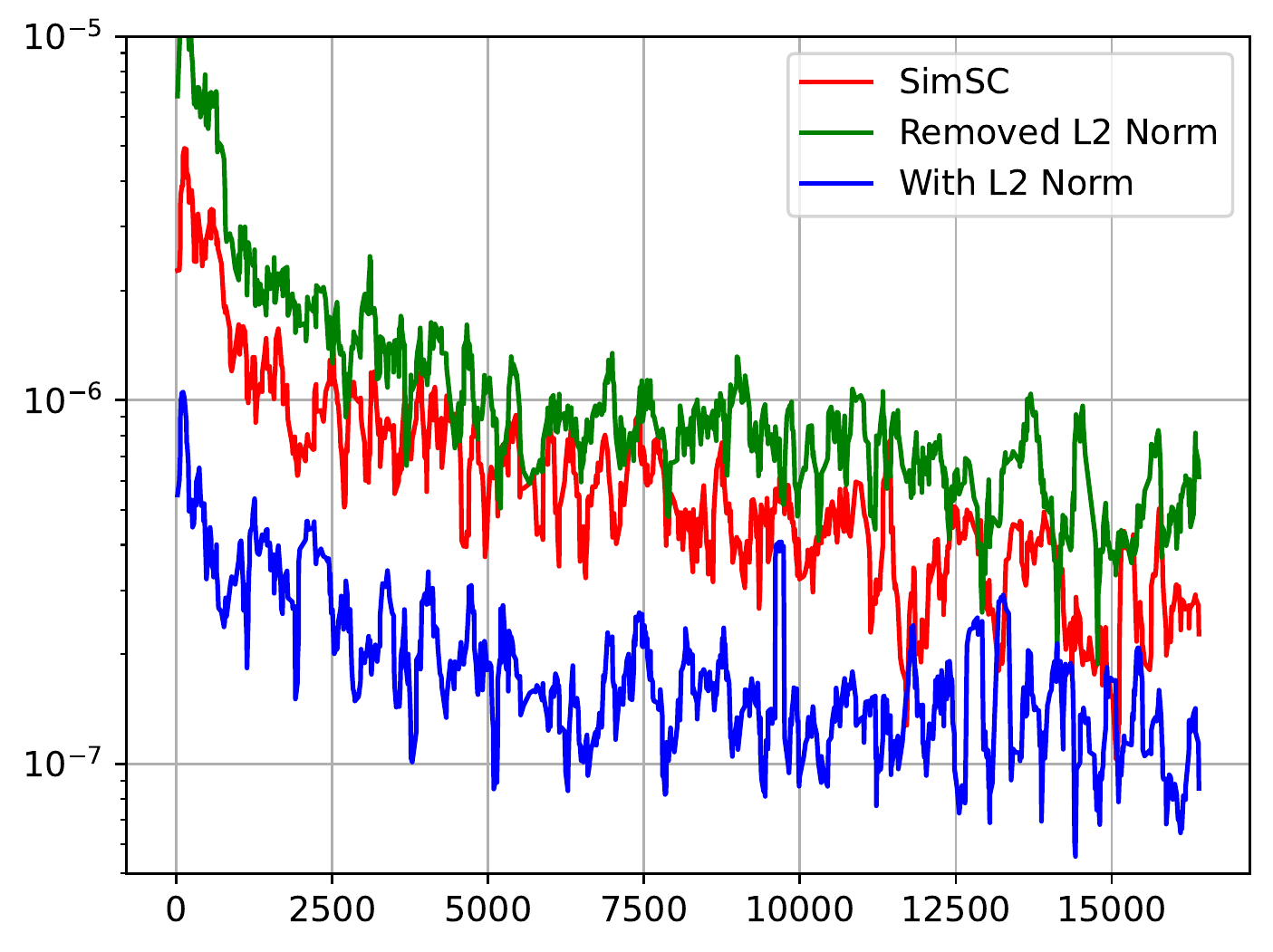}
        \caption{Gradient of \texttt{conv\_2}}
    \end{subfigure}
    \begin{subfigure}{.5\linewidth}
        \centering
        \includegraphics[width=\linewidth]{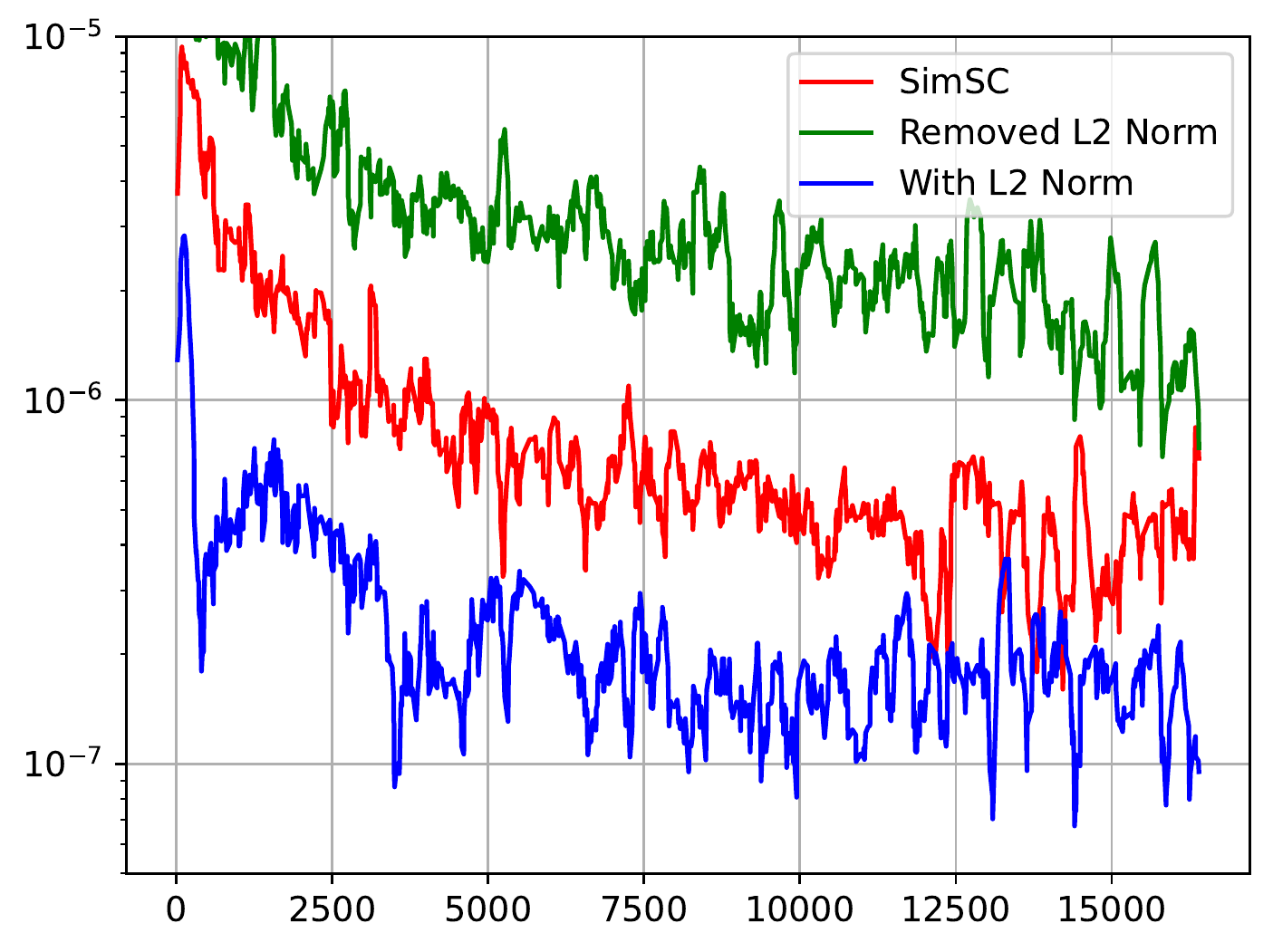}
        \caption{Gradient of \texttt{conv\_3}}
    \end{subfigure}
    \caption{The comparison between absolute mean values of gradient backpropagated to \texttt{conv\_1}, \texttt{conv\_2}, \texttt{conv\_3} of the last bottleneck block when training ``SimSC-ResNet101*'', ``NoL2Norm-ResNet101*'' and ``WithL2Norm-ResNet101*''. To avoid overly noisy plot, the gradient plots have been smoothed.}
    \label{fig:investigate_grad}
\end{figure}

\xh{In order to demonstrate the effect of the L2 normalization on the gradient backpropagated to the network when $\beta_{trn}=1$, we train three configurations of ResNet101*: (1) With L2 normalization, (2) Without L2 normalization, (3) SimSC, and track the gradients of \texttt{conv\_1}, \texttt{conv\_2}, \texttt{conv\_3} of the last bottleneck block of them. As illustrated in \cref{fig:investigate_grad}, the absolute mean values of gradients of ``NoL2Norm'' and ``SimSC'' are significantly and consistently larger than that of ``WithL2Norm'' during training by one order of magnitude. This agrees with our analysis of the impact of L2 normalization on the training of the model described in Sec. 3.2 of the main manuscript.}

\section{Manual temperature values for training}
\begin{figure}[t]
    \centering
    \begin{subfigure}{.5\linewidth}
        \centering
        \includegraphics[width=\linewidth]{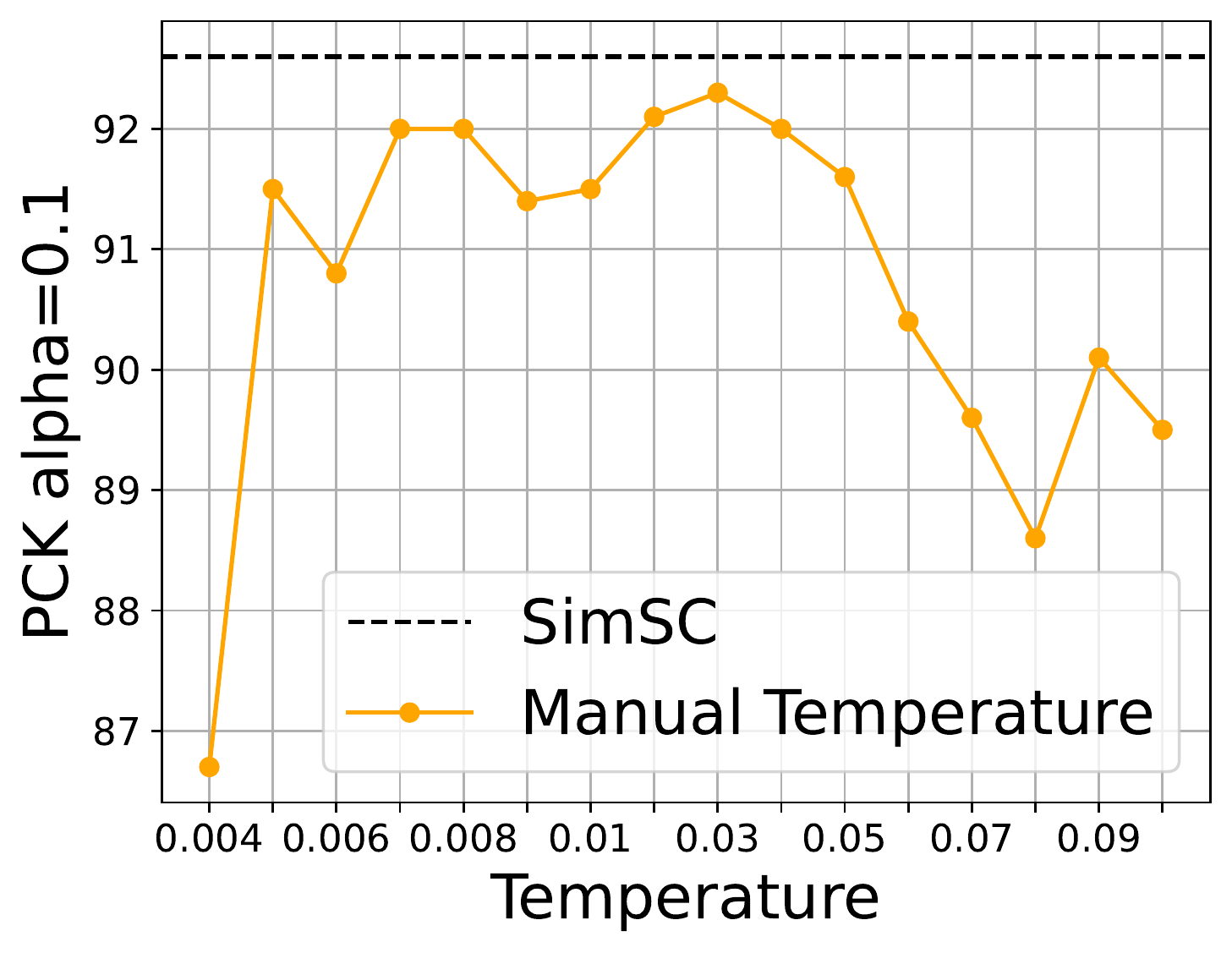}
        \caption{PF-Pascal}
    \end{subfigure}%
    \begin{subfigure}{.5\linewidth}
        \centering
        \includegraphics[width=\linewidth]{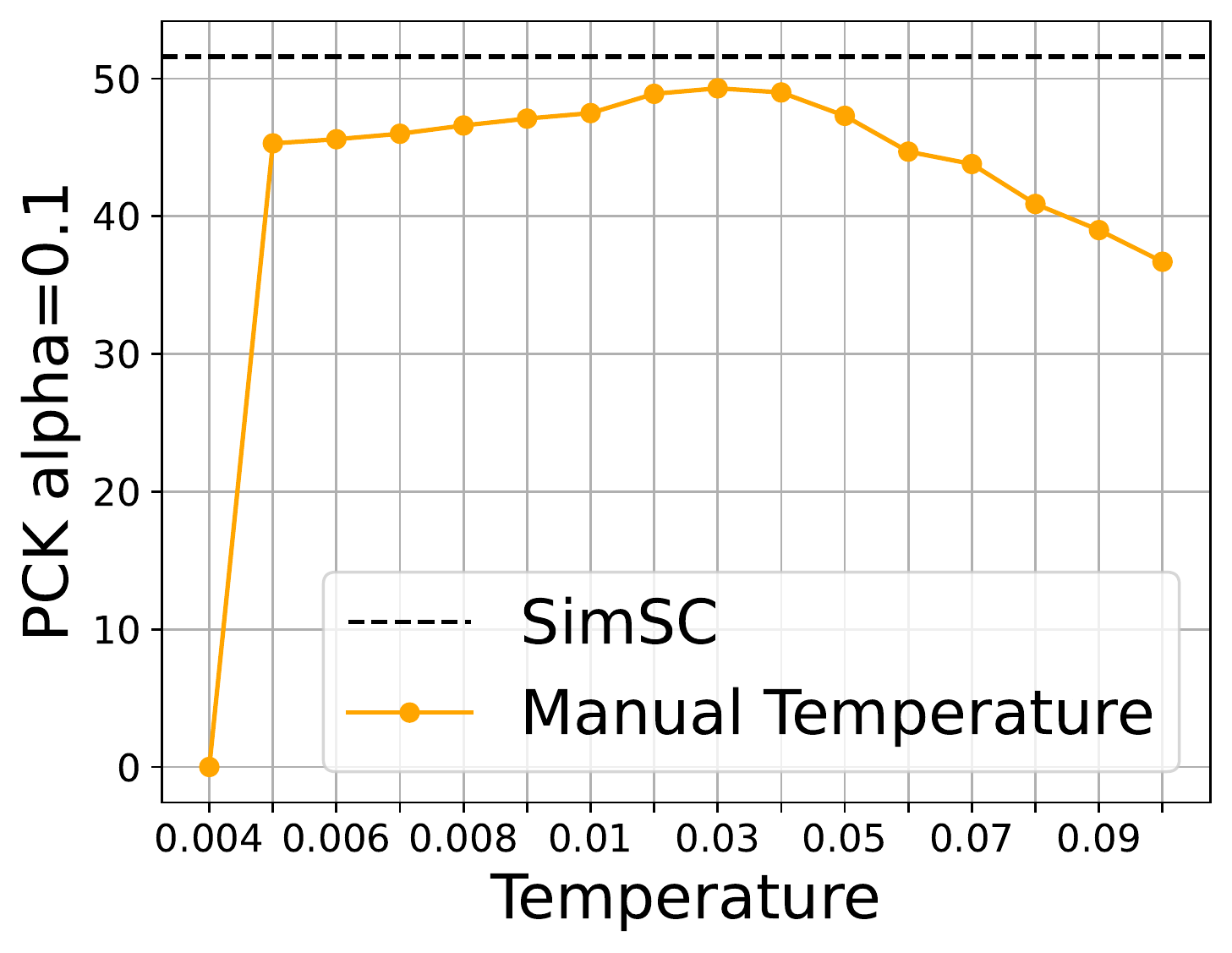}
        \caption{SPair-71K}
    \end{subfigure}
    \caption{Comparison between fine-tuning ResNet101* with different manual temperatures and SimSC on the PF-Pascal and SPair-71K datasets. For each sample of manual temperature, the evaluation temperature is set to be the same as the training temperature.}
    \label{fig:fixlr_varyingtemp}
\end{figure}
We fine-tune ResNet101* with different manual training temperatures on the PF-Pascal and SPair-71K datasets and plot the results together with results of SimSC in \cref{fig:fixlr_varyingtemp}. The evaluation temperature $\beta_{eval}$ is set to be the same as $\beta_{trn}$ for each sample. At $\beta_{trn}=0.004$, we obtain NaN error during training in both PF-Pascal and SPair-71K datasets, which is the reason for the poor performance at this temperature. We suspect that this is because the temperature is too small and the gradient around the maximum is too steep, causing instability during training. The performance increases until reaching the peak at $\beta_{trn}=0.03$ and starts decreasing.
\xh{The proposed temperature learning method can successfully predict the optimal temperatures across different datasets without the exhaustive grid search.}

\section{Learning-rate vs temperature effects}
As the temperature is important for the learning process by affecting the training signal (thus the gradients), one may wonder whether the same effects of the temperature can be trivially achieved by using a different learning rate. To validate this, we fine-tune ResNet101* and iBOT* with learning rates from $2\times 10^{-6}$ to $2\times 10^{-2}$ on PF-Pascal dataset using the baseline fine-tuning framework with $\beta_{trn}=1$ and summarize the results in \cref{tab:exp:lr_and_temp1}. As demonstrated, with $\beta_{trn}=1$, the baseline framework cannot achieve the same performance as our method regardless of the learning rate. This result indicates that the importance of temperature in softmax operation cannot be replaced by simply changing the learning rate. 

\begin{table}[ht]
\centering
\caption{Effect of different learning rate in fine-tuned feature extractors with $\beta_{trn}=1$. Apart from ``SimSC'', all models are fine-tuned on PF-Pascal dataset with $\beta_{trn}=1$ and evaluated with $\beta_{eval}=0.02$. For models trained with learning rate of $2\times 10^{-5}$, we also include the results that are read from Figure 6 in the main manuscript in the bracket because $\beta_{eval}=0.02$ is not the optimal value to them.}
\resizebox{.9\linewidth}{!}{
\begin{tabular}{cc|cc}
\hline
\multicolumn{2}{c|}{\textbf{ResNet101*}} & \multicolumn{2}{c}{\textbf{iBOT*}} \\
 & PF-Pascal &  & PF-Pascal \\
LR & $\alpha=0.1$ & LR & $\alpha=0.1$ \\ \hline
$2\times 10^{-6}$ & 79.9 & $2\times 10^{-6}$ & 76.3 \\
$2\times 10^{-5}$ & 80.3 (81.4) & $2\times 10^{-5}$ & 81.3 (83.5) \\
$2\times 10^{-4}$ & 78.0 & $2\times 10^{-4}$ & 74.6 \\
$2\times 10^{-3}$ & 63.5 & $2\times 10^{-3}$ & 4.5 \\
$2\times 10^{-2}$ & 54.6 & $2\times 10^{-2}$ & 4.4 \\ \hdashline
SimSC & 92.6 & SimSC & 95.6 \\ \hline
\end{tabular}}
\label{tab:exp:lr_and_temp1}
\end{table}

\section{Effect on other methods}
\begin{table}[t]
\centering
\caption{Coupling SimSC with various methods on PF-Pascal, given at threshold $\alpha$ = 0.05/0.1.}
\label{tab:simsc_with_sota}
\resizebox{.95\linewidth}{!}{
\begin{tabular}{c|c|c|c}
\hline
 & SFNet & NCNet & CATs \\ \hline
\multirow{2}{*}{\begin{tabular}[c]{@{}c@{}}Default\\ $\beta_{eval}$=0.02\end{tabular}} & \multirow{2}{*}{81.5 / 92.7} & \multirow{2}{*}{80.0 / 92.9} & \multirow{2}{*}{75.3 / 89.6} \\
 &  &  &  \\ \hline
\multirow{2}{*}{\begin{tabular}[c]{@{}c@{}}+ SimSC\\ $\beta_{eval}$=0.02\end{tabular}} & \multirow{2}{*}{80.6 / \textbf{93.0}} & \multirow{2}{*}{80.3 / 93.0} & \multirow{2}{*}{76.8 / 90.5} \\
 &  &  &  \\ \hline
\multirow{2}{*}{\begin{tabular}[c]{@{}c@{}}+ SimSC\\ $\beta_{eval}$=SimSC\end{tabular}} & \multirow{2}{*}{\textbf{82.3} / 92.9} & \multirow{2}{*}{\textbf{82.8} / \textbf{93.2}} & \multirow{2}{*}{\textbf{78.6} / \textbf{90.9}} \\
 &  &  &  \\ \hline
\end{tabular}}
\end{table}
\xh{Another question that one may consider is the effect of applying temperature learning to matching heads. We therefore apply the framework to three methods: SFNet, NCNet and CATs, corresponding to 2D conv, 4D conv and transformer matching heads respectively. Particularly, SFNet and CATs originally have a default temperature value of 0.02 while NCNet had a temperature value of 1. All methods are trained and tested on PF-Pascal dataset with 256$\times$256 images and cross-entropy loss. The results are in \cref{tab:simsc_with_sota}. As shown, the effect of temperature learning is not as significant as its effect on finetuning backbone. The reasons are two-fold. First, SFNet and CATs have already set a temperature value. Second, the final correlation tensors are processed by their respective matching heads. Therefore, the values in the correlation tensor are no longer confined between $[-1, 1]$ and they are not as overly smooth as the tensors directly constructed from feature maps after L2 Norm. However, we still observe performance gains when adding SimSC to NCNet and CATs. We additionally provide the evaluations on $\beta_{eval}$=0.02 for comparisons.}

\begin{table*}[t]
\centering
\caption{Comparison between the single-parameter design and the MLP design of the temperature learning module $g(\cdot)$. The evaluation metric is the same as described Sec. 4.1 and Table 1 of the main manuscript.}
\begin{tabular}{c|c|c|c|c|c}
\hline
\textbf{Feat. Ext.} & \multirow{2}{*}{$g(\cdot)$} & \textbf{PF-Pascal} & \textbf{PF-Willow} & \textbf{SPair-71K (T)} & \textbf{SPair-71K (F)} \\
(Eval. Size) &  & $\alpha(0.05/0.1/0.15)$ & $\alpha(0.05/0.1/0.15)$ & $\alpha(0.05/0.1)$ & $\alpha(0.05/0.1)$ \\ \hline
\textbf{ResNet101*} & single param. & 85.1 / 92.6 / 94.6 & 42.2 / \textbf{65.7} / \textbf{78.3} & 18.0 / 31.0 & \textbf{33.1} / 51.1 \\
(256) & MLP & \textbf{86.2} / 92.6 / \textbf{95.3} & \textbf{42.5} / 65.4 / 78.1 & \textbf{18.2} / \textbf{31.4} & 31.3 / \textbf{51.6} \\ \hdashline
\textbf{ResNet101*} & single param. & 85.0 / 92.8 / 95.5 & 39.5 / \textbf{68.5} / \textbf{82.2} & 24.5 / 35.9 & 35.6 / 48.2 \\
(ori.) & MLP & \textbf{85.3} / \textbf{93.3} / \textbf{96.0} & \textbf{40.7} / 68.0 / 80.6 & 24.5 / \textbf{36.0} & \textbf{36.6} / \textbf{51.5} \\ \hdashline
\textbf{iBOT*} & single param. & \textbf{89.7} / 94.9 / 96.7 & 44.3 / 68.9 / 82.2 & 20.1 / 33.4 & \textbf{43.3} / 62.0 \\
(256) & MLP & 88.4 / \textbf{95.6} / \textbf{97.3} & \textbf{44.9} / \textbf{71.4} / \textbf{84.5} & \textbf{22.0} / \textbf{37.9} & 43.0 / \textbf{63.5} \\ \hline
\end{tabular}
\label{supp:tab:singleVsMLP}
\end{table*}

\begin{table*}[t]
\centering
\caption{Comparison between the effective temperatures of the single-parameter design and the MLP design. For the MLP design, we calculate the mean and the standard deviation of temperatures for all image pairs in the testing set of each dataset. They are presented in the format of ``mean $\pm$ standard deviation''.}
\begin{tabular}{c|c|c|c|c|c}
\hline
\textbf{Feat. Ext.} & \multirow{2}{*}{$g(\cdot)$} & \multirow{2}{*}{\textbf{PF-Pascal}} & \multirow{2}{*}{\textbf{PF-Willow}} & \multirow{2}{*}{\textbf{SPair-71K (T)}} & \multirow{2}{*}{\textbf{SPair71-K (F)}} \\
(Eval. Size) &  &  &  &  &  \\ \hline
\textbf{ResNet101*} & single param. & 0.0333 & 0.0333 & 0.0333 & 0.0379 \\
(256) & MLP & 0.0303 $\pm$ 0.0017 & 0.0308 $\pm$ 0.0017 & 0.0305 $\pm$ 0.0015 & 0.0341 $\pm$ 0.0030 \\ \hdashline
\textbf{ResNet101*} & single param. & 0.0333 & 0.0333 & 0.0333 & 0.0379 \\
(ori.) & MLP & 0.0307 $\pm$ 0.0018 & 0.0314 $\pm$ 0.0025 & 0.0370 $\pm$ 0.0016 & 0.0408 $\pm$ 0.0031 \\ \hdashline
\textbf{iBOT*} & single param. & 0.0289 & 0.0289 & 0.0289 & 0.0352 \\
(256) & MLP & 0.0251 $\pm$ 0.0025 & 0.0254 $\pm$ 0.0025 & 0.2585 $\pm$ 0.0030 & 0.0324 $\pm$ 0.0037 \\ \hline
\end{tabular}
\label{supp:tab:single_mlp_temp}
\end{table*}

\section{The single-parameter design}
We explore a simplified design of the temperature learning module that consists of one single learnable parameter $\beta_{c}$, which takes no input and does not depend on anything. The effective temperature is replaced by $\beta^{2}_{c}$. We evaluate this simplified design on ResNet101* and iBOT*. We notice that the single-parameter design is sensitive to the learning rate of the parameter and requires careful tuning. We experiment with several learning rates on the PF-Pascal dataset using ResNet101* and choose the best value $0.005$ based on the performance on the validation set. For a fair comparison with the MLP design of the main manuscript, we apply this value to both ResNet101* and iBOT* across all datasets. The rest of the configuration is the same as in Sec. 4.2 in the main manuscript. We summarize the results of both the single-parameter design and the MLP design in \cref{supp:tab:singleVsMLP}. 

On the PF-Pascal dataset, the single-parameter design and the MLP design achieve close results in ResNet101*, provided that the MLP design slightly outperforms the single-parameter design on PF-Pascal and SPair-71K (F) at the original image size. However, the MLP design noticeably outperforms the single-parameter design in iBOT* on all three datasets. When switching to the single-parameter design, we notice a moderate drop in the generalizability of transferring iBOT* trained on PF-Pascal to PF-Willow and SPair-71K. \xh{This is because different feature extractors may have different learning curves, so the learning rate found for ResNet101* may not be optimal for iBOT*. The single-parameter design does not have the contextual information of the feature map in the temperature learning. The temperature update solely depends on the learning rate. The optimal learning rate may depend on the architecture and dataset. }

We also include the learned effective temperature of each feature extractor on each dataset in \cref{supp:tab:single_mlp_temp} for reference. For the MLP design, since different image pairs have different temperatures, we calculate the mean and the standard deviation of temperatures of all image pairs in testing set of each dataset.


\section{Qualitative examples}
\begin{figure*}
    \centering
        \begin{subfigure}{.33\linewidth}
        \centering
        \includegraphics[width=\linewidth]{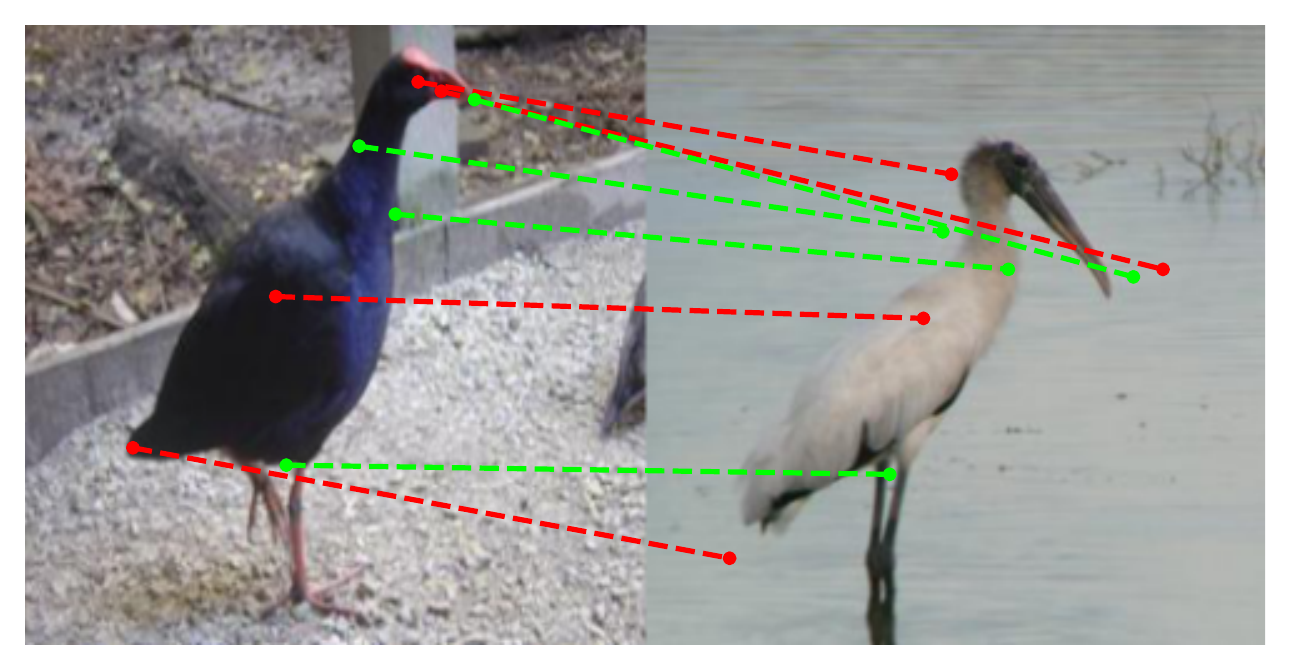}
    \end{subfigure}%
    \begin{subfigure}{.33\linewidth}
        \centering
        \includegraphics[width=\linewidth]{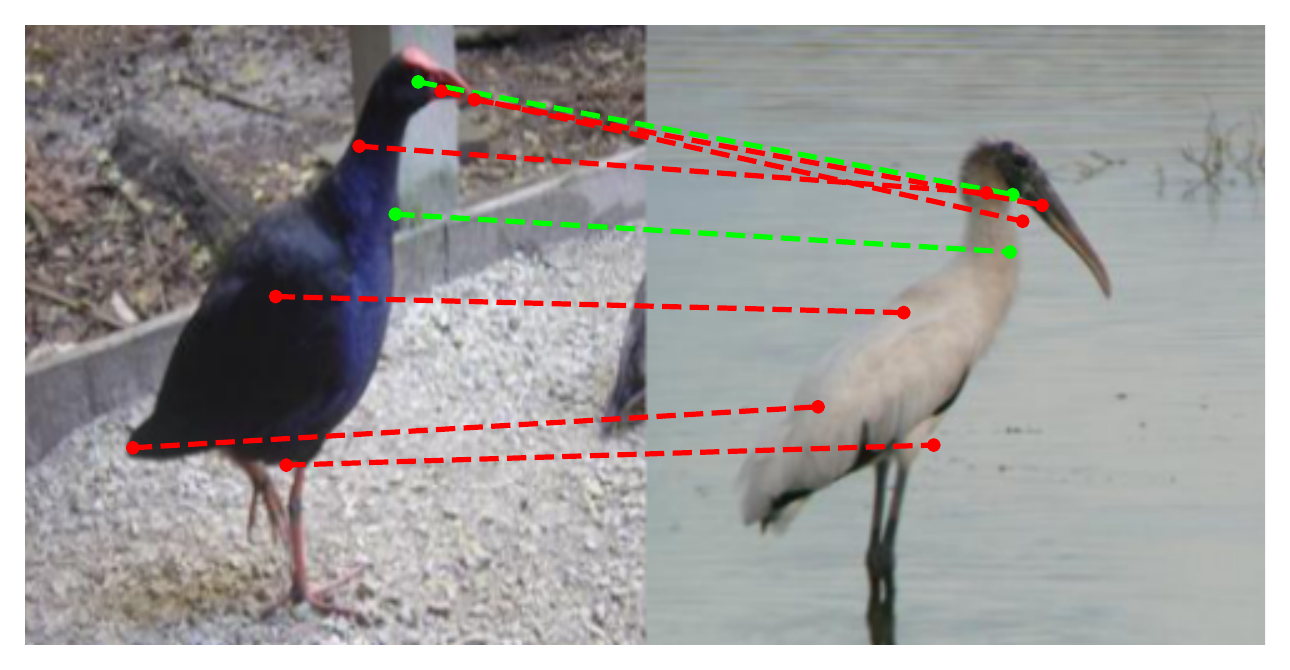}
    \end{subfigure}%
    \begin{subfigure}{.33\linewidth}
        \centering
        \includegraphics[width=\linewidth]{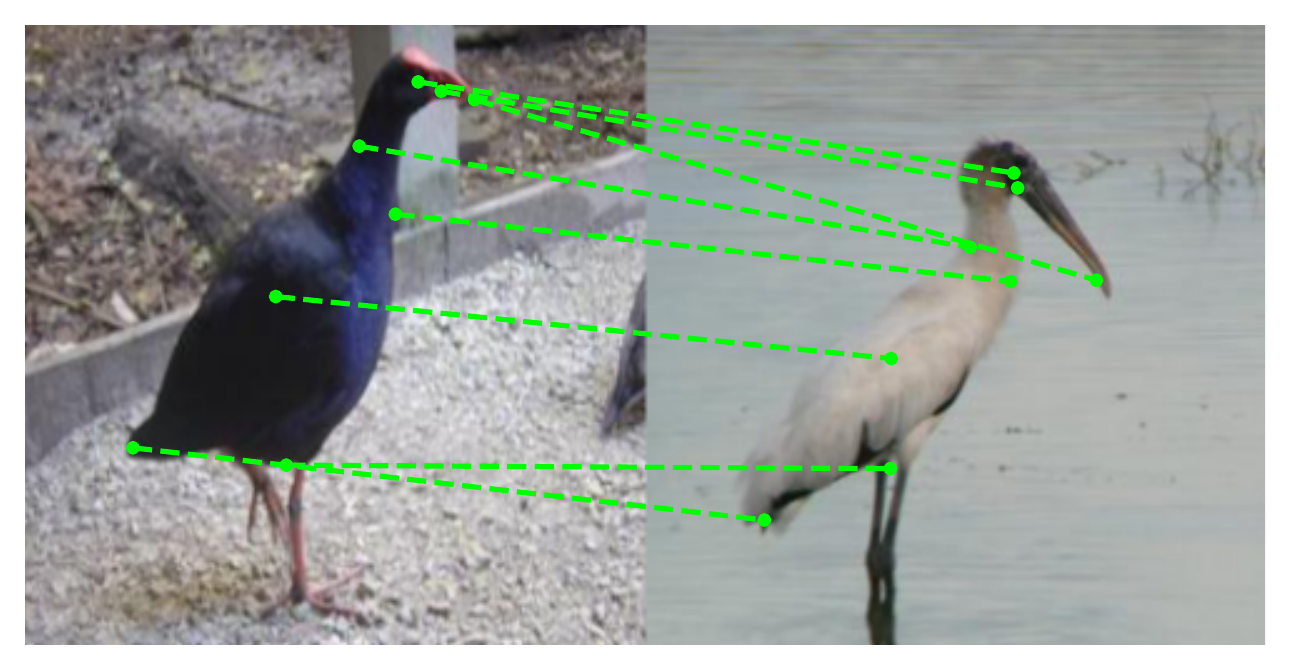}
    \end{subfigure}
        \begin{subfigure}{.33\linewidth}
        \centering
        \includegraphics[width=\linewidth]{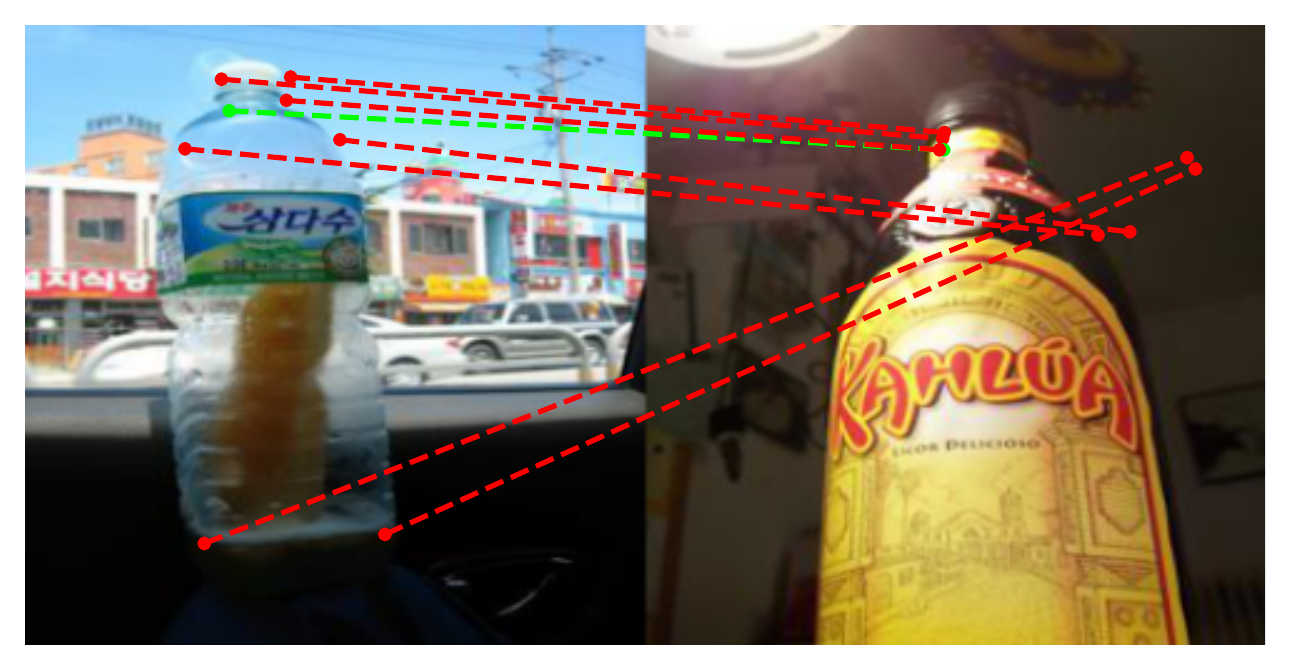}
    \end{subfigure}%
    \begin{subfigure}{.33\linewidth}
        \centering
        \includegraphics[width=\linewidth]{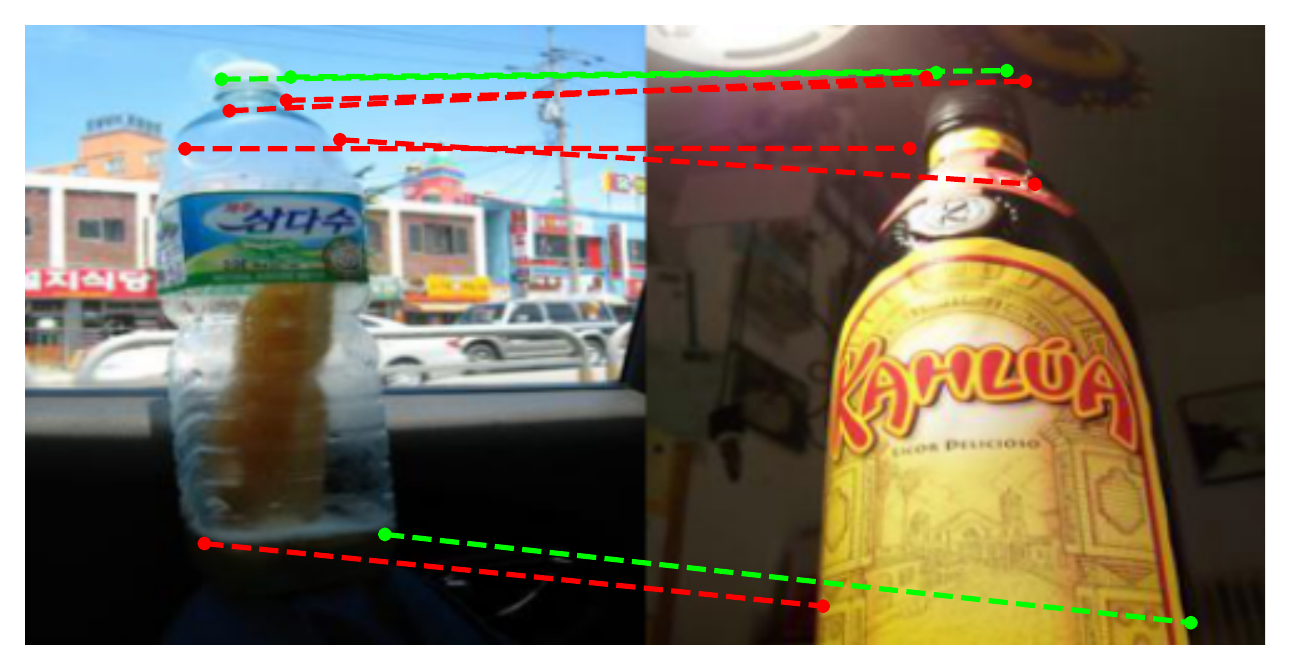}
    \end{subfigure}%
    \begin{subfigure}{.33\linewidth}
        \centering
        \includegraphics[width=\linewidth]{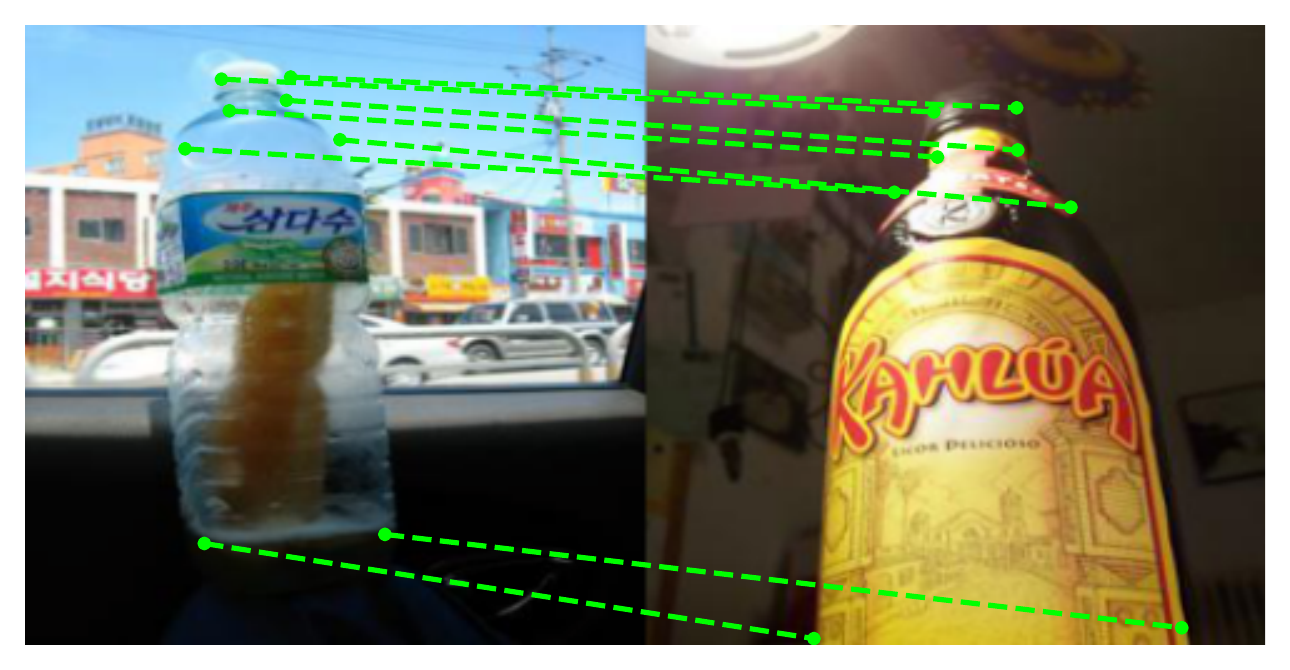}
    \end{subfigure}
        \begin{subfigure}{.33\linewidth}
        \centering
        \includegraphics[width=\linewidth]{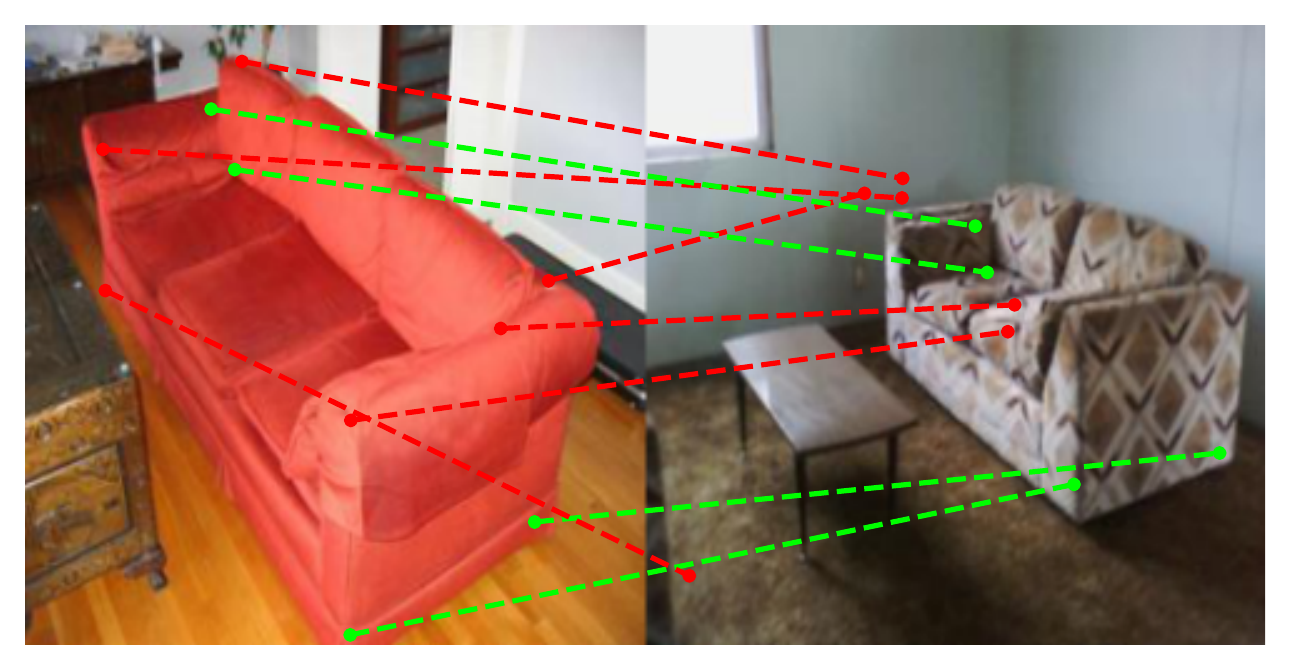}
        \caption{ImageNet pre-trained}
    \end{subfigure}%
    \begin{subfigure}{.33\linewidth}
        \centering
        \includegraphics[width=\linewidth]{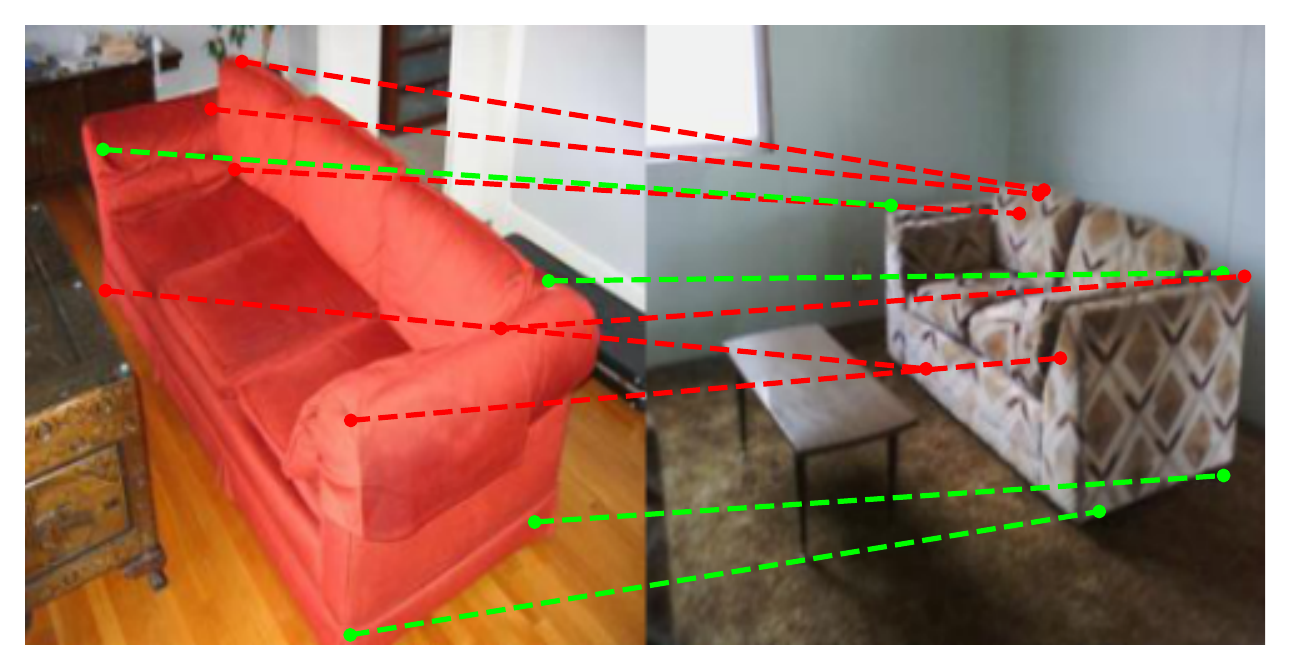}
        \caption{Finetuned with $\beta_{trn}=1$}
    \end{subfigure}%
    \begin{subfigure}{.33\linewidth}
        \centering
        \includegraphics[width=\linewidth]{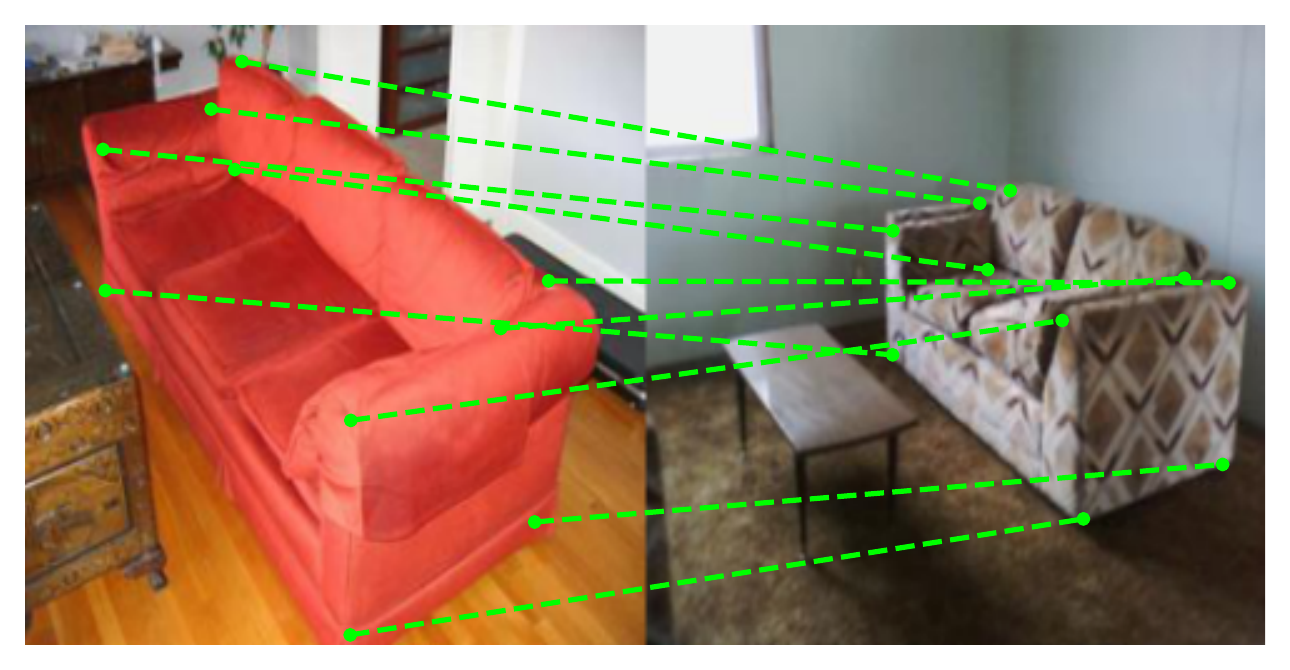}
        \caption{Finetuned with SimSC}
    \end{subfigure}%
    \caption{Comparison between semantic matching of ImageNet pre-trained ResNet101*, finetuned ResNet101* with $\beta_{trn}=1$ and SimSC-ResNet101* on the PF-Pascal dataset.}
    \label{supp:fig:imgnet_temp1_simsc}
\end{figure*}

\begin{figure*}
    \centering
    \begin{subfigure}{.33\linewidth}
        \centering
        \includegraphics[width=\linewidth]{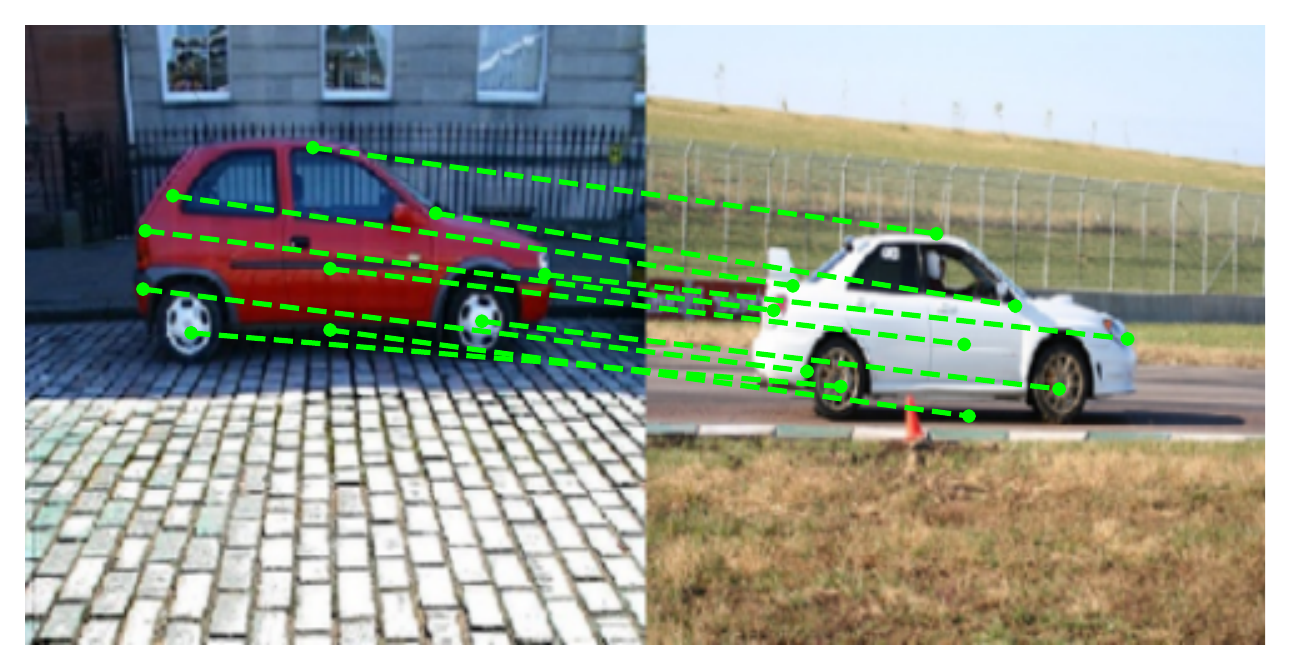}
    \end{subfigure}%
    \begin{subfigure}{.33\linewidth}
        \centering
        \includegraphics[width=\linewidth]{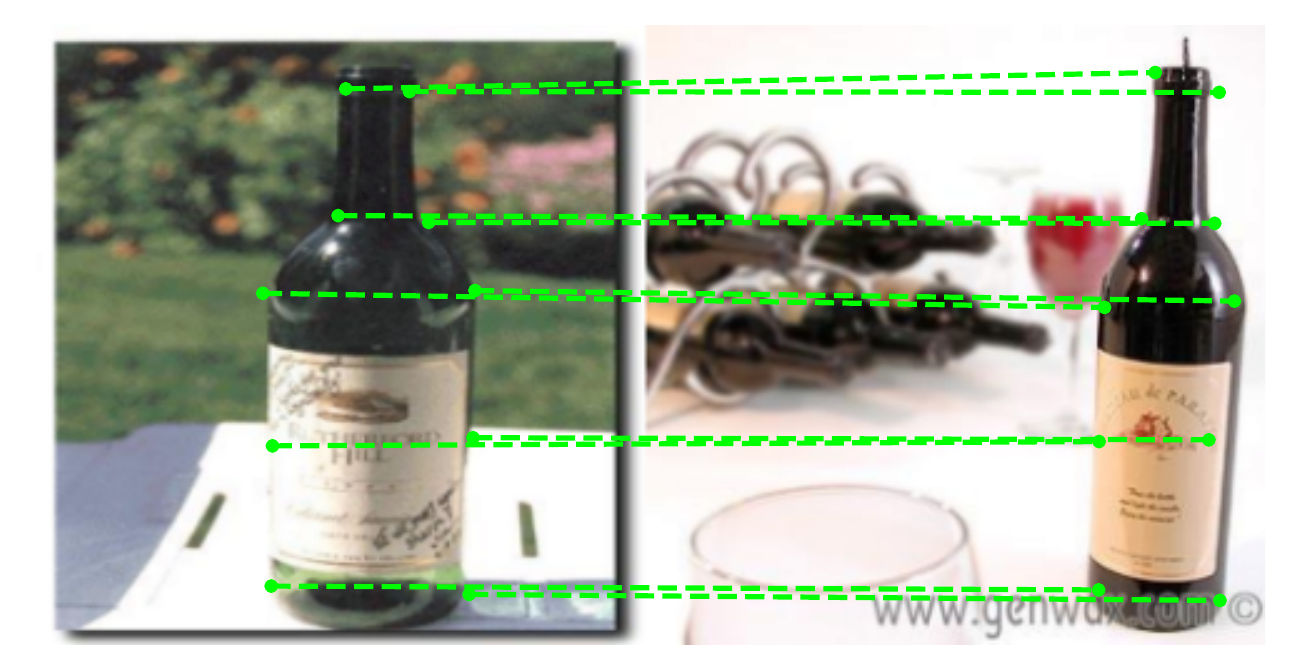}
    \end{subfigure}%
    \begin{subfigure}{.33\linewidth}
        \centering
        \includegraphics[width=\linewidth]{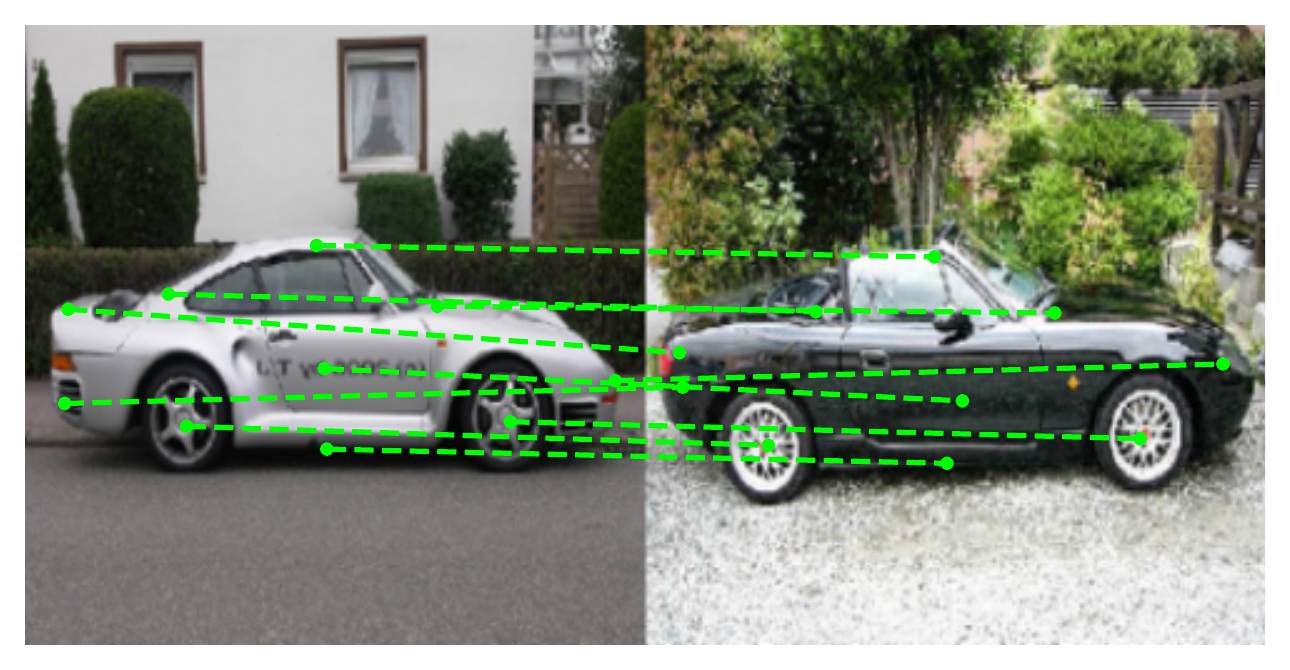}
    \end{subfigure}
    \caption{Examples of semantic matching by SimSC-ResNet101* on the PF-Willow dataset.}
    \label{supp:fig:resnet101_pfwillow}
\end{figure*}

\begin{figure*}
    \centering
    \begin{subfigure}{.33\linewidth}
        \centering
        \includegraphics[width=\linewidth]{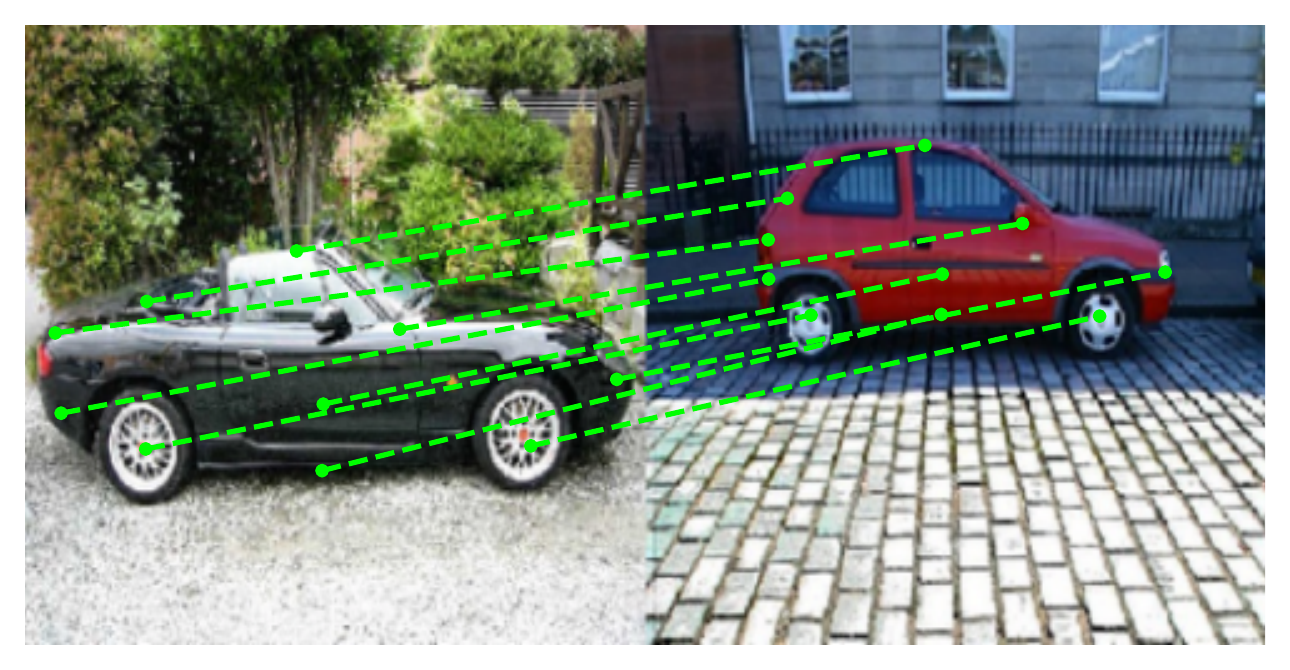}
    \end{subfigure}%
    \begin{subfigure}{.33\linewidth}
        \centering
        \includegraphics[width=\linewidth]{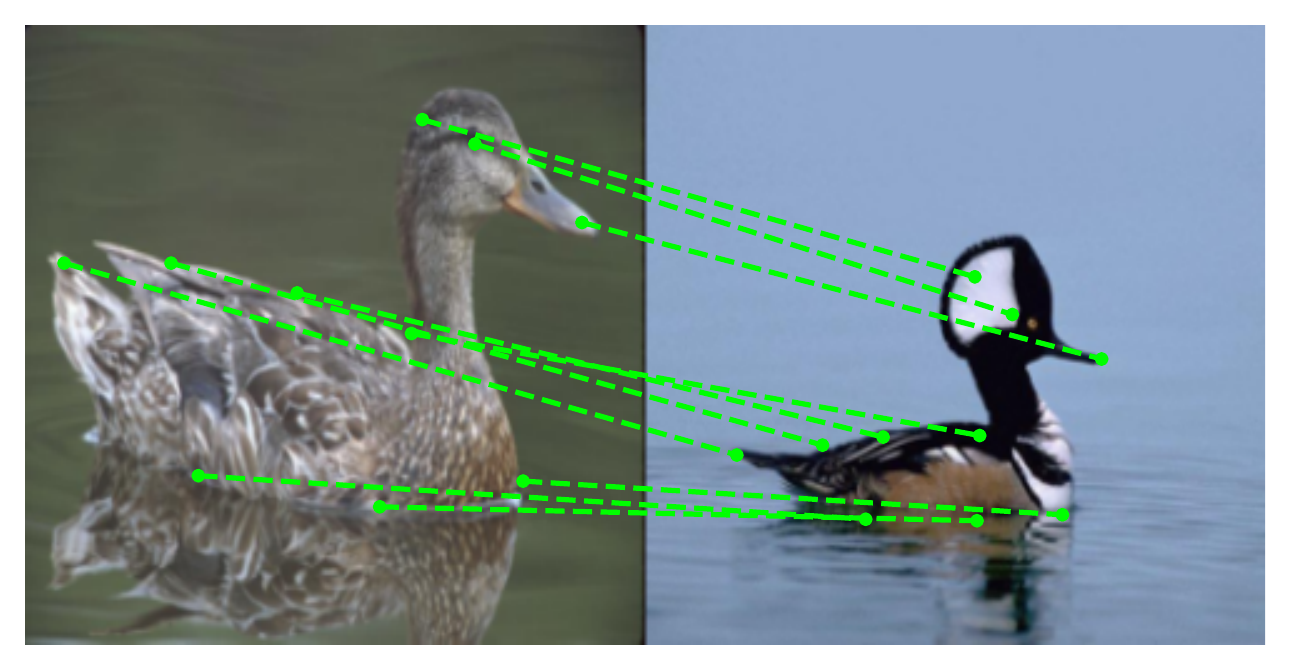}
    \end{subfigure}%
    \begin{subfigure}{.33\linewidth}
        \centering
        \includegraphics[width=\linewidth]{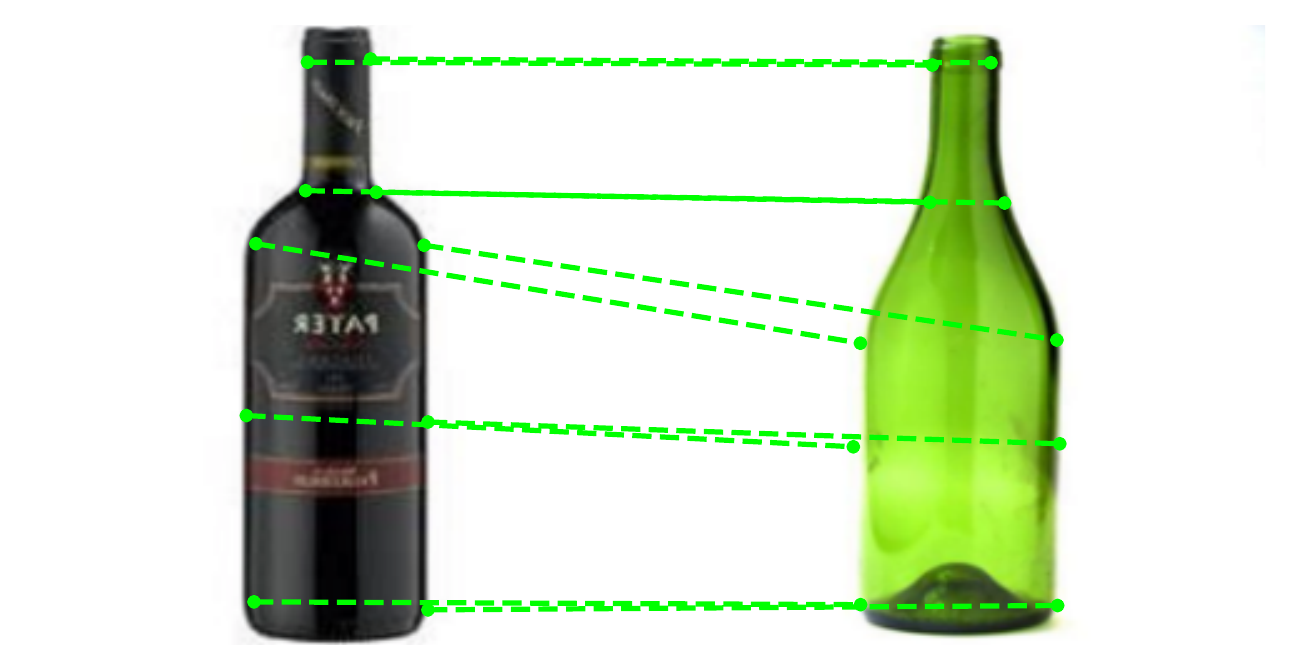}
    \end{subfigure}
    \caption{Examples of semantic matching by SimSC-iBOT* on the PF-Willow dataset.}
    \label{supp:fig:ibot_pfwillow}
\end{figure*}

\begin{figure*}
    \centering
    \begin{subfigure}{.33\linewidth}
        \centering
        \includegraphics[width=\linewidth]{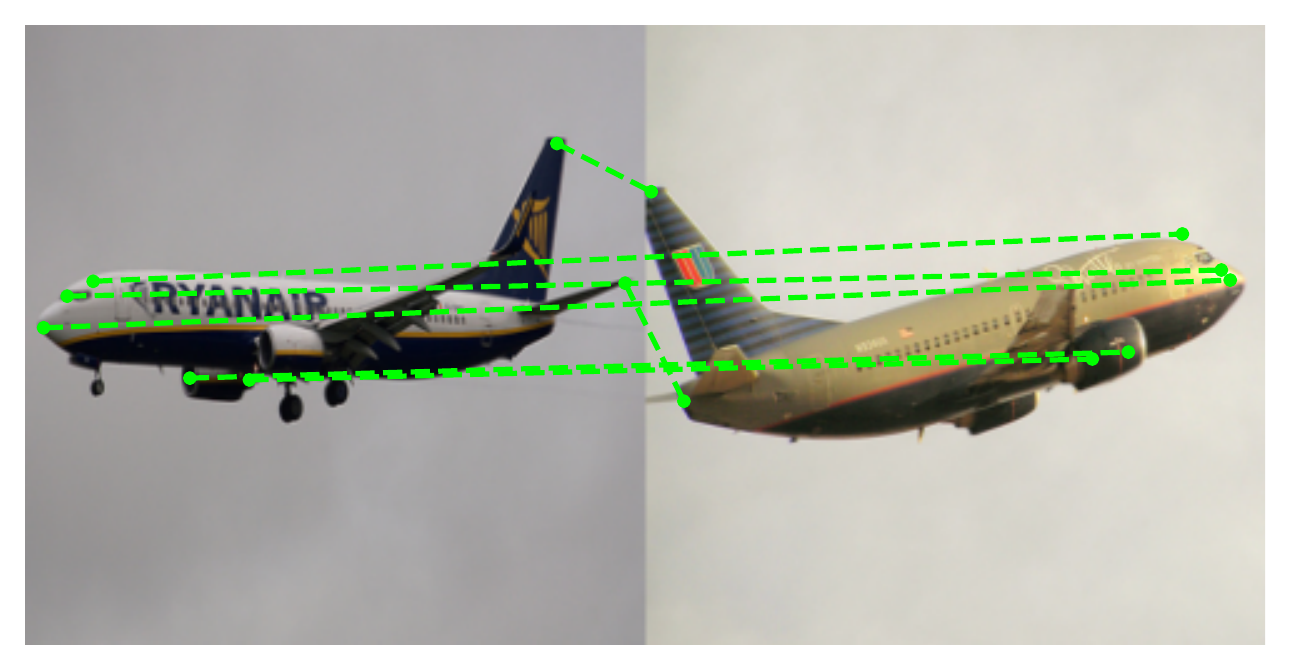}
    \end{subfigure}%
    \begin{subfigure}{.33\linewidth}
        \centering
        \includegraphics[width=\linewidth]{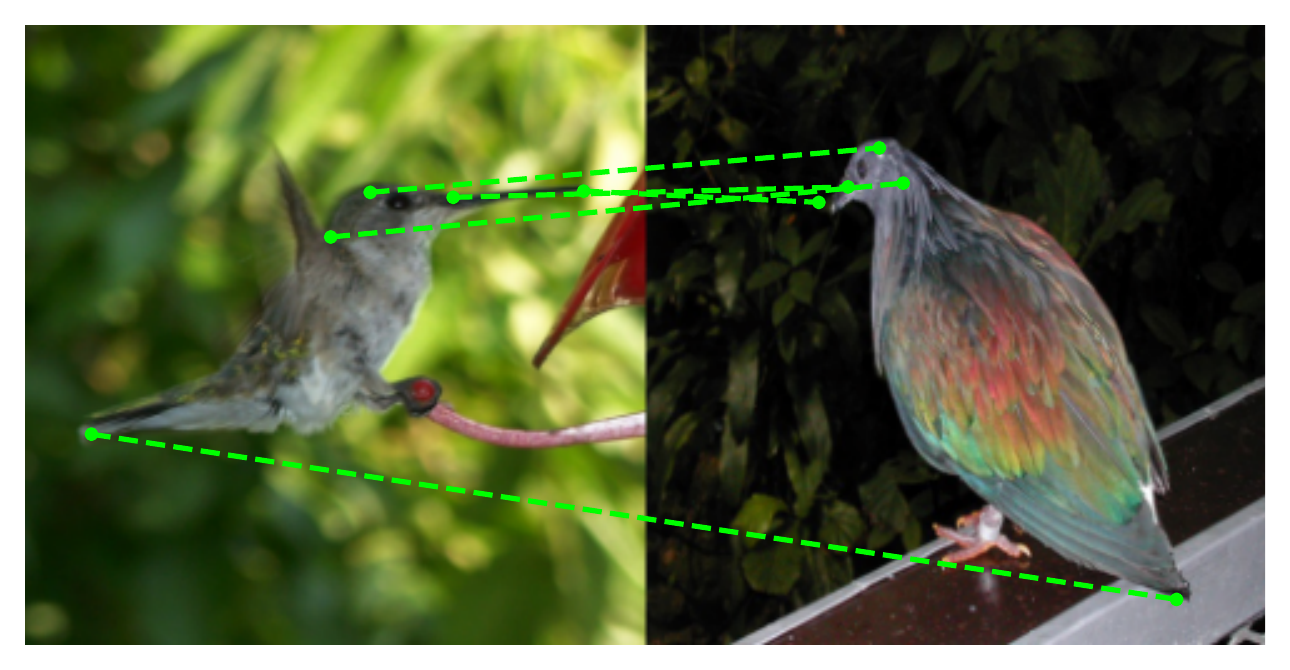}
    \end{subfigure}%
    \begin{subfigure}{.33\linewidth}
        \centering
        \includegraphics[width=\linewidth]{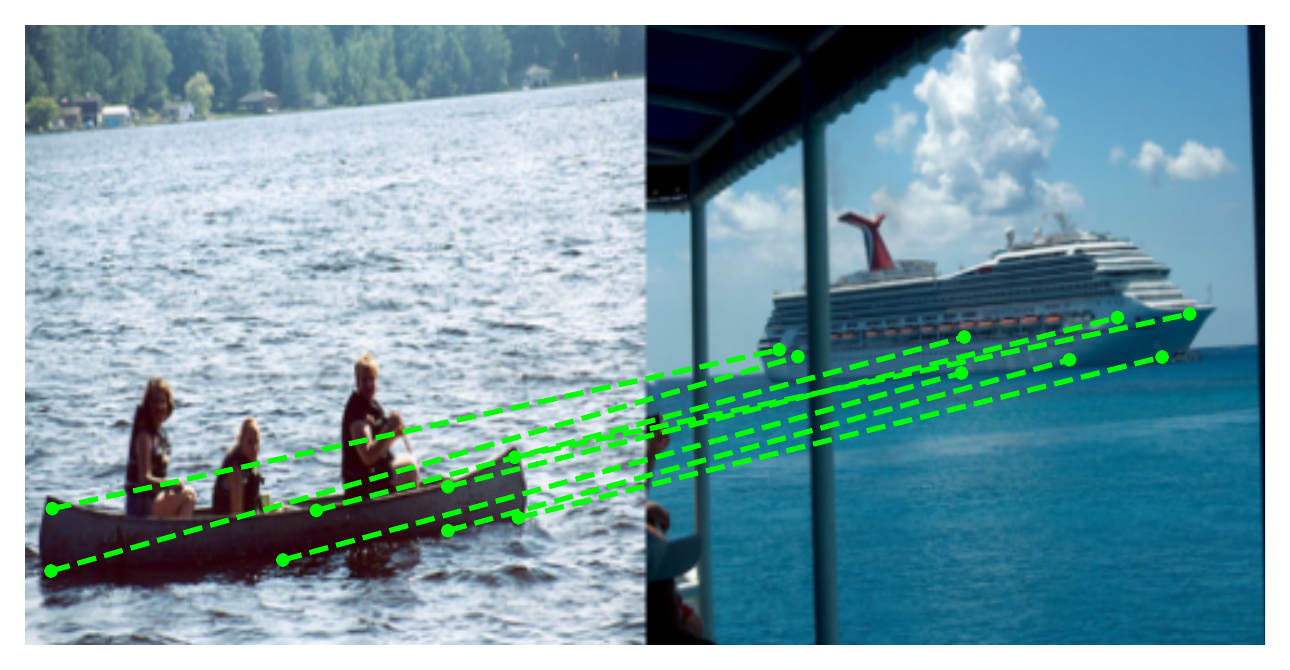}
    \end{subfigure}
        \begin{subfigure}{.33\linewidth}
        \centering
        \includegraphics[width=\linewidth]{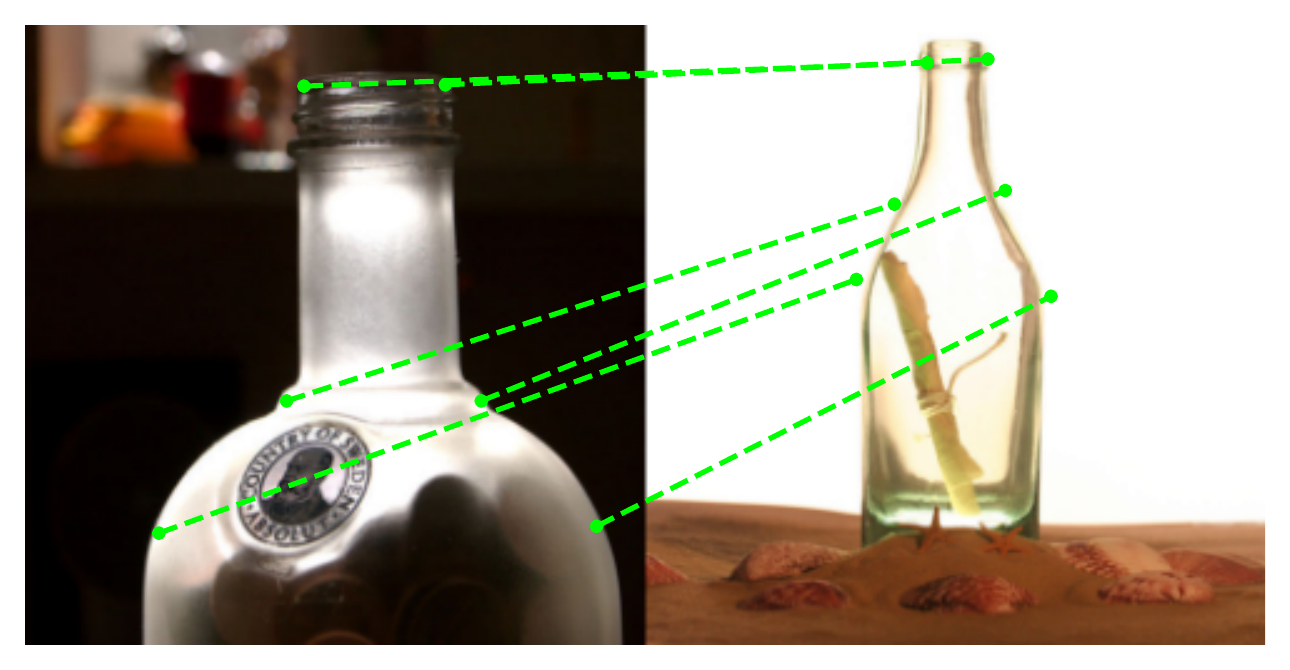}
    \end{subfigure}%
    \begin{subfigure}{.33\linewidth}
        \centering
        \includegraphics[width=\linewidth]{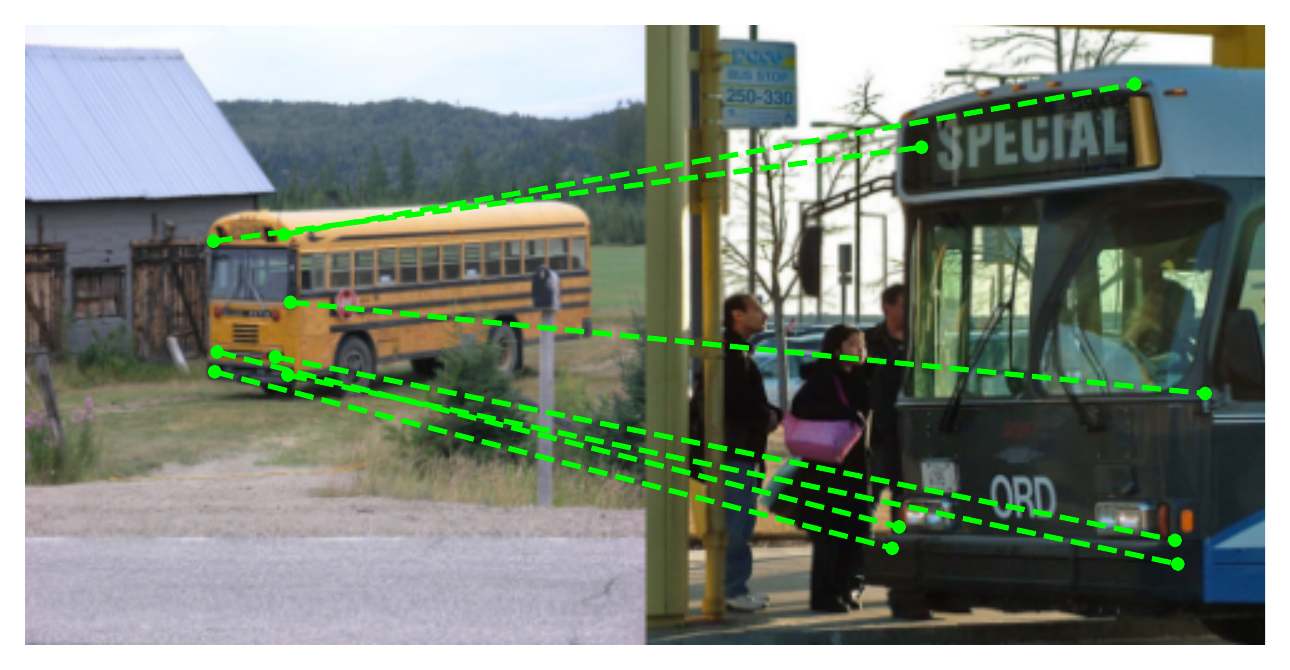}
    \end{subfigure}%
    \begin{subfigure}{.33\linewidth}
        \centering
        \includegraphics[width=\linewidth]{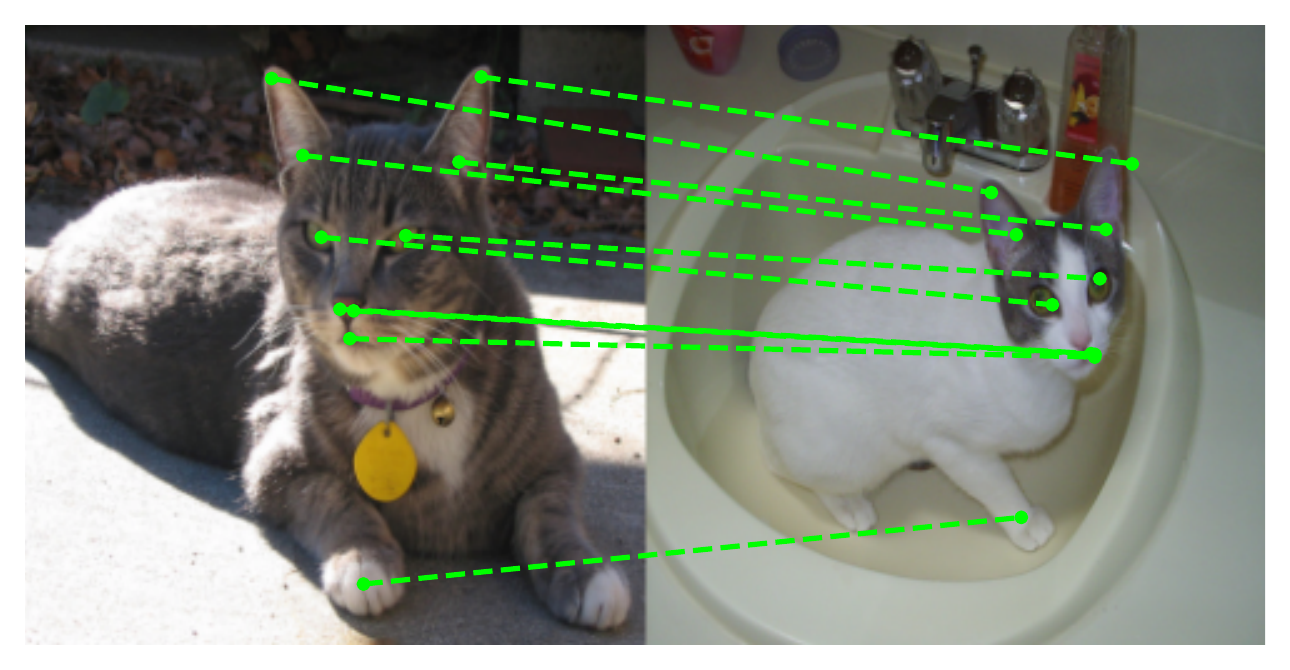}
    \end{subfigure}
        \begin{subfigure}{.33\linewidth}
        \centering
        \includegraphics[width=\linewidth]{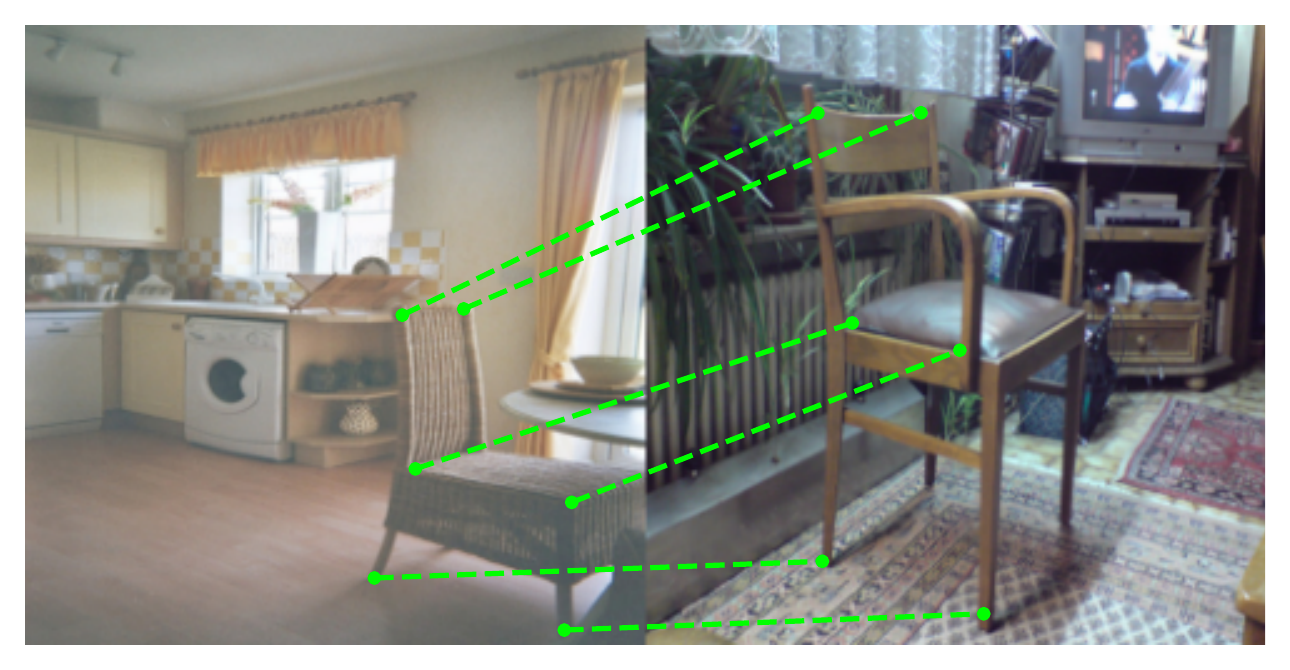}
    \end{subfigure}%
    \begin{subfigure}{.33\linewidth}
        \centering
        \includegraphics[width=\linewidth]{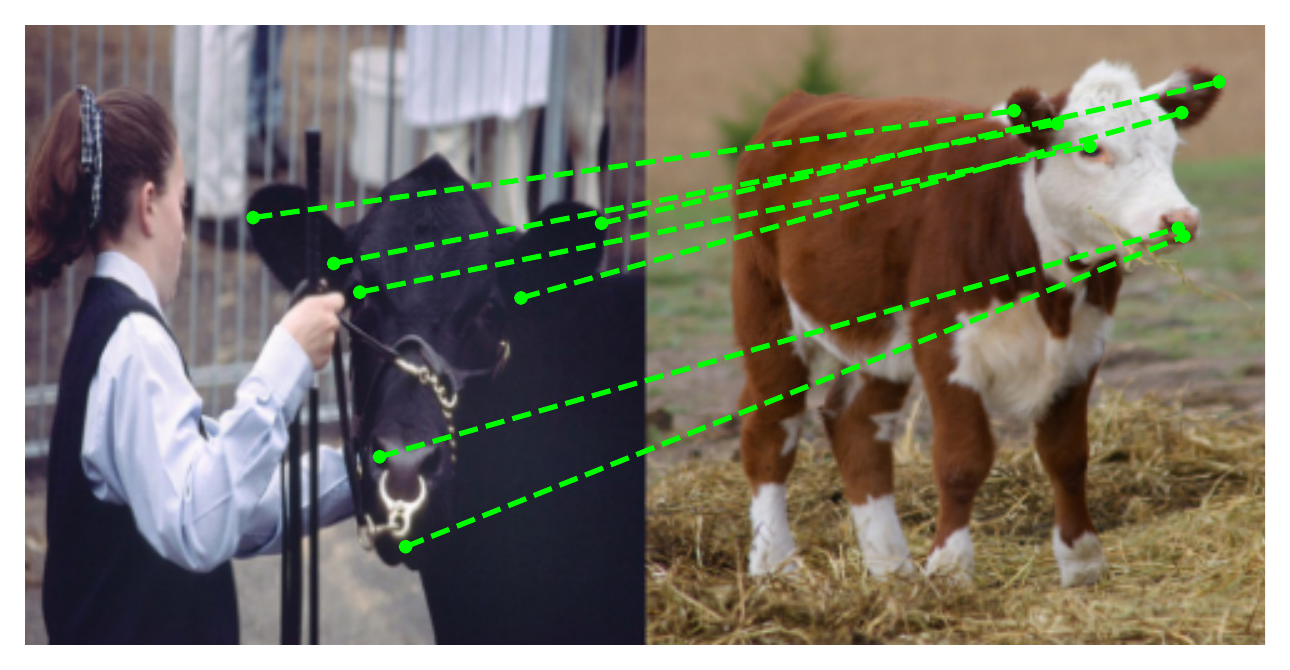}
    \end{subfigure}%
    \begin{subfigure}{.33\linewidth}
        \centering
        \includegraphics[width=\linewidth]{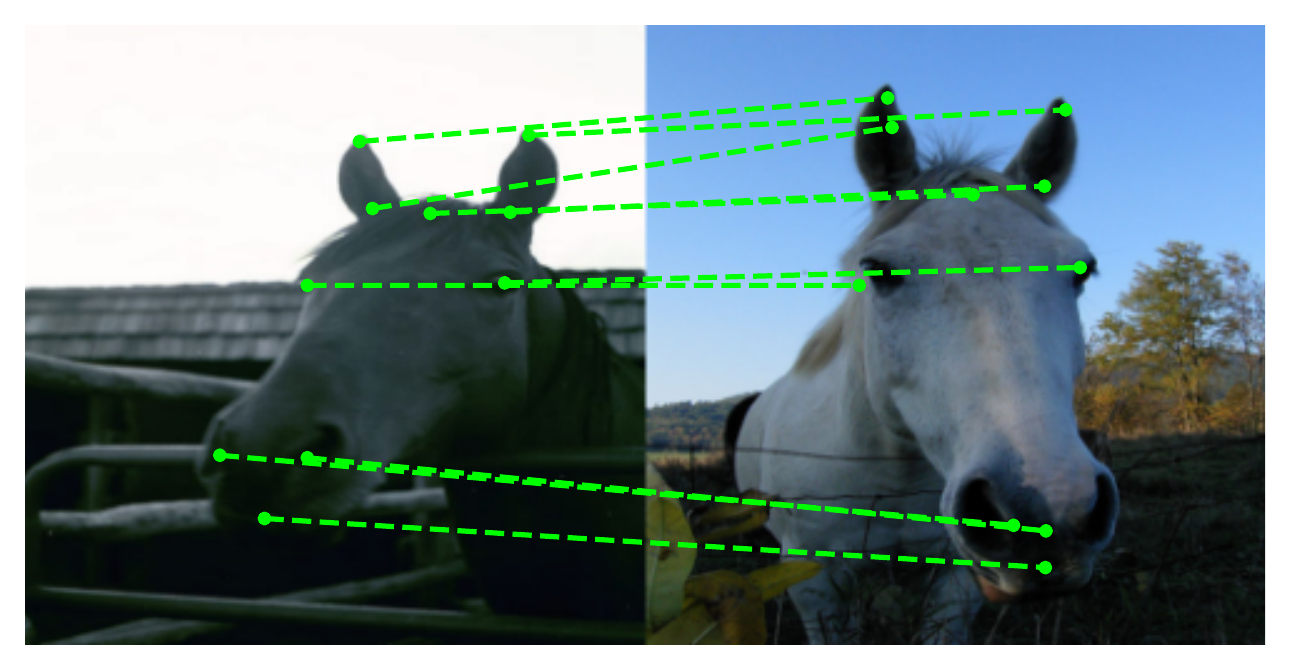}
    \end{subfigure}
    \caption{Examples of semantic matching by SimSC-ResNet101* on the SPair-71K dataset.}
    \label{supp:fig:resnet101_spair}
\end{figure*}

\begin{figure*}
    \centering
    \begin{subfigure}{.33\linewidth}
        \centering
        \includegraphics[width=\linewidth]{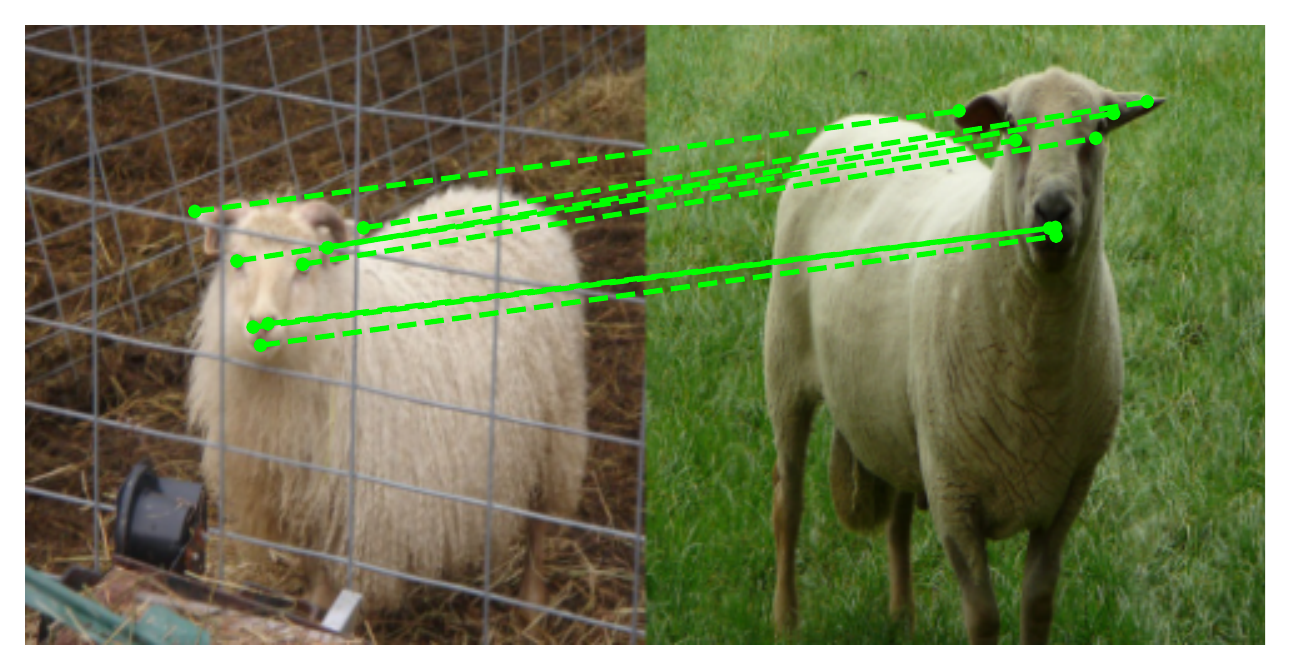}
    \end{subfigure}%
    \begin{subfigure}{.33\linewidth}
        \centering
        \includegraphics[width=\linewidth]{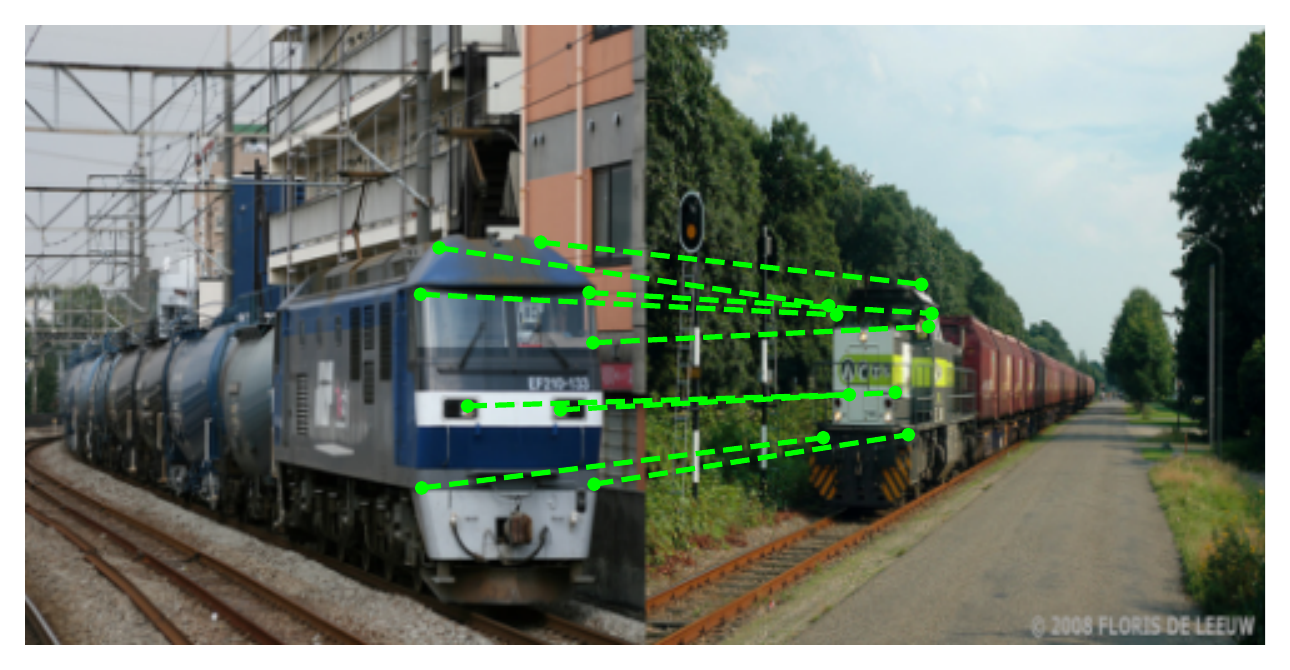}
    \end{subfigure}%
    \begin{subfigure}{.33\linewidth}
        \centering
        \includegraphics[width=\linewidth]{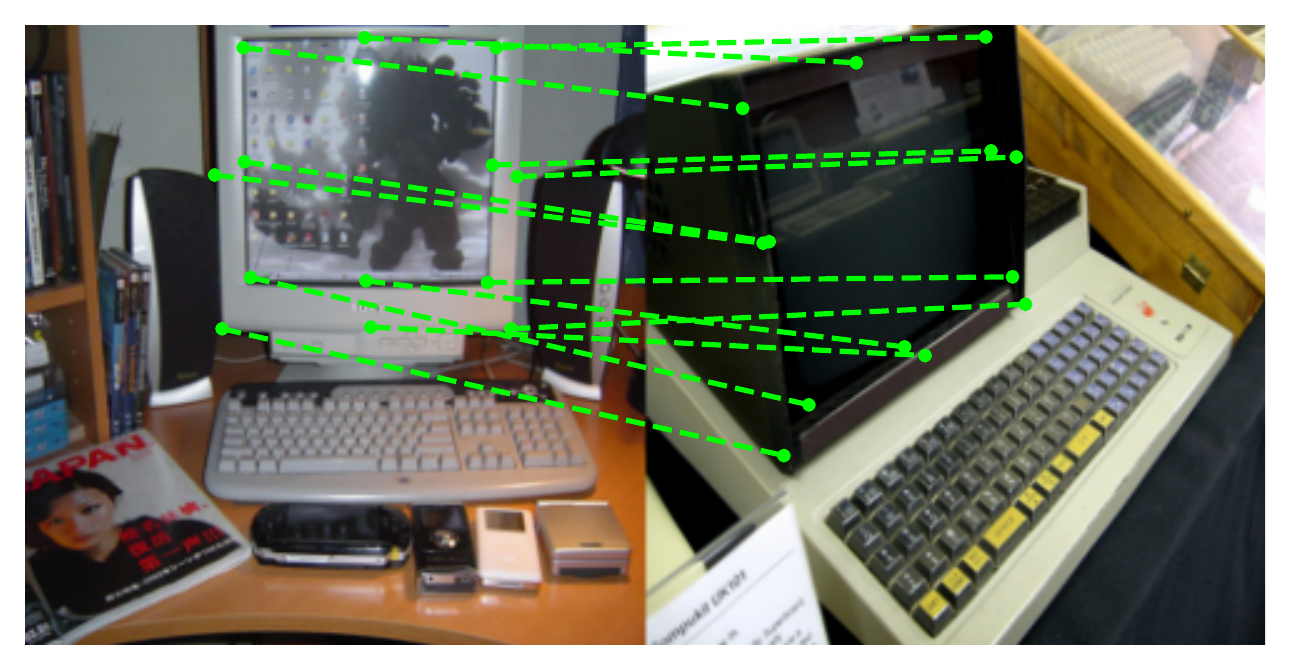}
    \end{subfigure}
        \begin{subfigure}{.33\linewidth}
        \centering
        \includegraphics[width=\linewidth]{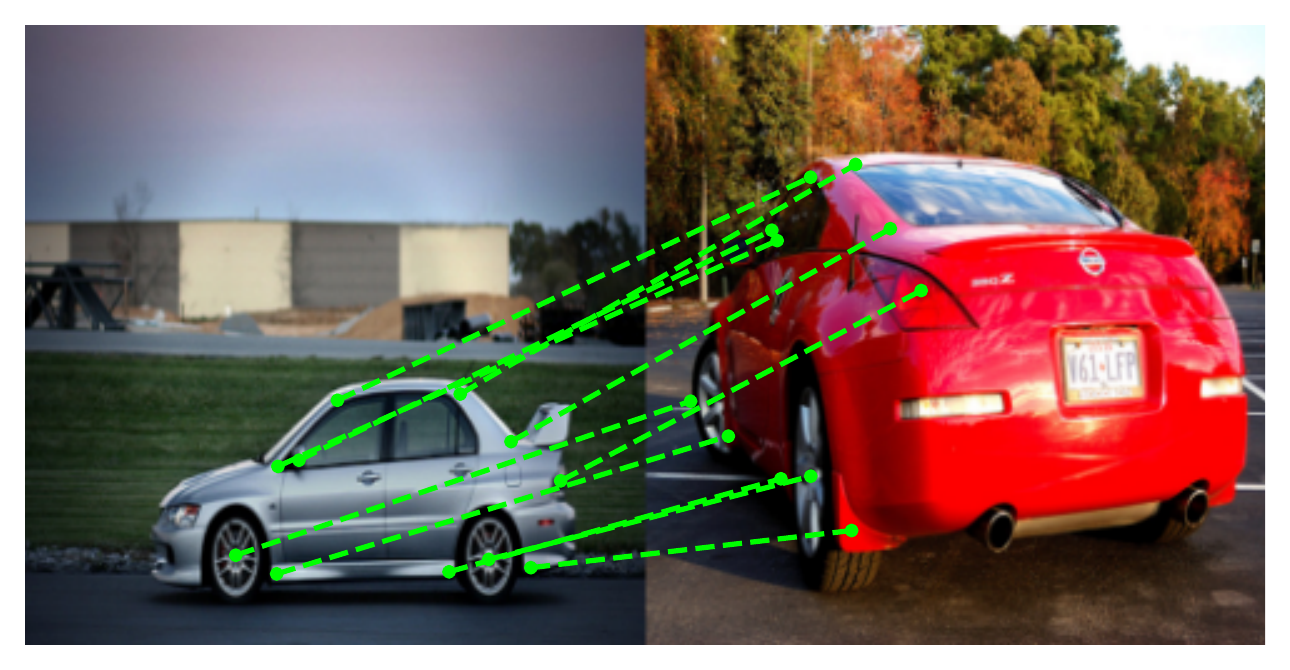}
    \end{subfigure}%
    \begin{subfigure}{.33\linewidth}
        \centering
        \includegraphics[width=\linewidth]{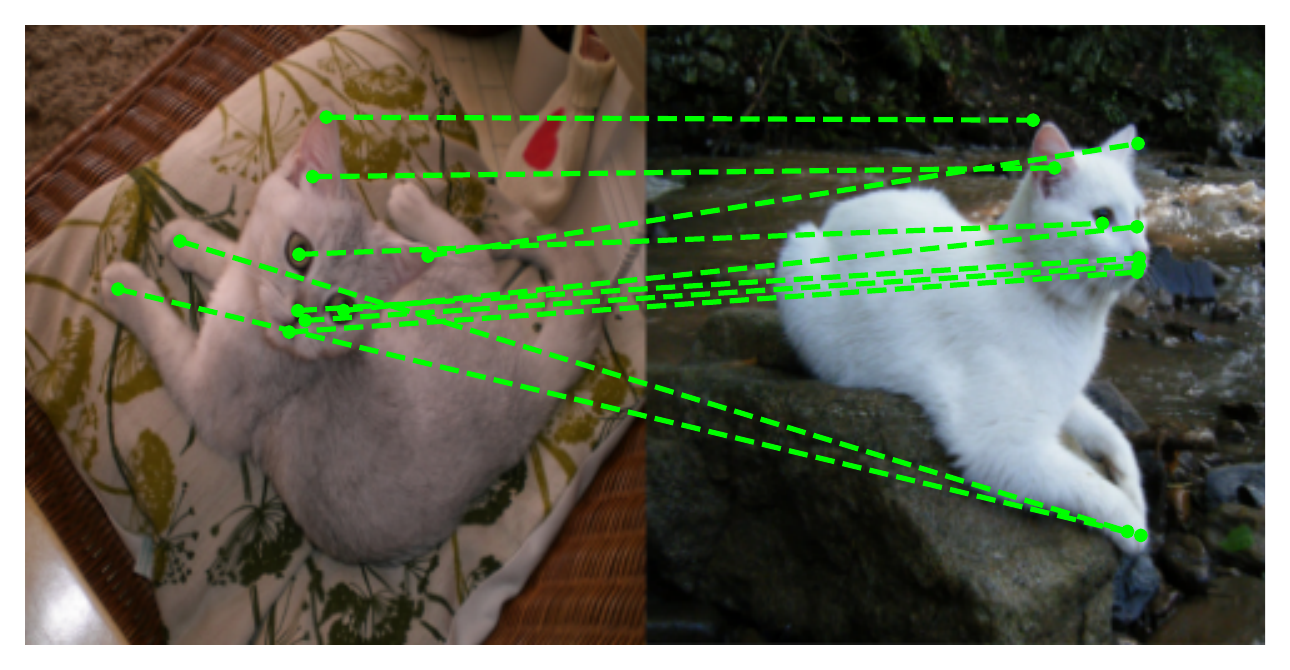}
    \end{subfigure}%
    \begin{subfigure}{.33\linewidth}
        \centering
        \includegraphics[width=\linewidth]{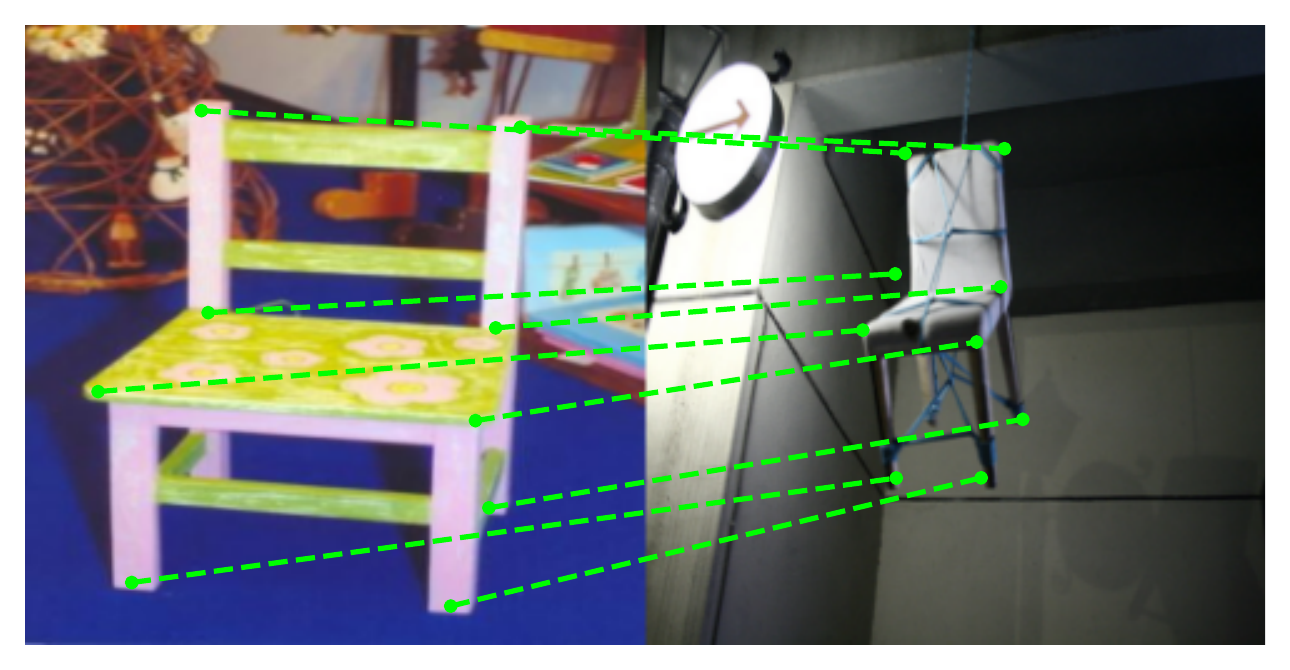}
    \end{subfigure}
        \begin{subfigure}{.33\linewidth}
        \centering
        \includegraphics[width=\linewidth]{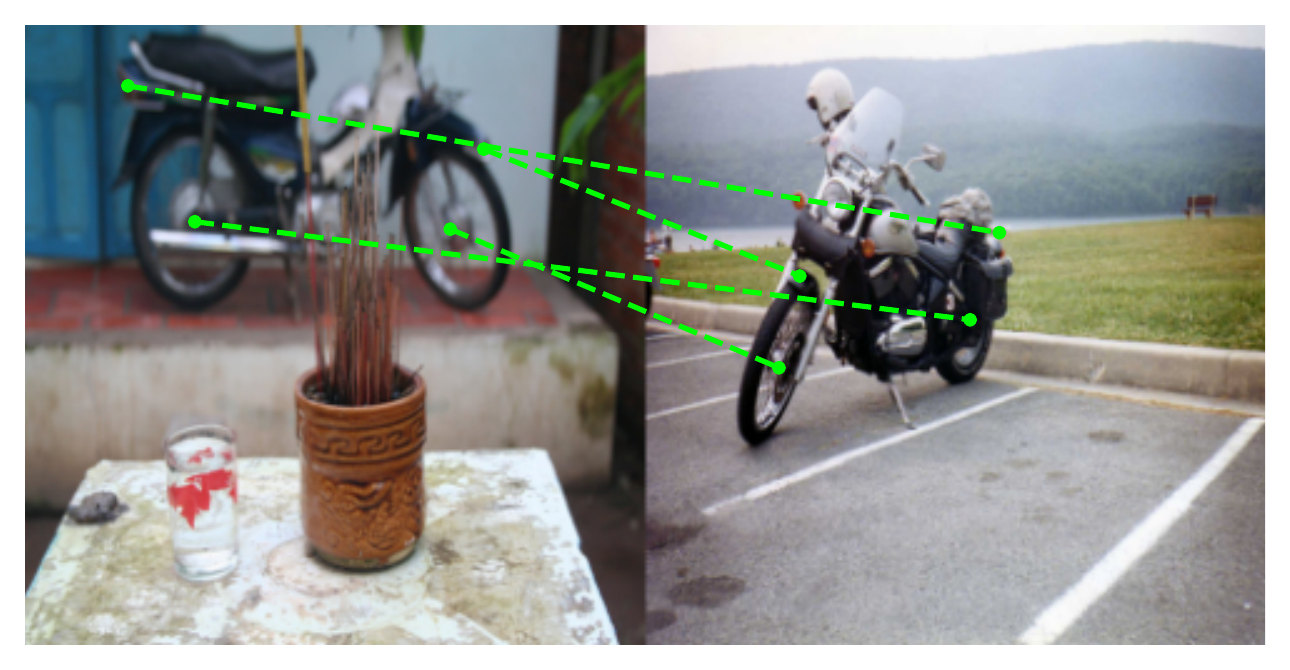}
    \end{subfigure}%
    \begin{subfigure}{.33\linewidth}
        \centering
        \includegraphics[width=\linewidth]{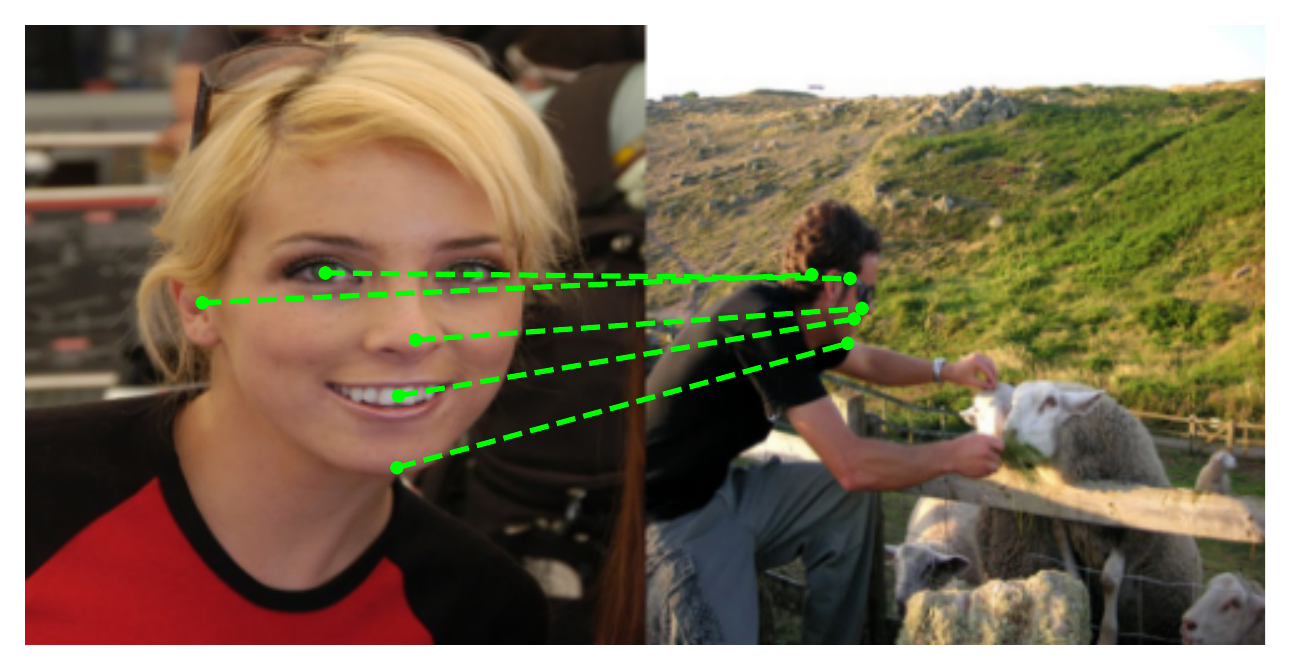}
    \end{subfigure}
    \begin{subfigure}{.33\linewidth}
        \centering
        \includegraphics[width=\linewidth]{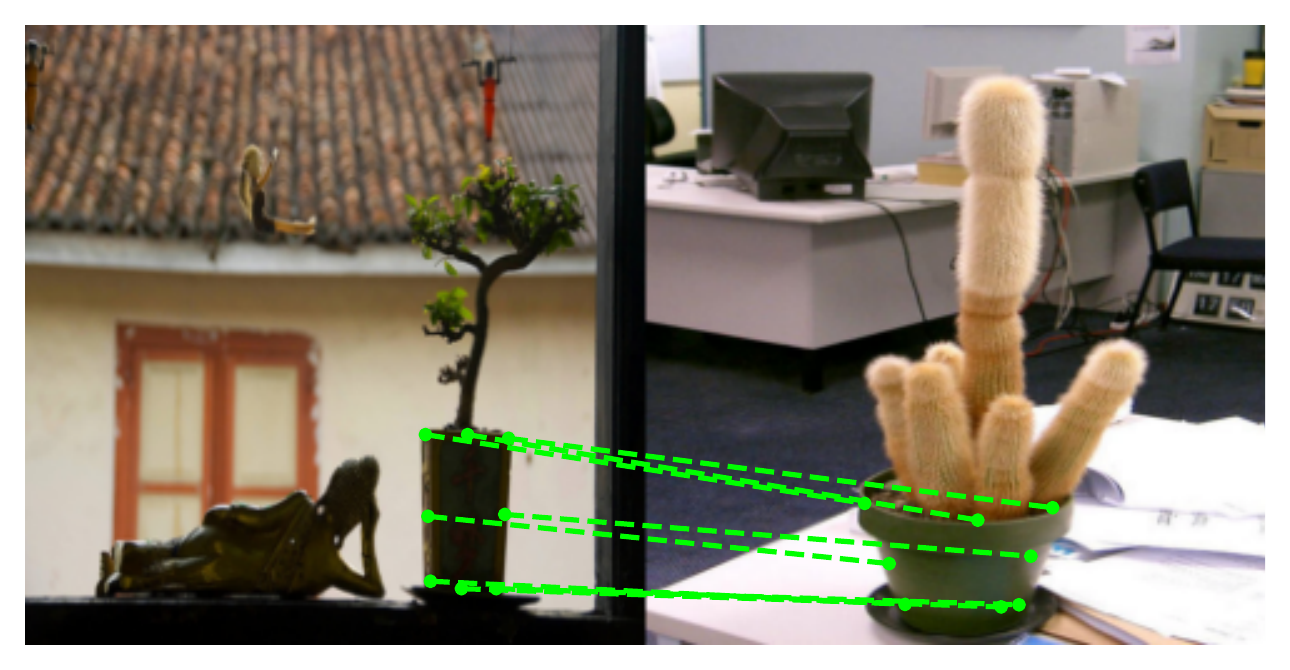}
    \end{subfigure}%
    \caption{Examples of semantic matching by SimSC-iBOT* on the SPair-71K dataset.}
    \label{supp:fig:ibot_spair}
\end{figure*}

We include more qualitative examples in \cref{supp:fig:imgnet_temp1_simsc} - \cref{supp:fig:ibot_spair}.

